\documentclass[10pt]{article} % For LaTeX2e
\usepackage[dvipsnames,svgnames]{xcolor}
\usepackage[accepted]{tmlr}
\usepackage{amsmath,amsfonts,bm}

\def\eqref#1{equation~\ref{#1}}
\def\1{\bm{1}}

\DeclareMathAlphabet{\mathsfit}{\encodingdefault}{\sfdefault}{m}{sl}
\SetMathAlphabet{\mathsfit}{bold}{\encodingdefault}{\sfdefault}{bx}{n}

\usepackage[toc,page,header]{appendix}

\usepackage{minitoc}

\usepackage[T1]{fontenc}    % use 8-bit T1 fonts
\usepackage[utf8]{inputenc} % allow utf-8 input
\usepackage[colorlinks=true,citecolor=ForestGreen]{hyperref}       % hyperlinks
\usepackage{url}            % simple URL typesetting
\usepackage{booktabs}       % professional-quality tables
\usepackage{amsfonts}       % blackboard math symbols
\usepackage{amssymb}
\usepackage{nicefrac}       % compact symbols for 1/2, etc.
\usepackage{microtype}      % microtypography

\usepackage{enumitem}
\setlist[itemize]{nosep}       % no extra separation
\setlist[enumerate]{nosep}
\setlist[itemize]{topsep=0pt}	% no top separation
\setlist[enumerate]{topsep=0pt}
\usepackage{setspace}

\usepackage{color}
\usepackage{listings}
\usepackage[ruled,vlined,algo2e]{algorithm2e} 

\usepackage{graphicx}
\usepackage{subcaption}
\usepackage{multirow}

\usepackage{enumitem}
\setlist[itemize]{align=parleft,left=5pt..1.5em}

\newcommand{\nocontentsline}[3]{}
\newcommand{\tocless}[2]{\bgroup\let\addcontentsline=\nocontentsline#1{#2}\egroup}

\newcommand{\methodlong}{Twin Learning for Dimensionality Reduction\xspace}
\newcommand{\method}{TLDR\xspace}
\newcommand{\cne}{\method}
\newcommand{\tldr}{\method}

\newcommand{\methodgauss}{$\textrm{\method}_{\mathcal{G}}$}
\newcommand{\lcne}{$\textrm{\tldr}$}
\newcommand{\ltldrpq}{$\textrm{\method}_{\mathcal{PQ}}$}
\newcommand{\flcne}[1]{$\textrm{\cne}_{#1}$}
\newcommand{\mlpcne}[1]{$\textrm{\cne}^\star_{#1}$}
\newcommand{\pcaw}{$\textrm{PCA}_w$}
\newcommand{\icaw}{$\textrm{ICA}_w$}
\newcommand{\supervised}{Oracle}

\newcommand{\contrproj}{Contrastive\xspace}
\newcommand{\gemap}{GeM-AP\xspace}

\newcommand{\tsne}{$t$-SNE\xspace}
\newcommand{\knn}{$k$-NN\xspace}

\newcommand{\linear}{linear\xspace}
\newcommand{\flinear}{factorized linear\xspace}

\newcommand{\rox}{$\mathcal{R}$Oxford\xspace}
\newcommand{\rpa}{$\mathcal{R}$Paris\xspace}

\newcommand{\gld}{GLD-v2\xspace}
\newcommand{\imnet}{ImageNet\xspace}
\newcommand{\landmarks}{Landmarks\xspace}

\newcommand{\mrcell}[1]{{\begin{tabular}[c]{@{}l@{}}#1\end{tabular}}}
\newcommand{\mrcellc}[1]{{\begin{tabular}[c]{@{}c@{}}#1\end{tabular}}}

\newcommand{\resnet}{ResNet-50\xspace}
\newcommand{\vit}{ViT-S/16\xspace}

\newcommand{\di}{D}
\newcommand{\dd}{d}
\newcommand{\ddd}{d^\prime}

\newcommand{\rdi}{\mathbb{R}^\di}
\newcommand{\rdd}{\mathbb{R}^\dd}
\newcommand{\rddd}{\mathbb{R}^{\ddd}}

\newcommand{\zz}{\hat{z}}

\newcommand{\ef}{f_\theta}
\newcommand{\proj}{g}
\newcommand{\pg}{g_\phi}

\newcommand{\cN}{\mathcal{N}}
\newcommand{\cX}{\mathcal{X}}
\newcommand{\cZ}{\mathcal{Z}}

\newcommand{\cC}{\mathcal{C}}
\newcommand{\cLbt}{\mathcal{L}_{BT}}

\definecolor{DarkGreen}{rgb}{0.0, 0.5, 0.0} 
\definecolor{DarkRed}{rgb}{0.25, 0.0, 0.0}

\newcommand{\gain}[1]{\textbf{\color{Green}{$\uparrow$ #1\%}}}
\newcommand{\losss}[1]{\textbf{\color{DarkRed}{$\downarrow$ #1\%}}}

\newcommand{\dmidrule}{\specialrule{0.8pt}{3pt}{3pt}}

\newcommand{\dlt}[1]{{\color{CornflowerBlue}{#1}}}

\newcommand{\cl}[1]{{\color{PineGreen}{C: #1}}}

\newcommand{\ykt}[1]{{\color{BrickRed}{#1}}}

\renewcommand{\dlt}[1]{{\color{black}{#1}}}

\renewcommand{\ykt}[1]{{\color{black}{#1}}}

\makeatletter
\DeclareRobustCommand\onedot{\futurelet\@let@token\@onedot}
\def\@onedot{\ifx\@let@token.\else.\null\fi\xspace}

\def\eg{\emph{e.g}\onedot} 
\def\ie{\emph{i.e}\onedot} 
\def\cf{\emph{cf}\onedot}

\makeatother

\newlist{inlinelist}{enumerate*}{1}
\setlist*[inlinelist,1]{label=\roman*),itemjoin={{, }},itemjoin*={{, and }}}

\usepackage{pifont}% http://ctan.org/pkg/pifont
\newcommand{\mypartight}[1]{\noindent {\bf #1}}
\usepackage{pgfplots, pgfplotstable}
\pgfplotsset{compat=newest}
\usepgfplotslibrary{fillbetween}

\newcommand{\baselineplotc}{Gray}
\newcommand{\gemwplotc}{Blue}
\newcommand{\cneplotflineartwoc}{RubineRed}
\newcommand{\cneplotflinearonec}{PineGreen}
\newcommand{\cneplotmlponec}{Orchid}
\newcommand{\cneplotlinearc}{CornflowerBlue}
\newcommand{\cneplotgaussianc}{WildStrawberry}
\newcommand{\cneplotsupc}{Turquoise}
\newcommand{\pcaplotc}{black}
\newcommand{\mseplotc}{RoyalPurple}
\newcommand{\contrastiveplotc}{Emerald}
\newcommand{\cneplotlinearplzeroc}{Periwinkle}
\newcommand{\cneplotlinearplonec}{Plum}
\newcommand{\cneplotlinearpltwoc}{MidnightBlue}
\newcommand{\drlimplotc}{Melon}
\newcommand{\cneplotflineartwoktc}{RedViolet}
\newcommand{\cneplotflineartwokhc}{TealBlue}
\newcommand{\cneplotlinearpqc}{RubineRed}

\newcommand{\leg}[1]{\addlegendentry{#1}}

\pgfmathsetmacro{\stdgrad}{30}

\tikzset{every mark/.append style={solid}}
\pgfplotsset{
	grid=both, width=\linewidth, try min ticks=5,
	legend cell align=left, legend style={fill opacity=0.8},
	ylabel near ticks,
    xlabel near ticks,
    every tick label/.append style={font=\footnotesize},
}

\pgfplotsset{
    baselineplot/.style={thick, color=\baselineplotc, mark=x,mark size=5pt, only marks},
    cneplotflineartwo/.style={thick, color=\cneplotflineartwoc, mark=star},
    cneplotflinearone/.style={thick, color=\cneplotflinearonec, mark=triangle*},
    cneplotmlpone/.style={thick, color=\cneplotmlponec, mark=triangle*},
    cneplotlinear/.style={thick, color=\cneplotlinearc, mark=*}, 
    cneplotlinearpq/.style={thick, color=\cneplotlinearpqc, mark=square*}, 
    cneplotgaussian/.style={thick, color=\cneplotgaussianc, mark=diamond*},
    cneplotsup/.style={thick, color=\cneplotsupc, mark=diamond*},
    pcaplot/.style={thick, color=\pcaplotc, mark=square*},   
    mseplot/.style={thick, color=\mseplotc, mark=star},  
    contrastiveplot/.style={thick, color=\contrastiveplotc, mark=star},  
    cneplotlinearplzero/.style={thick, color=\cneplotlinearplzeroc, mark=star},
    cneplotlinearplone/.style={thick, color=\cneplotlinearplonec, mark=triangle*},
    cneplotlinearpltwo/.style={thick, color=\cneplotlinearpltwoc, mark=diamond*},
    drlimplot/.style={thick, color=\drlimplotc, mark=square*},
    cneplotflineartwokt/.style={thick, color=\cneplotflineartwoktc, mark=triangle*},
    cneplotflineartwokh/.style={thick, color=\cneplotflineartwokhc, mark=diamond*},
    gemwplot/.style={thick, color=\gemwplotc, mark=x,mark size=5pt, only marks},
    lppplot/.style={thick, color=\cneplotgaussianc, mark=star}, 
}

\title{TLDR: Twin Learning for Dimensionality Reduction}
\author{\name Yannis Kalantidis \email yannis.kalantidis@naverlabs.com \\ 
\addr NAVER LABS Europe
\AND
\name Carlos Lassance \email carlos.lassance@naverlabs.com \\
\addr NAVER LABS Europe
\AND
\name Jon Almaz\'{a}n \email jon.almazan@naverlabs.com \\
\addr NAVER LABS Europe
\AND
\name Diane Larlus \email diane.larlus@naverlabs.com \\ 
\addr NAVER LABS Europe
}

\def\month{06}  % Insert correct month for camera-ready version
\def\year{2022} % Insert correct year for camera-ready version
\def\openreview{\url{https://openreview.net/forum?id=86fhqdBUbx}} % Insert correct link to OpenReview for camera-ready version

\begin{document}

\maketitle

% -------------------------------------------------------------------------
% Overview
% -------------------------------------------------------------------------

\vspace{-10pt}
% \begin{center}
% %     \large{\url{https://github.com/naver/tldr}}
% \url{https://github.com/naver/tldr}
% \end{center} 

\begin{figure}[h!]
  \centering
    \includegraphics[width=\linewidth]{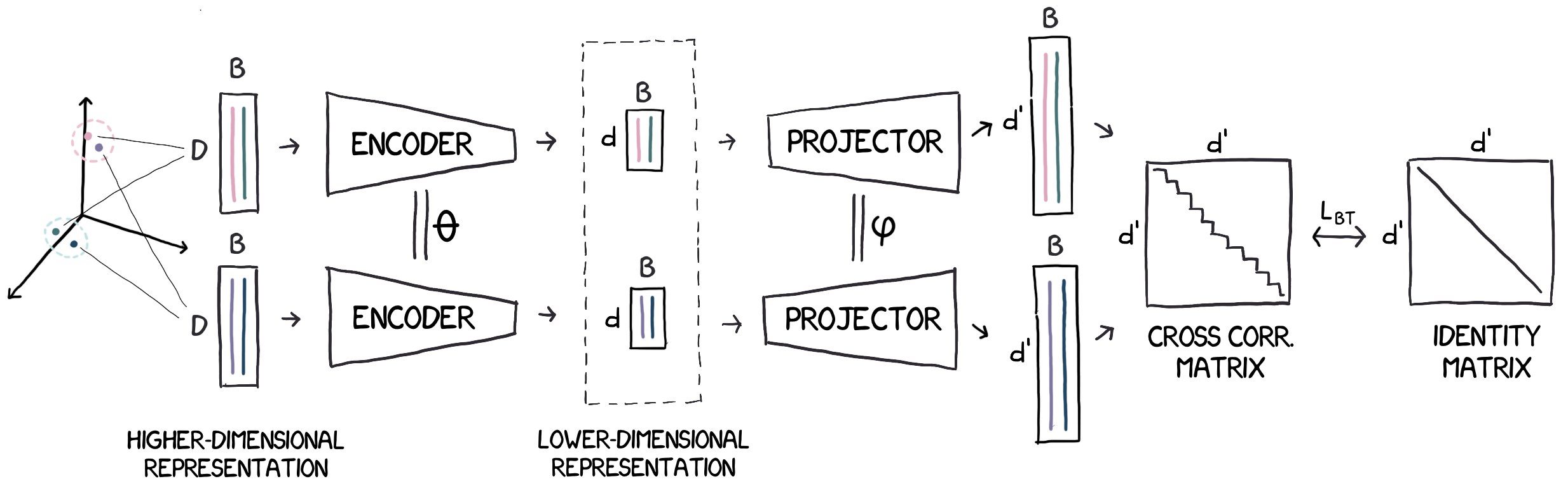}  
    \caption{
    \looseness=-1
    \textbf{Overview of the proposed \tldr dimensionality reduction method.} Given feature vectors from a generic input space, we use nearest neighbors to define a set of feature pairs whose proximity we want to preserve. We then learn a dimensionality-reduction function (the {\em encoder}) by encouraging neighbors in the input space to have similar low dimensional representations using the Barlow Twins loss~\citep{zbontar2021barlow}.
    % We learn it jointly with an auxiliary {\em projector} that produces high dimensional representations, where we compute the Barlow Twins~\citep{zbontar2021barlow} loss over the $\ddd \times \ddd$ cross-correlation matrix averaged over the batch.
    }
  \label{fig:overview}
\end{figure}
% }
% -------------------------------------------------------------------------

\begin{abstract}
\looseness=-1

Dimensionality reduction methods are unsupervised approaches which learn low-dimensional spaces where some properties of the initial space, typically the notion of ``neighborhood'', are preserved. Such methods usually require propagation on large \knn graphs or complicated optimization solvers. On the other hand, self-supervised learning approaches, typically used to learn representations from scratch, rely on simple and more scalable frameworks for learning.
In this paper, we propose \textbf{\tldr}, a dimensionality reduction method for generic input spaces that is porting the recent self-supervised learning framework of~\citet{zbontar2021barlow} to the specific task of dimensionality reduction, over arbitrary representations.
We propose to use nearest neighbors to build pairs from a training set and a redundancy reduction loss to learn an encoder that produces representations invariant across such pairs. 
\tldr is a method that is simple, easy to  train, and 
of broad applicability; it consists of an offline nearest neighbor computation step that can be highly approximated, and a straightforward learning process.
Aiming for scalability, 
we focus on improving \textit{linear} dimensionality reduction,
and  show consistent gains on image and document retrieval tasks, \eg gaining +4\% mAP over PCA on \rox for \gemap, improving the performance of DINO on \imnet or retaining it with a $10\times$ compression.

\begin{center}
Code available at: \url{https://github.com/naver/tldr}    
\end{center}

\end{abstract}

\section{Introduction}
\label{sec:introduction}

\vspace{-5pt}

\looseness=-1
Self-supervised representation learning (SSL) has been shown to produce representations that are highly transferable to a wide number of downstream tasks via encoding invariance to image distortions
% like data augmentations
~\citep{chen2020simple,he2020momentum,caron2020unsupervised}. 
Methods like BYOL~\citep{grill2020bootstrap}, DINO~\citep{caron2021emerging} or Barlow Twins~\citep{zbontar2021barlow} start from \textit{structured inputs} 
% of a specific nature 
(i.e. a 2D grid of pixel intensities) and learn \textit{a large encoder} with hundreds of millions of parameters that transforms this input into a representation vector useful for 
% several 
downstream tasks. 
Central to their success % of such methods 
is the scalable and easy-to-optimize learning framework that such methods adopt, often based on pairwise loss functions with or without constrasting pairs. 

Dimensionality reduction refers to a set of (generally unsupervised) approaches which aim at learning low-dimensional spaces where properties of an initial higher-dimensional input space, \eg proximity or ``neighborhood'', are preserved. Unlike SSL, %it 
dimensionality reduction starts from \textit{any} trustworthy, potentially blackbox, representation and learns a \textit{simple} %(linear or MLP) 
(often linear) encoder without any further assumptions on the nature of the input space
beyond the fact that structure should be preserved. 
% Assuming that data in the input space lie on a lower-dimensional ``manifold'', dimensionality reduction is also referred to as \textit{manifold learning}.
Dimensionality reduction is \textit{a crucial component} in many areas as diverse as biology or AI, and used for tasks ranging from visualization and compression, to indexing and retrieval. This is because, more often than not, learning a small encoder on top of an existing representation is easier than training a lower dimensionality representation end-to-end;
in many cases, training from scratch or fine-tuning a representation is too costly or simply not possible
%. In several cases, training or fine-tuning is simply not possible 
as features might be produced by a blackbox module or sensor. 
% Consequently, dimensionality reduction methods like PCA are still part of state-of-the-art pipelines e.g. for image retrieval~\citep{tolias2020learning,superfeatures}.
Considering
%Seeing 
how manifold learning methods lack scalability and usually require propagation on large \knn graphs or complicated optimization solvers, % one cannot help but wonder:
it is only natural to wonder:
\textit{Can we borrow from  the highly successful learning frameworks of self-supervised representation learning to design dimensionality reduction approaches?}

In this paper, we 
% unify these two families of approaches from the angle of manifold learning and 
propose \textit{\methodlong} or \textbf{\method}, a generic dimensionality-reduction technique where the only prior is that data lies on a reliable manifold we want to preserve. 
It is based on the intuition that comparing a data point and its nearest neighbors is a good ``distortion'' to learn from, and hence a good way of approximating the local manifold geometry.
Similar to other manifold learning methods~\citep{roweis2000nonlinear,van2008visualizing,belkin2003laplacian,donoho2003hessian, hadsell2006dimensionality} we use Euclidean nearest neighbors as a way of defining distortions of the input that the dimensionality reduction function should be invariant to. 
However, unlike other manifold learning methods, \method does not require eigendecompositions, negatives to contrast, or cumbersome optimization solvers; it simply consists of an offline nearest neighbor computation step that can be highly approximated without loss in performance and a straightforward stochastic gradient descent learning process. 
This leads to a highly scalable method that can learn linear and non-linear encoders for dimensionality reduction while trivially handling out-of-sample generalization. We show an overview of the proposed method in Figure~\ref{fig:overview}.

\tldr is meant as a post-hoc dimensionality reduction method, \ie \textit{on top of} learned representations. 
We show that it provides %practitioners with
an easy way of not only compressing representations, but also improving the performance of large state-of-the-art models relying on those reduced representations, \textit{without the need to fine-tune} large %representation 
encoders. 
Training end-to-end or fine-tuning requires large resources, specially when starting from large pre-trained models composed of millions or billions of parameters, and application to certain domains or tasks requires non-trivial know-how.
%End-to-end training requires large resources, especially when starting from ``foundation'' models composed of millions or billions of parameters; fine-tuning those models to certain domains or tasks requires non-trivial know-how. 
%It is much simpler to t
Those issues fade when large models are treated as feature extractors, 
%save the features 
and we only learn a linear or MLP encoder to compress those features vectors in a representation that is more suited to the desired task or dataset. 
% As we show, one can get significant gains and compression when starting from \eg DINO features.
This highlights the natural link between
%This makes an obvious link between 
representation learning and dimensionality reduction: we could use the output of models like Barlow Twins, BYOL, DINO, BERT or any representation learning method as the input for our TLDR encoder. Yet, those two steps are not interchangeable and we focus on the latter.

\looseness=-1
% We are interested in explicitly targeting applications like image and document search where training labels are non-existent and dimensionality reduction is an important part of the state-of-the-art pipelines. 
Aiming at large-scale search applications, we focus on improving \textit{linear} dimensionality reduction with a compact encoder, an integral part of the first-stage of most retrieval systems, and an area where PCA~\citep{pca} is still the default method used in practice~\citep{tolias2020learning, superfeatures}. 
We present a large set of ablations and experimental results on common benchmarks for image retrieval, as well as on the \textit{natural language processing} task of argument retrieval.
We show that one can achieve significant gains without 
altering the encoding and search complexity: for example we can improve landmark image retrieval with
%\ykt{
\gemap~\citep{revaud2019aploss} on ROxford5K~\citep{radenovic2018revisiting} by almost \textit{4 mAP points} for 128 dimensions, a commonly used dimensionality, by simply replacing PCA with \ykt{a linear} \tldr encoder. Similarly, we are able to improve the state-of-the-art retrieval performance of DINO~\citep{caron2021emerging} representations on \imnet~\citep{russakovsky2015imagenet}, even when compressing the vectors tenfold.
%by $10\times$.

We also perform extensive evaluations using DINO models with a ViT backbone and report state-of-the-art retrieval results for fully self-supervised learning methods on multiple datasets like \rox, \rpa, and \imnet. We further show that, given access to unlabeled data from the downstream domain, simply learning a linear \tldr encoder on top of generic 
% (\imnet pre-trained) 
DINO representations leads to large performance gains on the downstream task, without the need to fine-tune the backbone.
% (+2.7\%/+5.4\% over PCA and the original DINO \resnet features when evaluating on \rox after simply learning a linear encoder with unlabeled landmark data)
Finally, we show that \tldr is robust to both approximate neighbor pair generation, as well as subsequent vector quantization.
% and report consistent gains over PCA also after Product Quantization~\citep{jegou2010product}.}
%}

\mypartight{Contributions.} We introduce \method, a dimensionality reduction method that achieves neighborhood embedding learning with the simplicity and effectiveness of recent self-supervised visual representation learning losses. Aiming for scalability, we focus on large-scale image and document retrieval where dimensionality reduction is still an integral component. We show that replacing PCA~\citep{pca} with a \emph{linear} \method encoder can greatly improve the performance of state-of-the-art methods without any additional computational complexity. We thoroughly ablate parameters and show that our design choices allow \tldr to be robust to a large range of hyper-parameters and is applicable to a diverse set of tasks and input spaces. 
% \onlyonarxiv{We also made the code for TLDR publily available at \url{https://github.com/naver/tldr}.}

% -------------------------------------------------------------------------
% \arxivorsub{
    {\section{\methodlong \label{sec:method}}}
% }{
%     {\section{Method \label{sec:method}}}
% }
% -------------------------------------------------------------------------

Starting from a set of unlabeled and high-dimensional features, our goal is to
learn a lower-dimensional space which preserves the local geometry of the larger
input space. Assuming that we have no prior knowledge other than the
reliability of the local geometry of the input space, we use nearest neighbors to define 
a set of feature pairs whose proximity we want to preserve. 
We then learn the parameters of a dimensionality-reduction function (the {\em encoder}) using a loss that encourages neighbors in the input space to have similar representations, while also minimizing the redundancy between the components of these vectors. Similar to other works~\citep{chen2020improved, chen2020simple,zbontar2021barlow} we append a {\em projector} to the encoder that produces a representation in a very high dimensional space, where the Barlow Twins~\citep{zbontar2021barlow} loss is computed. At the end of the learning process, the projector is discarded.
All aforementioned components are detailed next.
We call our method \textbf{\methodlong} or \textbf{\method}, in homage to the Barlow Twins loss. An overview is provided
~in Figure~\ref{fig:overview} and in Algorithm~\ref{alg:overview}.

% \arxivorsub{~in Figure~\ref{fig:overview} and in Algorithm~\ref{alg:overview}.\input{tex/algo}}{~in Figure~\ref{fig:overview}.}

\mypartight{Preserving local neighborhoods.}
Recent self-supervised learning methods define positive pairs via hand-crafted
distortions that exploit prior information from the input space. In absence of
any such prior knowledge, defining local distortions can only be achieved via
assumptions on the input manifold.
Assuming a locally linear manifold, for
example, would allow using the Euclidean distance as a local measure of on-manifold distortion and using nearest neighbors over the training set would be a good approximation for local neighborhoods.
Therefore, we construct pairs of neighboring training vectors, and learn invariance to the distortion from one such vector to another. 
Practically, we define the local neighborhood of each training sample as its $k$ nearest neighbors.
Although defining local neighborhood in such a uniform way over the whole manifold might seem naive, we experimentally show that not only it is sufficient, but also that our algorithm is robust across a wide range of values for $k$ (see Section~\ref{sec:results_image_retrieval}). 

Using nearest neighbors is of course not the only way of defining neighborhoods. In fact, we show alternative results with a simplified variant of \method (denoted as \methodgauss) where we construct pairs by simply adding Gaussian noise to an input vector. This is a baseline resembling denoising autoencoders, although in our case we are using a) an asymmetric encoder-decoder architecture and b) the Barlow twins loss instead of a reconstruction loss. 

\mypartight{Notation.}
Our goal is to learn an encoder $\ef: \rdi \rightarrow \rdd$ that takes as input
a vector $x \in \rdi$ and outputs a corresponding reduced vector $z = \ef(x) \in
\rdd$, with $\dd << \di$. Without loss of generality, we define the encoder to be a neural network
with learnable parameters $\theta$.
Let $\cX$ be a (training) set of datapoints in $\rdi$, the $\di$-dimensional
input space.
Let $x \in \rdi$ be a vector from $\cX$. $\cN_k(x)$ is composed of the $k$
nearest neighbors of $x$. For a
vector $y \in \cX$ from the training set: $y \in \cN_k(x) \Leftrightarrow y
\in \arg_k \min_{y \in \cX} d(x,y)$, where $d(\cdot,\cdot)$ denotes the
Euclidean distance.
Although the definition above can be trivially extended to non-Euclidean distances and adaptive neighborhoods (\eg defined by a radius), without loss of generality we present our method and results with pairs from $k$ Euclidean neighbors. We define \emph{neighbor pairs} as pairs $(x,y) \in \cX \times \cX$ where $y \in \cN_k(x)$.

\mypartight{Learning \`a la Barlow Twins.}
Although contrastive losses were proven highly successful for visual representation learning, explicitly minimizing the redundancy of the output dimension is highly desirable for dimensionality reduction: having a highly informative output space is more important than a highly discriminative one. We therefore choose to learn the parameters of our encoder by minimizing the Barlow Twins loss
function~\citep{zbontar2021barlow}, \cl{which} suits \cl{the problem} perfectly. 
Similar to~\citep{zbontar2021barlow}, we append a projector $\pg$ to the encoder
$\ef$, allowing to calculate the loss in a (third) representation space which is not the one that will be used for subsequent tasks. That extended space can possibly be much larger.
We detail the encoder and the projector later.
Let $\zz = \pg(\ef(x))$ be the output vector of the projector, $\zz \in \rddd$. 
Given a pair of neighbors $(x^A, x^B)$ and the corresponding vectors $\zz^A, \zz^B$ after the projector, the loss function $\cLbt$ is given by:
% \arxivorsub{
\begin{equation}
    \cLbt =\sum_i (1 - \cC_{ii})^2 + \lambda \sum_i \sum_{i \neq j} \cC_{ij}^2, \hspace{6pt} \textrm{ where }  \hspace{2pt} \cC_{ij} = \frac{\sum_b \zz^A_{b,i} \zz^B_{b,j}}{\sqrt{\sum_b(\zz_{b,i}^A)^2} \sqrt{\sum_b(\zz_{b,j}^B)^2}},
\label{eq:bt}
\end{equation}
% }{
% \begin{equation}
%     \cLbt =\sum_i (1 - \cC_{ii})^2 + \lambda \sum_i \sum_{i \neq j} \cC_{ij}^2, 
% \label{eq:bt}
% \end{equation}
% where 
% \begin{equation}
% \cC_{ij} = \frac{\sum_b \zz^A_{b,i} \zz^B_{b,j}}{\sqrt{\sum_b(\zz_{b,i}^A)^2} \sqrt{\sum_b(\zz_{b,j}^B)^2}},
% }
% \label{eq:btc}
% \end{equation}

where $b$ indexes the positive pairs in a batch, $i$ and $j$ are two dimensions
from $\rddd$ (\ie $0 \leq i,j \leq d'$) and $\lambda$ is a hyper-parameter. $\cC$ is the
$\ddd \times \ddd$ cross-correlation matrix computed and averaged over all
positive pairs $(\zz^A, \zz^B)$ from the current batch.
The loss is composed of two terms. The first term encourages
the diagonal elements to be equal to 1. This makes the learned representations
invariant to applied distortions, \ie the datapoints moving along the input manifold in the neighborhood of a training vector are encouraged to share similar representations in the output space. The second term is pushing off-diagonal elements towards 0, reducing the redundancy between output dimensions, a highly desirable property for dimensionality reduction. 

The redundancy reduction term can be viewed as a soft-whitening constraint on the representations and, as shown in~\citet{zbontar2021barlow}, it works better than performing ``hard'' whitening on the representations~\citep{ermolov2021whitening}. 
Finally, it is worth noting that understanding the dynamics of learning without contrasting pairs is far from trivial and beyond the scope of this paper; we refer the reader to the recent work by~\citet{tian2021understanding} that studies this learning paradigm in depth and discusses why trivial solutions are avoided when learning without negatives as in Eq.~(\ref{eq:bt}).

% For dimensionality reduction, every dimension of the output space counts; implicitly enforcing every output dimension to be maximally de-correlated make this loss a very well fitting loss for dimensionality reduction

% ------------------------------------------------------------------------
\begin{algorithm2e}[t]
\SetAlgoLined
%\setstretch{1.7}
\setstretch{1}
\SetKwInput{KwData}{Input} 
\SetKwInput{KwResult}{Output}

\KwData{Training set of high-dimensional vectors $\cX$, hyper-parameter $k$}
\KwResult{Corresponding set of lower-dimensional vectors $\cZ$, \newline
  dimensionality-reduction function $\ef: \rdi \rightarrow \rdd$}
  
1: Calculate the $k$ (approximate) nearest neighbors, \ie $\cN_k(x)$, for every $x \in \cX$.

2: Create positive pairs $(x,y)$ by sampling $y$ from the set $\cN_k(x)$.

3: Learn the parameters $\theta$ of $\ef$ and $\phi$ of $\pg$ by optimizing
the Barlow Twins' loss (Eq.~(\ref{eq:bt})).

\caption{\methodlong (\method)}
\label{alg:overview}
\end{algorithm2e}
% ------------------------------------------------------------------------

\mypartight{The encoder $\ef$.} We consider a number of different architectures for the encoder:
\begin{itemize}[topsep=0pt, align=parleft,left=2pt..1.5em]
   \item \emph{\linear}: The most straight-forward choice for encoder $\ef$ is a linear function parametrized by a $\di \times \dd$ weight matrix $W$ and bias term $b$, \ie $\ef(x) = Wx + b$. Beyond computational benefits, and given that we are mostly interested in medium-sized output spaces where $\dd \in \{8,\ldots,512\}$, we argue that, given a meaningful enough input space, a linear encoder could suffice in preserving neighborhoods of the input. 
   \item \emph{\flinear}: Exploiting the fact that batch normalization
     (BN)~\citep{ioffe2015batch} is linear during inference,\footnote{Although
       batch normalization is non-linear during training because of the reliance
       on the current batch statistics, during inference and using the means and
       variances accumulated over training, it reduces to a linear scaling
       applied to the features, that can be embedded in the weights of an adjacent linear layer.} we formulate $\ef$ as a multi-layer linear
     model, where $\ef$ is a sequence of $l$  layers, each composed of a
     linear layer followed by a BN layer. This model introduces non-linear dynamics which can
     potentially help during training but the sequence of layers can still be
     replaced with a single linear layer after training for efficiently encoding new features.
   \item \emph{MLP}: $\ef$ can be a multi-layer perceptron with batch normalization (BN)~\citep{ioffe2015batch} and rectified linear units (reLUs) as non-linearities, \ie $\ef$ would be a sequence of $l$ linear-BN-reLU triplets, each with $H^i$ hidden units ($i = 1,..,l$), followed by a linear projection.
\end{itemize}

Our main goal is to develop a scalable alternative to PCA for
dimensionality reduction, so we are mostly interested in \linear
and \flinear encoders. It is worth already mentioning that, as we will show in our experimental validation, gains from introducing an MLP in the encoder are minimal and would not justify the added computational cost in practice.

\mypartight{The projector $\pg$.} As also recently noted in~\citet{tian2021understanding}, a crucial part of learning with non-contrastive pairs is the projector.
This module is present in a number of contrastive self-supervised learning
methods~\citep{chen2020simple,grill2020bootstrap,zbontar2021barlow, tian2021understanding}.
It is usually
implemented as an MLP inserted between the transferable representations and the loss function. Unlike other methods, however, where the projector takes the
representations to an even lower dimensional space for the contrastive loss to
operate on (\ie for SimCLR~\citep{chen2020simple} and BYOL~\citep{grill2020bootstrap},
$\ddd \ll \dd$), for the Barlow Twins objective, operating in large output
dimensions is crucial. 

In Section~\ref{sec:experiments}, we study the impact of the dimension $\ddd$ and experiment with a wide range of values. We empirically verify the findings of \cite{zbontar2021barlow} that calculating the de-correlation loss in higher dimensions ($\ddd \gg \dd$) is highly
beneficial. In this case, and as shown in Figure~\ref{fig:overview}, the
transferable representation is now the \emph{bottleneck} layer of this non-symmetrical
hour-glass model. Although Eq.~(\ref{eq:bt}) is applied after the
projector and only indirectly decorrelates the output representation components, having more dimensions to decorrelate leads to a representation that is more informative: the \textit{bottleneck} effect created by the projector's output being in a much larger dimensionality implicitly enables the network to learn an encoder that also has more decorrelated outputs. 

\section{Experimental validation \label{sec:experiments}}

In this section, we present a set of experiments validating the proposed \method on both visual and textual representations. In Section~\ref{sec:tasks_and_datasets} we first present a summary of the tasks we evaluate on. Then in Sections~\ref{sec:results_image_retrieval} and~\ref{sec:results_nlp} we present results for visual and textual applications, respectively.
% , focusing on linear dimensionality reduction and large-scale applications. 
Finally, in Section~\ref{sec:results_analysis} we study the different hyper-parameters of our method, compare to several manifold learning methods, and present results when approximating the nearest neighbor pairs, as well as results after subsequent vector quantization.

\subsection{Summary of tasks and evaluation protocol}
\label{sec:tasks_and_datasets}

A summary of all the tasks, datasets and representations that we consider
%the main paper 
in this section is presented in Table~\ref{tab:tasks_and_summary}.
We explore tasks like landmark image retrieval on \rox and \rpa, object class retrieval on \imnet, and argument retrieval on ArguAna.
For all experiments, we start from reliable feature vectors; it is a pre-requirement for the dimensionality reduction task.
We assume that any structured data (images or documents) is first encoded with a suitable representation, and, without loss of generality, we assume the Euclidean distance to be meaningful, at least locally, in this input representation space.
% to encode any data not in vector form (images/documents) with some representation learning model into a  dimensional input space where, without loss of generality, the Euclidean distance is meaningful at least locally. 
%\alert{got lost in this last paragraph - TODO}

In Section~\ref{sec:results_image_retrieval} we evaluate \tldr on popular image retrieval tasks like \textit{landmark image retrieval} on datasets like \rox, \rpa~\citep{radenovic2018revisiting}, or $k$-NN retrieval on \imnet~\citep{russakovsky2015imagenet}. We start from both specialized, retrieval-oriented representations like  \gemap~\citep{revaud2019aploss}, as well as more generic representations learned via self-supervised learning like DINO~\citep{caron2021emerging}. For all experiments, we use the DINO and AP-GeM models as feature extractors to encode images for visual tasks; we use global image features from publicly avalable models for \gemap\footnote{\url{https://github.com/naver/deep-image-retrieval}} or DINO\footnote{\url{https://github.com/facebookresearch/dino}}.
For the textual domain, we focus on the task of \textit{argument retrieval} (Section~\ref{sec:results_nlp}). We use 768-dimensional features from an off-the-shelf Bert-Siamese model called ANCE\footnote{\url{https://www.sbert.net/docs/pretrained_models.html}}~\citep{xiong2020approximate} trained for document retrieval, following the dataset definitions from%
% a recent benchmark
~\citet{thakur2021beir}. 

We %never 
do not use any supervision for learning TLDR nor any of the methods we compare to. Given a trained % learned
dimensionality reduction encoder, we then encode all downstream task data for each test dataset, and evaluate them in a “zero-shot” manner, using non-parametric classifiers ($k$-NN) for all retrieval tasks. Specifically, for landmark image retrieval on \rox/\rpa we use the common protocols presented %in
by~\citet{radenovic2018revisiting}, where specific queries are defined.
For every query, we measure the mean average precision metric across all other images depicting the same landmark. For \imnet~\cite{russakovsky2015imagenet}, we follow the exact process used by \citet{caron2021emerging} and others: 
The gallery is composed of the full validation set, spanning 1000 classes, and each image from this validation (val) set is used in turn as a query.
%Each image in the validation (val) set is used as a query for each class, and the ``database'' consists of the full validation set across all 1000 classes. 
Once again, we use $k$-NN to compute a list of images ranked by their relevance to the query, and we aggregate the labels from the top 20 images, assigning the predicted query label to the most prominent class.
%a ranked list for the database, and aggregate the labels from the top-20 ranked database images for each query, assigning the most prominent class as the predicted one.
In Table~\ref{tab:tasks_and_summary} we report a summary of the most notable results in each case. We \dlt{chose} $d=128$ (resp. 64) for the visual (resp. textual) domains as the most commonly used setting in practice. 

% \onlyinsub{
% \emph{\textbf{Due to space constraints, further ablations, results  on FashionMNIST, on duplicate query retrieval, as well as 2D visualizations can be found in the supplementary material.}}

% }

% ----------------------------------------------------------------
% Table with tasks and datasets
% \input{tables/table_tasks}
\begin{table}
\caption{\textbf{Tasks, models and summary for \textit{linear} dimensionality reduction.}  $^\dagger$ denotes the use of the dataset without labels. The \gemap and DINO models are from~\citet{revaud2019aploss} and \citet{caron2021emerging}, respectively. 
% The underlined result highlights a new \textit{state-of-the-art} for self-supervised learning on \rox (\rpa); \tldr with $d=256$ achieves \textbf{+4.6} (\textbf{+2.3}) mAP higher than the result reported in~\citet{caron2021emerging}.
Different values of $d$ for \tldr and PCA are denoted in parenthesis next to the \tldr performance when $d \ne 128$. $^\ddagger$ input space $D=384$ for \vit; $^\mathsection$ input space $D=768$ for BERT. \label{tab:tasks_and_summary}}
    \setlength{\tabcolsep}{7pt}
    \renewcommand{\arraystretch}{1.1}
    \resizebox{\linewidth}{!}{
    \centering
    \begin{tabular}{ll | ll | l | ccc}
    \toprule    
    \multirow{2}{*}{\textbf{Task}} & 
    \multirow{2}{*}{\mrcell{\textbf{Retrieval dataset} \\ \textbf{(Metric)}}} & 
    \multicolumn{2}{c|}{\textbf{Representation}} & 
    \textbf{Dim. red.} & 
    \multicolumn{3}{c}{\textbf{Result Summary}}  \\
    &    &    \textbf{Model}  & 
    \textbf{Dataset} &
    \textbf{dataset} &
    \mrcellc{\textbf{\tldr} \\ ($d=128$)}&
    \mrcellc{\textbf{vs \pcaw}  \\  ($d=128$)}&
    \mrcellc{\textbf{vs Orig.}  \\ ($D=2048$)} \\ \midrule
    \multirow{8}{*}{\mrcell{Landmark \\ Retrieval} } &  
    % % Paris GEM-AP
    % \multirow{2}{*}{\rpa (\textit{mAP})} & \gemap (\resnet)  & \landmarks & \gld$^\dagger$ &  0.67 & \gain{1.2} & \gain{0.6} \\  \cmidrule{3-8}
    % && DINO (\vit) & \gld$^\dagger$  & \gld$^\dagger$ & \underline{0.??} & \gain{?.?} & \gain{?.?}$^\ddagger$ \\     \cmidrule{2-8} 
    % %
    % Oxford GEM-AP
     \multirow{6}{*}{\rox (\textit{mAP}) } &  \gemap (\resnet)  & \landmarks & \gld$^\dagger$ & 0.49 & \gain{4.1} & \gain{0.7}  \\   \cmidrule{3-8}
    %
    % DINO Resnet
    & & \multirow{2}{*}{DINO (\resnet)}   & \multirow{2}{*}{\imnet$^\dagger$}    & \imnet$^\dagger$ & 0.22 & \gain{2.9} & \losss{0.8}\\% \cmidrule{5-8}
    & & & & \gld$^\dagger$ & 0.29 & \gain{3.7} &\gain{6.8}  \\  \cmidrule{3-8}
    %
    % DINO ViT
    & & \multirow{3}{*}{DINO (\vit)}  &    
    \multirow{2}{*}{\imnet$^\dagger$}  & \imnet$^\dagger$
    & 0.25 & \gain{2.6} & \losss{0.2}$^\ddagger$ \\  %\cmidrule{5-8}
    & & & & \gld$^\dagger$ 
    & 0.31 & \gain{2.7} & \gain{5.4}$^\ddagger$

    \\ \cmidrule{4-8}
    % & & & \gld$^\dagger$  & \gld$^\dagger$ & \underline{0.41} & \gain{2.8} & \underline{\gain{4.6}}$^\ddagger$ \\ 
    & & & \gld$^\dagger$  & \gld$^\dagger$ & {0.43} (256) & \gain{1.9} & {\gain{4.6}}$^\ddagger$ \\ 
    % 
    % Paris GEM-AP
    \cmidrule{2-8}
    & \multirow{1}{*}{\rpa (\textit{mAP})} & \gemap (\resnet)  & \landmarks & \gld$^\dagger$ &  0.67 & \gain{1.2} & \gain{0.6} \\  
    % \cmidrule{3-8}
    % && DINO (\vit) & \gld$^\dagger$  & \gld$^\dagger$ 
    % & \underline{0.63} & \gain{1.6} & \gain{0.0}$^\ddagger$ \\    
    % & \underline{0.66} (256) & \gain{0.0} & \gain{0.0}$^\ddagger$ \\    
    %
    \dmidrule
    % Imagenet
    \multirow{2}{*}{\mrcell{ImageNet \\ Retrieval}}&
    \multirow{2}{*}{\mrcell{\imnet \\ ($k$-NN \textit{Top-1 Acc.})}} &
    \multirow{1}{*}{DINO (\resnet)}&
    \multirow{1}{*}{\imnet$^\dagger$}&
    \multirow{1}{*}{\imnet$^\dagger$} &
    % ($d=256$)& ($d=256$) &  \\  % \cmidrule{5-8}
    % &&&&& 67.9 & \gain{1.8} & \gain{0.4} \\  
    % &&&&& 
    % ($d=512$)& ($d=512$) &  \\% ($D=2048$) \\  % \cmidrule{5-8}
    % for TLDR linear
    % &&&&& 68.4 (512) & \gain{1.3} & \gain{0.9} \\  
     68.4 (512) & \gain{1.3} & \gain{0.9} \\  
    % for TLDR-F-linear
    % &&&&& 68.9$^\mathparagraph$ & \gain{1.8} & \gain{1.4} \\  
    &&\multirow{1}{*}{DINO (\vit)}&
    \multirow{1}{*}{\imnet$^\dagger$}&
    \multirow{1}{*}{\imnet$^\dagger$} &
    % ($d=256$)& ($d=256$) &  \\% ($D=2048$) \\  % \cmidrule{5-8}
    % for TLDR linear
    % &&&&& 74.8 (256) & \gain{0.6} & \gain{0.3}$^\ddagger$ \\  
     74.8 (256) & \gain{0.6} & \gain{0.3}$^\ddagger$ \\  
    % for TLDR-F-linear
    % &&&&& 74.9 & \gain{0.7} & \gain{0.4} \\  
    \dmidrule
    \mrcell{Argument \\ Retrieval}
    % &&&&&($d=64$)& ($d=64$) & \\%($D=768$) \\  % \cmidrule{5-8}
     & ArguAna (\textit{Rec@100}) & ANCE (BERT) & MSMarco & WT20$^\dagger$ 
     & 94.5 (64) & \gain{0.8} & \gain{0.5}$^\mathsection$ \\ 
    \bottomrule
    \end{tabular}
    }

\end{table}
% -------------------------------------------------------------------------

% ----------------------------------------------------------------

\mypartight{Implementation details.} 
We do not explicitly normalize representations during learning \method; yet, we follow the common protocol and L2-normalize the features before retrieval for both tasks. 
Results reported for PCA use whitened PCA; we tested multiple whitening power values and kept the ones that performed best. Further implementation details are reported in the 
% \arxivorsub{Appendix}{supplementary material}.
Appendix.
It is noteworthy that we used the \textit{exact same hyper-parameters} for the learning rate, weight decay, scaling, and $\lambda$ suggested in~\citet{zbontar2021barlow}, despite having very different tasks and encoder architectures.
We were able to run PCA on up to millions of data points and hundreds of dimensions using out-of-the-box tools. More precisely, we use the PCA implementation from scikit-learn.\footnote{\url{https://scikit-learn.org/stable/modules/generated/sklearn.decomposition.PCA.html}} 
For large matrices (> 500x500, which is our case) it uses the randomized SVD approach by \citet{halko2011finding}, which is an approximation of the full SVD. This has been the standard way of scaling PCA to large matrices.

\mypartight{Measuring variance.} The proposed method has some inherent stochasticity, e.g. from SGD or from the neighbor pair sampling step (Step 2 in Algorithm 1). To make sure we properly measure variance, we run each variant of TLDR shown in Figures 2 and 6 for five times and average the output results; the error bars report the standard deviation across those 5 runs. The reason we only report error bars for TLDR is because it has some noticeable variance; all other methods either correspond to deterministic algorithms or had negligible variance and therefore the error bars are not visible.

\mypartight{Further ablations, results on FashionMNIST and 2D visualizations.} 
Additional results and interesting ablations can be found in the Appendix. 
In particular, we explore the effect of the training set size, the batch size and report results on another NLP task: duplicate query retrieval.
Moreover, and although beyond the scope of what \tldr is designed for, in Appendix~\ref{sec:experiments_fashionmnist} we present results on FashionMNIST when using \tldr on raw pixel data and for 2D visualization. 
We show that for cases where the input pixels are forming an informative space, \tldr can achieve top performance for $\dd \ge 8$.

% -------------------------------------------------------------------------
{\subsection{Results on image retrieval \label{sec:results_image_retrieval}}}
% -------------------------------------------------------------------------

We first focus on landmark image retrieval. For large-scale experiments on this task, it is common practice to apply dimensionality reduction to global normalized image representations using PCA with whitening~\citep{jegouchum12,tolias16particular,revaud2019aploss,tolias2020learning}. For the experiments in this section, we simply replace the standard PCA step with our proposed \method. 
%\ykt{
We conduct most comparisons and ablations using 
the optimized, retrieval-oriented \gemap representations proposed by \citet{revaud2019aploss} that are tailored to the landmark image retrieval task and present results in Section~\ref{sec:landmark_gemap}. We then report results using the generic DINO representations that are learned in a self-supervised manner in Section~\ref{sec:landmark_dino}.
%}

% ----------------------------------------------------
{\subsubsection{Landmark image retrieval using \gemap representations \label{sec:landmark_gemap}}}
% ----------------------------------------------------

\mypartight{Experimental protocol.}
We start from 2048-dimensional features obtained from the pre-trained ResNet-50 of~\citep{revaud2019aploss}, which uses Generalized-Mean pooling~\citep{radenovic2018fine} and has been specifically trained for landmark retrieval using the AP loss (GeM-AP).
To learn the dimensionality reduction function, we use \gld, a dataset composed of 1.5 million landmark images~\citep{weyand2020GLDv2}.
We learn different output spaces whose dimensions range from 32 to 512. 
We evaluate these spaces on two standard image retrieval benchmarks~\citep{radenovic2018revisiting}, the revisited Oxford and Paris datasets (\rox and \rpa). Each dataset comes with two test sets of increasing difficulty, the ``Medium'' and ``Hard''. Following these datasets' protocol, we
apply the learned dimensionality reduction function to encode both the gallery images and the set of query images whose 2048-dim features have been extracted beforehand with the model of~\citet{revaud2019aploss}. We then evaluate landmark image retrieval on ROxford5K and RParis6K and report mean average precision (mAP), the standard metric reported for these datasets. 
For brevity, we report the ``Mean'' mAP metric, \ie the average of the mAP of the ``Medium'' and ``Hard'' test sets; we include the individual plots for ``Medium'' and ``Hard'' in the Appendix for completeness.

\mypartight{Compared approaches.}
We report results for several flavors of our approach. \textit{\method} uses a
linear projector, \textit{\flcne{1}} uses a factorized linear one, and \textit{\mlpcne{1}} an MLP encoder with 1 hidden layer. 
As an alternative, we also report \textit{\methodgauss}, which uses Gaussian noise to create synthetic neighbour pairs. All variants use an MLP with 2 hidden layers and 8192 dimensions as a projector. 

% \onlyonarxiv{
\begin{table}
\caption{\textbf{Compared Methods.} For \textit{unsupervised} methods, the objective is based on reconstruction. \textit{Neighbor-supervised} methods use nearest neighbors as pseudo-labels to guide the learning. \emph{Denoising} learns to ignore added Gaussian noise. 
Note that in Figure~\ref{fig:roxford5k_umap}, we compare to a few additional manifold learning methods (\ie ICA, LLE, UMAP, LPP and LTSA) that could not scale to large output dimensions.
% \ykt{Note that in Figure~\ref{fig:roxford5k_umap} we further compare to manifold learning methods like ICA, Locally Linear Embedding (LLE)~\citep{roweis2000nonlinear}, Local Tangent Space Alignment (LTSA)~\citep{zhang2004principal}, \ykt{Locality Preserving Projections (LPP)~\citep{he2003locality}} and UMAP~\citep{mcinnes2018umap}; these methods could only scale comfortably to output dimensions $\le 32$.} 
\label{tab:comparedmethods}}
    \centering
    \resizebox{\textwidth}{!}{

\begin{tabular}{l|l|lll|l}
\toprule
Method       & (Self-) supervision           & Encoder           & Projector & Loss                               & 
Notes
     
\\ \midrule
\begin{tabular}[c]{@{}l@{}}PCA\\~~\citep{pca} \end{tabular}        & unsupervised & linear            & linear    & \begin{tabular}[c]{@{}l@{}} Reconstruction MSE \\ + orthogonality \end{tabular}    & 
\begin{tabular}[c]{@{}l@{}}
Used for dimensionality reduction \\
in SoTA methods like DELF, GeM, \gemap and HOW \end{tabular} \\
\midrule
 DrLim & neighbor-supervised           & MLP               & None      & Contrastive                        &            \begin{tabular}[c]{@{}l@{}} \citep{hadsell2006dimensionality} (\textit{very low performance}) \end{tabular}
 \\
\contrproj   & neighbor-supervised           & linear            & MLP       & Contrastive                        & \citet{hadsell2006dimensionality} with projector

\\ \midrule
 MSE          & unsupervised  & linear            & MLP       & Reconstruction MSE                        &    \begin{tabular}[c]{@{}l@{}}      \tldr with MSE loss  \end{tabular}
\\ 
\methodgauss       & denoising              & linear            & MLP       & Barlow Twins                      &     \tldr with noise as distortion                                                                                                              
\\ \midrule
\lcne         & neighbor-supervised           & linear            & MLP       & Barlow Twins                      &                                                                                                                                 \\
\flcne{1,2}  & neighbor-supervised           & fact. linear & MLP       & Barlow Twins                      &                                                                                                                                 \\
\mlpcne{1,2}  & neighbor-supervised           & MLP               & MLP       & Barlow Twins                      &               \\ \bottomrule                                                                                                                
\end{tabular}
}   
% \vspace{4pt}

\end{table}
% }

We compare with a number of un- and self-supervised methods
% \onlyonarxiv{
(see also Table~\ref{tab:comparedmethods} for a summary).
% } 
First, and foremost, we compare to reducing the dimension with \textit{PCA} with whitening, which is still standard practice for these datasets~\citep{revaud2019aploss,radenovic2018fine,tolias2020learning}. 
We also report results for our approach but trained with the Mean Square Error reconstruction loss instead of the Barlow Twins' (as we discuss in Section~\ref{sec:discussion}, PCA can be rewritten as learning a linear encoder and projector with a reconstruction loss), and refer to this method as \textit{MSE}. In this case, the projector's output is reduced to 2048 dimensions in order to match the input's dimensionality.
Following a number of approaches that use nearest neighbors as (self-)supervision for contrastive learning~\citep{hadsell2006dimensionality}, the \textit{\contrproj} approach uses a contrastive loss on top of the projector's output. This draws inspiration from~\citet{hadsell2006dimensionality}, and is a variant where we replace the Barlow Twins loss, with the loss from~\citet{hadsell2006dimensionality}. 
It is worth noting that we omit results from a more faithful reimplementation of~\citet{hadsell2006dimensionality}, \ie  using a max-margin loss directly applied on the lower dimensional space and without a projector, as they were very low.
Note that we considered other manifold learning methods, 
but none of the ones we tested was able to neither scale, nor outperform PCA in output dimensions $\dd \ge 8$; we present comparisons for smaller $\dd$ in Section~\ref{sec:results_analysis}.
Finally, we report retrieval results obtained on the initial features from~\citet{revaud2019aploss} (\textit{\gemap}), \ie without dimensionality reduction. 
For all flavours of \method, we fix the number of nearest neighbors to $k=3$, although, and as we show in Figure~\ref{fig:roxford5k_paris6k_ablative}, TLDR performs well for a wide range of number of neighbors. 

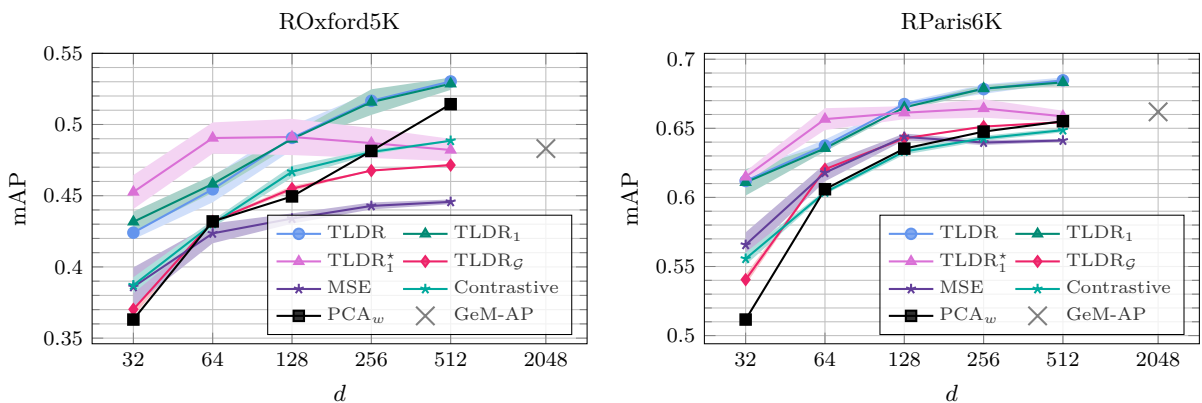
\begin{figure}
\begin{center}
% -------------------------------------------------------------------------
% Oxford results
% -------------------------------------------------------------------------
\resizebox{\linewidth}{!}{
    \begin{subfigure}{.5\linewidth}
        \centering
        \begin{tikzpicture}
\begin{axis}[%
  width=\linewidth,
  xlabel={$\dd$},
  xtick = {1,2,3,4,5,6.2},
  xticklabels = {32,64,128,256,512,2048},
  ylabel={mAP},
  legend pos=south east,
  ylabel near ticks, xlabel near ticks, legend style={font=\scriptsize}, legend columns=2,
  height=5.5cm,
  minor y tick num=4,
  title={ROxford5K},
  label style={font=\small},
  title style={font=\small}
  ]

\pgfplotstableread{
d CNE_0-h2048 CNE_0-h2048-std CNE_1-h2048 CNE_1-h2048-std CNE_2-h2048 CNE_2-h2048-std CNE_1*-h2048 CNE_1*-h2048-std CNE_2*-h2048 CNE_2*-h2048-std Recon_0 Recon_0-std Synth-nn Synth-nn-std Contrastive Contrastive-std Contrastive-MLP Contrastive-MLP-std Contrastive+Proj Contrastive+Proj-std Supervised Supervised-std PCA-whiten baseline
1	0.42404	0.00460678846920498	0.43165	0.00712550349098223	0.43073	0.00667260818570969	0.45259	0.0119099832073769	0.40014	0.0174142039726196	0.38598	0.0136581861899741	0.37022	0.0047986769009801	0.26708	0.000568726647872245	0.24203	0.0293233507635127	0.38712	0.00563619108973427	0.41153	0.00385303127420477	0.36304	nan
2	0.45450	0.00915293668720592	0.45823	0.00612896402338927	0.45933	0.00714059521328579	0.49050	0.0109571095641141	0.45578	0.0108068982599079	0.42356	0.00696536072289153	0.43191	0.000831504660239496	0.26738	0.00106773123959169	0.22332	0.00951803551159587	0.43024	0.00257166288614974	0.45343	0.00648217941744904	0.43195	nan
3	0.49060	0.0086457648591666	0.48981	0.00396137602355545	0.49280	0.00261961829280527	0.49124	0.0127173365922272	0.47042	0.0104079440813256	0.43397	0.00360112482427366	0.45500	0.00184295957633368	0.26666	0.00204015930750518	0.24050	0.0202901577125462	0.46685	0.00413043581235685	0.49858	0.00351688072018373	0.44950	nan
4	0.51662	0.00586	0.51571	0.008988228412763	0.50844	0.00427041567063442	0.48695	0.0104983522516631	0.45945	0.00923851990310136	0.44282	0.0024481217289996	0.46764	0.000303726850969749	0.26841	0.00136625400273888	0.23162	0.0136512362077579	0.48067	0.00169477137101144	0.51928	0.00183723977749231	0.48142	nan
5	0.53022	0.00272767300092955	0.52858	0.00432195557589386	0.52079	0.00551962408140265	0.48210	0.00782007672596631	0.46916	0.00474704645016246	0.44568	0.00166950591493412	0.47148	0.000865852181379709	0.26991	0.000338008875623111	0.23357	0.0192809556298437	0.48858	0.000450111097397076	0.53192	0.00320035154319022	0.51426	nan
6.2	nan	nan	nan	nan	nan	nan	nan	nan	nan	nan	nan	nan	nan	nan	nan	nan	nan	nan	nan	nan	nan	nan	nan	0.48309
}{\map}
    
    \addplot[cneplotlinear]      table[x=d,  y=CNE_0-h2048]   \map; \leg{\lcne}
    \addplot [name path=upper,draw=none, forget plot] table[x=d,y expr=\thisrow{CNE_0-h2048}+\thisrow{CNE_0-h2048-std}] {\map};
    \addplot [name path=lower,draw=none, forget plot] table[x=d,y expr=\thisrow{CNE_0-h2048}-\thisrow{CNE_0-h2048-std}] {\map};
    \addplot [fill=\cneplotlinearc!\stdgrad, forget plot] fill between[of=upper and lower];
    
    \addplot[cneplotflinearone]      table[x=d,  y=CNE_1-h2048]   \map; \leg{\flcne{1}}
    \addplot [name path=upper,draw=none, forget plot] table[x=d,y expr=\thisrow{CNE_1-h2048}+\thisrow{CNE_1-h2048-std}] {\map};
    \addplot [name path=lower,draw=none, forget plot] table[x=d,y expr=\thisrow{CNE_1-h2048}-\thisrow{CNE_1-h2048-std}] {\map};
    \addplot [fill=\cneplotflinearonec!\stdgrad, forget plot] fill between[of=upper and lower];

    \addplot[cneplotmlpone]      table[x=d,  y=CNE_1*-h2048]   \map; 
    \leg{\mlpcne{1}}
    \addplot [name path=upper,draw=none, forget plot] table[x=d,y expr=\thisrow{CNE_1*-h2048}+\thisrow{CNE_1*-h2048-std}] {\map};
    \addplot [name path=lower,draw=none, forget plot] table[x=d,y expr=\thisrow{CNE_1*-h2048}-\thisrow{CNE_1*-h2048-std}] {\map};
    \addplot [fill=\cneplotmlponec!\stdgrad, forget plot] fill between[of=upper and lower];

    \addplot[cneplotgaussian]      table[x=d,  y=Synth-nn]   \map; 
    \leg{\methodgauss}
    \addplot [name path=upper,draw=none, forget plot] table[x=d,y expr=\thisrow{Synth-nn}+\thisrow{Synth-nn-std}] {\map};
    \addplot [name path=lower,draw=none, forget plot] table[x=d,y expr=\thisrow{Synth-nn}-\thisrow{Synth-nn-std}] {\map};
    \addplot [fill=\cneplotgaussianc!\stdgrad, forget plot] fill between[of=upper and lower];
    
    \addplot[mseplot]      table[x=d,  y=Recon_0]   \map; 
    \leg{MSE}
    \addplot [name path=upper,draw=none, forget plot] table[x=d,y expr=\thisrow{Recon_0}+\thisrow{Recon_0-std}] {\map};
    \addplot [name path=lower,draw=none, forget plot] table[x=d,y expr=\thisrow{Recon_0}-\thisrow{Recon_0-std}] {\map};
    \addplot [fill=\mseplotc!\stdgrad, forget plot] fill between[of=upper and lower];

    \addplot[contrastiveplot]      table[x=d,  y=Contrastive+Proj]   \map; 
    \leg{\contrproj}
    \addplot [name path=upper,draw=none, forget plot] table[x=d,y expr=\thisrow{Contrastive+Proj}+\thisrow{Contrastive+Proj-std}] {\map};
    \addplot [name path=lower,draw=none, forget plot] table[x=d,y expr=\thisrow{Contrastive+Proj}-\thisrow{Contrastive+Proj-std}] {\map};
    \addplot [fill=\contrastiveplotc!\stdgrad, forget plot] fill between[of=upper and lower];

    \addplot[pcaplot]      table[x=d,  y=PCA-whiten]   \map; \leg{\pcaw}

    \addplot[baselineplot] table[x=d,  y=baseline]   \map; \leg{\gemap}

\end{axis}
\end{tikzpicture}
        % \caption{ROxford5K-Medium test set.}
        \label{fig:roxford5k_avg}
    \end{subfigure}%
    \begin{subfigure}{.5\linewidth}
        \centering
        \begin{tikzpicture}
\begin{axis}[%
  width=\linewidth,
  xlabel={$\dd$},
  xtick = {1,2,3,4,5,6.2},
  xticklabels = {32,64,128,256,512,2048},
  ylabel={mAP},
  legend pos=south east,
  ylabel near ticks, xlabel near ticks, legend style={font=\scriptsize},
  legend columns=2,
  height=5.5cm, 
  minor y tick num=4,
  title={RParis6K},
  label style={font=\small},
  title style={font=\small}
  ]

\pgfplotstableread{
d CNE_0-h2048 CNE_0-h2048-std CNE_1-h2048 CNE_1-h2048-std CNE_2-h2048 CNE_2-h2048-std CNE_1*-h2048 CNE_1*-h2048-std CNE_2*-h2048 CNE_2*-h2048-std Recon_0 Recon_0-std Synth-nn Synth-nn-std Contrastive Contrastive-std Contrastive-MLP Contrastive-MLP-std Contrastive+Proj Contrastive+Proj-std Supervised Supervised-std PCA-whiten baseline
1	0.61189	0.00605871686085429	0.61088	0.00955569202098937	0.60517	0.00882266399677558	0.61483	0.00397288937676347	0.60649	0.0114024778008992	0.56570	0.00902616474478502	0.54050	0.00364159992311072	0.45910	0.000262773666869418	0.46939	0.00369802650071629	0.55565	0.00450249930594109	0.60784	0.00219009132229686	0.51161	nan
2	0.63746	0.00441692766524425	0.63553	0.00218563491919396	0.63459	0.00341052781838823	0.65665	0.00790154731682346	0.63149	0.00759530447052651	0.61791	0.00631245198001537	0.62038	0.000485231903320464	0.45962	0.000330151480384384	0.46285	0.000531507290636732	0.60362	0.00141823834386185	0.63233	0.00459497551680093	0.60591	nan
3	0.66752	0.00147229412822303	0.66508	0.00201853907566834	0.66252	0.00235992584629263	0.66127	0.00470551272445416	0.64674	0.00904390678855106	0.64379	0.00246483265152018	0.64271	0.000521775813927783	0.45886	0.00145500859103993	0.47064	0.00368917334913934	0.63321	0.00231101709210469	0.65832	0.00307942364737299	0.63529	nan
4	0.67836	0.003454287191303	0.67876	0.00245944099339667	0.67492	0.00295601082541996	0.66435	0.00666983133220024	0.65601	0.0054662647209955	0.63971	0.00180227633841206	0.65125	0.000286094390018399	0.46400	0.00252643028797551	0.45889	0.0115277274429959	0.64273	0.00186409495466299	0.67590	0.00200282300765694	0.64761	nan
5	0.68473	0.00198777513818842	0.68324	0.00201872484504451	0.68003	0.00142051047162631	0.65867	0.00409011002297004	0.65715	0.00910479543976689	0.64118	0.00102181211580212	0.65444	0.000540092584655631	0.46463	0.000360555127546399	0.46037	0.00188373299594183	0.64867	0.00156438486313311	0.67888	0.00508944987203922	0.65511	nan
6.2	nan	nan	nan	nan	nan	nan	nan	nan	nan	nan	nan	nan	nan	nan	nan	nan	nan	nan	nan	nan	nan	nan	nan	0.66191
}{\map}
    
    \addplot[cneplotlinear]      table[x=d,  y=CNE_0-h2048]   \map; \leg{\lcne}
    \addplot [name path=upper,draw=none, forget plot] table[x=d,y expr=\thisrow{CNE_0-h2048}+\thisrow{CNE_0-h2048-std}] {\map};
    \addplot [name path=lower,draw=none, forget plot] table[x=d,y expr=\thisrow{CNE_0-h2048}-\thisrow{CNE_0-h2048-std}] {\map};
    \addplot [fill=\cneplotlinearc!\stdgrad, forget plot] fill between[of=upper and lower];
    
    \addplot[cneplotflinearone]      table[x=d,  y=CNE_1-h2048]   \map; \leg{\flcne{1}}
    \addplot [name path=upper,draw=none, forget plot] table[x=d,y expr=\thisrow{CNE_1-h2048}+\thisrow{CNE_1-h2048-std}] {\map};
    \addplot [name path=lower,draw=none, forget plot] table[x=d,y expr=\thisrow{CNE_1-h2048}-\thisrow{CNE_1-h2048-std}] {\map};
    \addplot [fill=\cneplotflinearonec!\stdgrad, forget plot] fill between[of=upper and lower];
    
    \addplot[cneplotmlpone]      table[x=d,  y=CNE_1*-h2048]   \map; 
    \leg{\mlpcne{1}}
    \addplot [name path=upper,draw=none, forget plot] table[x=d,y expr=\thisrow{CNE_1*-h2048}+\thisrow{CNE_1*-h2048-std}] {\map};
    \addplot [name path=lower,draw=none, forget plot] table[x=d,y expr=\thisrow{CNE_1*-h2048}-\thisrow{CNE_1*-h2048-std}] {\map};
    \addplot [fill=\cneplotmlponec!\stdgrad, forget plot] fill between[of=upper and lower];

    \addplot[cneplotgaussian]      table[x=d,  y=Synth-nn]   \map; 
    \leg{\methodgauss}
    \addplot [name path=upper,draw=none, forget plot] table[x=d,y expr=\thisrow{Synth-nn}+\thisrow{Synth-nn-std}] {\map};
    \addplot [name path=lower,draw=none, forget plot] table[x=d,y expr=\thisrow{Synth-nn}-\thisrow{Synth-nn-std}] {\map};
    \addplot [fill=\cneplotgaussianc!\stdgrad, forget plot] fill between[of=upper and lower];
        
    \addplot[mseplot]      table[x=d,  y=Recon_0]   \map; 
    \leg{MSE}
    \addplot [name path=upper,draw=none, forget plot] table[x=d,y expr=\thisrow{Recon_0}+\thisrow{Recon_0-std}] {\map};
    \addplot [name path=lower,draw=none, forget plot] table[x=d,y expr=\thisrow{Recon_0}-\thisrow{Recon_0-std}] {\map};
    \addplot [fill=\mseplotc!\stdgrad, forget plot] fill between[of=upper and lower];
    
    \addplot[contrastiveplot]      table[x=d,  y=Contrastive+Proj]   \map; 
    \leg{\contrproj}
    \addplot [name path=upper,draw=none, forget plot] table[x=d,y expr=\thisrow{Contrastive+Proj}+\thisrow{Contrastive+Proj-std}] {\map};
    \addplot [name path=lower,draw=none, forget plot] table[x=d,y expr=\thisrow{Contrastive+Proj}-\thisrow{Contrastive+Proj-std}] {\map};
    \addplot [fill=\contrastiveplotc!\stdgrad, forget plot] fill between[of=upper and lower];

    \addplot[pcaplot]      table[x=d,  y=PCA-whiten]   \map; \leg{\pcaw}

    \addplot[baselineplot] table[x=d,  y=baseline]   \map; \leg{\gemap}

\end{axis}
\end{tikzpicture}
        % \caption{ROxford5K-Hard test set.}
        \label{fig:rparis6k_avg}
    \end{subfigure}
}

\caption{\textbf{Landmark image retrieval experiments using \gemap representations}. Mean average precision (mAP) on  \rox (left) and \rpa (right) as a function of the output dimensions $\dd$. We report \method with different encoders: \linear (\lcne), \flinear with 1 hidden layer
  (\flcne{1}), and a MLP with 1 hidden layer (\mlpcne{1}), the projector remains
  the same (MLP with 2 hidden layers). We compare with PCA with whitening, two baselines based on
  \lcne, but which respectively train with a reconstruction (MSE) and a
  contrastive (\contrproj) loss, and also with \textit{\methodgauss}, a variant of \method which uses Gaussian noise to synthesize pairs. The original \gemap performance is also reported. \label{fig:roxford5k_paris6k_avg}} 
\end{center}
\end{figure}

\mypartight{Results.}
Figure~\ref{fig:roxford5k_paris6k_avg} reports mean average precision (mAP) results
for \rox and \rpa; as the output
dimensions $\dd$ varies. We report the average of the Medium and Hard
protocols for brevity, while results per protocol are presented in 
% \arxivorsub{Appendix~\ref{sec:med_hard}}{the supplementary material}.
Appendix~\ref{sec:med_hard}.
We make a number of observations.
First and most importantly, we observe that both linear flavors of our approach
outperform PCA by a significant margin. For instance, \method improves ROxford5K retrieval by almost 4 mAP points for 128 dimensions over the PCA baseline. 
The MLP flavor is very competitive for very small dimensions (up to 128) but
degrades for larger ones. Even for the former, it is not worth the
extra-computational cost. 
An important observation is that we are able to retain the performance of the input representation (\gemap) while using only 1/16th of its dimensionality.
Using a different loss (MSE and \contrproj) instead of the Barlow Twins' in
\method degrades the results. These approaches are comparable to or worse than PCA.
Finally, replacing true neighbors by synthetic ones, as in \methodgauss, performs worse.  

% % -------------------------------------------------------------------------
% % Oxford + Paris results

% % -------------------------------------------------------------------------

%\ykt{

% ----------------------------------------------------
{\subsubsection{Landmark image retrieval using DINO representations \label{sec:landmark_dino}}}
% ----------------------------------------------------

% -------------------------------------------------------------------------
% DINO results
\begin{figure}
\begin{center}
% -------------------------------------------------------------------------
% Oxford results
% -------------------------------------------------------------------------
\resizebox{\linewidth}{!}{
    \begin{subfigure}{.5\linewidth}
        \centering
        \begin{tikzpicture}
\begin{axis}[%
  width=\linewidth,
  xlabel={$\dd$},
  xtick = {2,3,4,5,6,6.5},
  xticklabels = {16,32,64,128,256,384},
  ylabel={mAP},
  legend pos=south east,
  ylabel near ticks, xlabel near ticks, legend style={font=\scriptsize}, legend columns=2,
  height=5.5cm,
%   minor y tick num=4,
%   title={DINO \vit (\imnet pretrained)},
%   ytick = {0.1,0.15,0.2,0.25,0.29},
%   yticklabels = {0.1,0.15,0.2,0.25,0.29},
  minor y tick num=1,
  label style={font=\small},
  title style={font=\small}
  ]

\pgfplotstableread{
d tldr-r-im-gld pca-r-im-gld dino-r-im tldr-r-im-im pca-r-im-im
2 0.2519 0.1561 nan 0.1562	0.1149
3 0.2943 0.2399 nan 0.1975	0.1679
4 0.3087 0.2827 nan 0.2360	0.1870
5 0.3095 0.2822 nan 0.2532	0.2275
6 0.3086 0.3066 nan 0.2761	0.2699
6.5 nan nan 0.2555 nan nan
}{\map}

    \addplot[cneplotlinear]     table[x=d,  y=tldr-r-im-gld]   \map; \leg{\tldr (\gld)}
    \addplot[pcaplot]           table[x=d,  y=pca-r-im-gld]   \map; \leg{\pcaw (\gld)}
    \addplot[cneplotlinear, dashed]     table[x=d,  y=tldr-r-im-im]   \map; \leg{\tldr (ImNet)}
    \addplot[pcaplot, dashed]           table[x=d,  y=pca-r-im-im]   \map; \leg{\pcaw (ImNet)}

    \addplot[baselineplot]      table[x=d,  y=dino-r-im]   \map; %\label{dino_lbl} %\leg{DINO}
    
    % Second "Legend" node
% \node [draw,fill=white] at (rel axis cs: 0.88,0.4) {\scriptsize{\ref{dino_lbl} DINO}};
\node [] at (rel axis cs: 0.92,0.6) {\scriptsize{{DINO}}};
    
\end{axis}
\end{tikzpicture}
        % \caption{}
        % \caption{Results using DINO learned.}
        \label{fig:oxford_dino_vit}
    \end{subfigure}%
    \begin{subfigure}{.5\linewidth}
        \centering
        \begin{tikzpicture}
\begin{axis}[%
  width=\linewidth,
  xlabel={$\dd$},
  xtick = {2,3,4,5,6,6.7},
  xticklabels = {16,32,64,128,256,384},
  ylabel={mAP},
  legend pos=south east,
  ylabel near ticks, xlabel near ticks, 
  legend style={font=\small}, 
  height=5.5cm,
%   minor y tick num=4,
    minor y tick num=1,
  label style={font=\small},
  title style={font=\small}
  ]

\pgfplotstableread{
d tldr-r-gld-gld pca-r-gld-gld dino-r-gld tldr-fl
% 1 0.2068	0.0861 nan
2 0.2741	0.1513 nan 0.2822
3 0.3450	0.2613 nan 0.3456
4 0.3855	0.3276 nan 0.3843
5 0.4116	0.3837 nan 0.4087
6 0.4251	0.4065 nan 0.4165
% 6.7 0.4212	0.4098 0.3720
6.7 nan nan 0.3720 nan
}{\map}

    \addplot[cneplotlinear]     table[x=d,  y=tldr-r-gld-gld]   \map; \leg{\tldr }
    % \addplot[cneplotflinearone]     table[x=d,  y=tldr-fl]   \map; \leg{\flcne{1} (\gld)}
    \addplot[pcaplot]           table[x=d,  y=pca-r-gld-gld]   \map; \leg{\pcaw }
    \addplot[baselineplot]      table[x=d,  y=dino-r-gld]   \map; \leg{DINO } %\label{dinoleg} 

    % Second "Legend" node
    % \node [draw,fill=white] at (rel axis cs: 0.15,0.8) {\small{\ref{dinoleg} DINO}};

\end{axis}
\end{tikzpicture}
        % \caption{}
        % \input{plots/fig_oxford_dino_vit_gld}
        % \caption{$k$-NN Accuracy on \imnet using DINO.}
        \label{fig:oxford_dino_vit_gld}
    \end{subfigure}
}
\caption{\textbf{Self-supervised landmark retrieval performance on \rox using DINO \vit backbones}. Left: mAP on \rox as a function of the output dimensions $\dd$ using representations from DINO pretrained on \imnet; dimensionality reduction is learned on either \imnet (dashed lines) or \gld (solid lines).
Right: Performance for \tldr and PCA when learning dimensionality reduction on \gld over representations from a DINO model pretrained on \gld. No labels are used at any stage. 
\label{fig:dino_results}}
\end{center}
\end{figure}
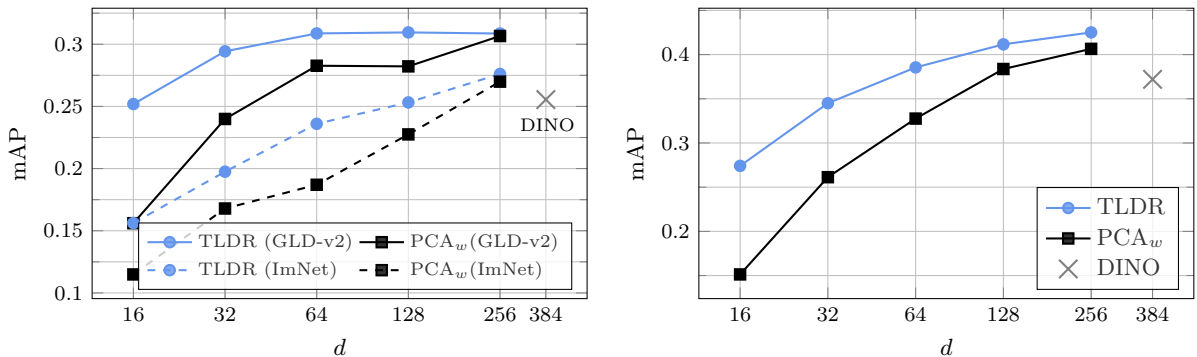

% -------------------------------------------------------------------------

In the previous section, we started from representations pretrained using supervision and tailored to the landmark retrieval task; we then learned dimensionality reduction on top of those in an unsupervised way. It is however interesting to also study the \textit{fully unsupervised case} and measure how well the method performs if we use representations 
%that are also 
learned in a self-supervised way. In this section, 
we assume that images are represented using
%present results using representations from 
DINO~\citep{caron2021emerging}, a state-of-the-art self-supervised approach that 
whose representations have lead to
%has shown 
impressive results for a wide variety of transfer learning tasks, including image retrieval. 
% We use multiple public pretrained DINO  models out-of-the-box as feature extractors.

\mypartight{Results.}
In Figure~\ref{fig:dino_results} we report results on \rox when learning dimensionality reduction on top of DINO features from a \vit backbone. In all cases, we follow the evaluation protocol presented in the previous section. In Figure~\ref{fig:dino_results} (left) we present results starting from the publicly available ViT DINO model trained on \imnet; similar to the \gemap case, we treat the ViT as a feature extractor and learn a linear encoder on top using either the \gld or \imnet datasets. In all cases, both features and dimensionality reduction are learned without any supervision. We see that \tldr shows strong gains over PCA with whitening, the best performing competing method, and that gains are consistent across multiple values of output dimension $d$ and across all setups. In fact, we see that if one assumes access to unlabeled data from the downstream landamark domain, one can achieve a large \textbf{+5.4 mAP gain} over DINO, \textbf{by simply learning a linear encoder on top}, and \textbf{without the need to fine-tune the ViT model}.  It is also noteworthy that \tldr is able to \textbf{match the DINO ViT performance} on \rox \textbf{using only 16-dimensions}. Results for \rpa are presented in the appendix.
% if one has access to unlabeled data from the downstream domain.

In Figure~\ref{fig:dino_results} (right) we start from a publicly available DINO model trained in an unsupervised way on \gld. \citet{caron2021emerging} evaluate their representations on \rox and \rpa using global image features from this model; they report 37.9\% mAP on \rox for the average setting (0.52/0.24 for medium/hard splits). We see that \tldr can improve on that result and is able to achieve \textbf{state-of-the-art} performance on \rox when using self-supervised learning, \ie 0.43 mAP (0.57/0.28 for medium/hard splits) for $d=256$, \textbf{+4.6\% mAP higher} than using the original DINO features learned on \gld.

% -------------------------------------------------------------------------
{\subsubsection{Results on ImageNet  \label{sec:results_imagenet_retrieval}}}
% -------------------------------------------------------------------------

% -------------------------------------------------------------------------
% imagenet and pq
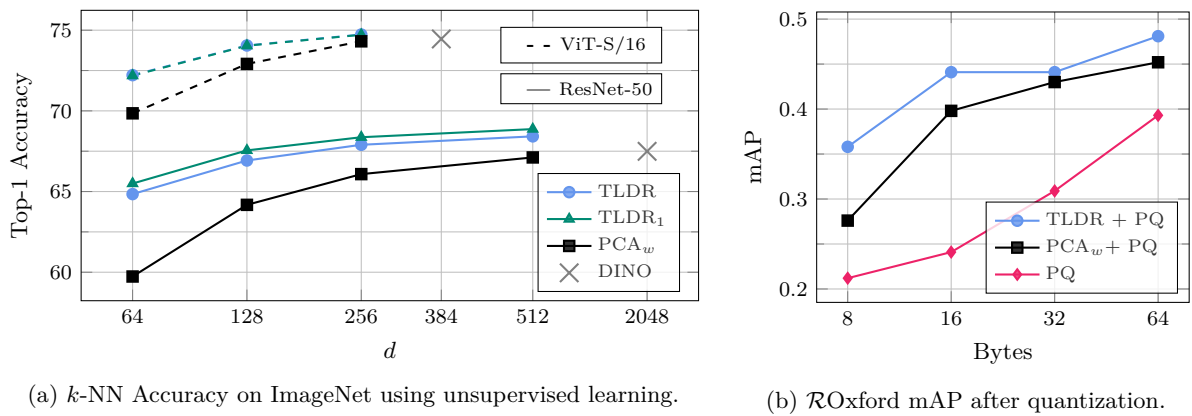
\begin{figure}
\begin{center}
% -------------------------------------------------------------------------
% Oxford results
% -------------------------------------------------------------------------
\resizebox{\linewidth}{!}{
    \begin{subfigure}{.6\linewidth}
        \centering
        \begin{tikzpicture}
\begin{axis}[%
  width=\linewidth,
  xlabel={$\dd$},
  xtick = {4,5,6,6.7,7.5,8.5},
  xticklabels = {64,128,256,384, 512,2048},
  ylabel={Top-1 Accuracy},
  legend pos=south east,
  ylabel near ticks, xlabel near ticks, 
  legend style={font=\scriptsize}, 
  height=5.5cm,
%   minor y tick num=4,
    minor y tick num=1,
  label style={font=\small},
  title style={font=\small}
  ]

\pgfplotstableread{
d tldr pca dino tldr-fl tldr-mlp
% 1 
% 2 
% 3 61.06 51.576 nan 61.59 60.364
4 64.842 59.734 nan 65.496 61.704
5 66.924 64.176 nan 67.55 62.634
6 67.898 66.08 nan 68.366 63.796
7.5 68.42 67.114 nan 68.87 64.21
8.5 nan nan 67.492 nan nan
% 6 
% 6.
}{\map}
\pgfplotstableread{
d pca tldr dino tldr-fl 
% 1 
% 2 
% 3 62.81 68.338 nan 68.43
4 69.848 72.214 nan 72.184
5 72.91 74.052 nan 74.038
6 74.31 74.732 nan 74.716
% 6.7 74.168 74.754 74.456 74.904
6.7 nan nan 74.456 nan
% 6 
% 6.
}{\mapvit}

    \addplot[cneplotlinear]     table[x=d,  y=tldr]   \map; \leg{\tldr}
    \addplot[cneplotflinearone]     table[x=d,  y=tldr-fl]   \map; \leg{\flcne{1}}
    \addplot[pcaplot]           table[x=d,  y=pca]   \map; \leg{\pcaw}
    \addplot[baselineplot]      table[x=d,  y=dino]   \map; \leg{DINO}

    \addplot[cneplotlinear, dashed]     table[x=d,  y=tldr]   \mapvit; %\leg{\tldr}
    \addplot[cneplotflinearone, dashed]     table[x=d,  y=tldr-fl]   \mapvit; %\leg{\flcne{1}}
    \addplot[pcaplot, dashed]           table[x=d,  y=pca]   \mapvit; %\leg{\pcaw}
    \addplot[baselineplot]      table[x=d,  y=dino]   \mapvit; %\label{dino} \leg{DINO}
    
    % \addplot[black,dashed,no marks] table[x=d1,  y=d2] \mapdummy; \label{dashed_line} 
        
    % Second "Legend" node
    \node [draw,fill=white,text width=55pt] at (rel axis cs: 0.81,0.73) {
    \scriptsize{~~\textbf{{---}}  \resnet}};

    \node [draw,fill=white,text width=55pt] at (rel axis cs: 0.81,0.88) {
    \scriptsize{~~\textbf{- -}  \vit}};

    % Second "Legend" node
    % \node [draw,fill=white] at (rel axis cs: 0.15,0.8) {\small{\ref{dino} DINO}};

\end{axis}
\end{tikzpicture}
        \caption{$k$-NN Accuracy on \imnet using unsupervised learning.}
        \label{fig:imagenet_knn}
    \end{subfigure}%
    \begin{subfigure}{.4\linewidth}
        \centering
        \begin{tikzpicture}
\begin{axis}[%
  width=\linewidth,
  xlabel={Bytes},
  xtick = {1,2,3,4},
  xticklabels = {8,16,32,64},
  ylabel={mAP},
  legend pos=south east,
  ylabel near ticks, xlabel near ticks, 
  legend style={font=\scriptsize}, 
height=5.5cm,
%   minor y tick num=4,
    minor y tick num=1,
  label style={font=\small},
  title style={font=\small}
  ]

\pgfplotstableread{
d tldr pca pcaw bestpca nopca
1 0.358 0.303 0.276 0.303 0.212
2 0.441 0.370 0.398 0.398 0.241
3 0.441 0.407 0.430 0.430 0.309
4 0.481 0.436 0.452 0.452 0.393

}{\map}

    \addplot[cneplotlinear]     table[x=d,  y=tldr]   \map; \leg{\tldr + PQ}
    % \addplot[mseplot]           table[x=d,  y=pca]   \map; \leg{PCA + PQ }
    \addplot[pcaplot]           table[x=d,  y=pcaw]   \map; \leg{\pcaw + PQ}
    % \addplot[contrastiveplot]           table[x=d,  y=bestpca]   \map; \leg{PCAb +PQ}
    \addplot[cneplotgaussian]           table[x=d,  y=nopca]   \map; \leg{PQ}
    
    % Second "Legend" node
    % \node [draw,fill=white] at (rel axis cs: 0.15,0.8) {\small{\ref{dino} DINO}};

\end{axis}
\end{tikzpicture}
        \caption{\rox mAP after quantization.}
        \label{fig:pq_study}
    \end{subfigure}
}
\caption{\textbf{Retrieval results on \imnet, and after vector quantization. } Left: Top-1  Accuracy as a function of the output dimensions $\dd$ for $k$-NN retrieval on \imnet following the protocol from~\cite{caron2021emerging} and using DINO \resnet and ViT representations trained on \imnet. Right: Performance on \rox after PQ quantization of the reduced features ($d=128$), as a function of the output vector size. \label{fig:imagenet_and_pq}}
\end{center}
\end{figure}

% -------------------------------------------------------------------------

We further evaluate the performance of the proposed \tldr on \imnet retrieval using $k$-NN. We follow the protocol from~\citet{caron2021emerging, wu2018unsupervised} and run the corresponding evaluation scripts provided in the DINO codebase. We query with all images in the \imnet val set, using the training set as the database; similar to DINO, we then report results using 20-NN. We report Top-1 Accuracy in Figure~\ref{fig:imagenet_knn} for the DINO \resnet and \vit models. We see that \tldr is consistently better than PCA also in this case, and for \resnet gives +9.4/5.1/2.7\% gains over PCA for $d$=32/64/128. For ViT, performance is higher and gains smaller, but still consistent. Generally, linear \tldr encoders (linear and factorized linear) \textbf{improve DINO's retrieval performance} on \imnet $k$-NN for both backbones and for all output dimension over $d=128$; this is not the case for PCA, which is not able to improve DINO's performance. We further see that \tldr is able to outperform the original 2048-dimensional features with only 256 dimensions, \ie \textbf{achieving a $10\times$ compression without loss in performance} for \resnet, or reach approximately 75\% Top-1 Accuracy on \imnet with 256-dimensional features for \vit DINO.
% For $d=512$ the gain becomes close to \gain{1\%} mAP.

% these results are super cool! On the right we can both claim to be able to outperform VIT on the same amount of dims and also outperform the resnet50 with just 32 dims

%}

% -------------------------------------------------------------------------
{\subsection{Results on first stage document retrieval \label{sec:results_nlp}}}
% -------------------------------------------------------------------------

For document retrieval, the process is generally divided into two stages: the first one selects a small set of candidates while the second one re-ranks them. Because it works on a smaller set, this second stage can afford costly strategies, but the first stage has to scale. The typical way to do this is to reduce the dimension of the representations used in the first retrieval stage, often in a supervised fashion~\citep{khattab2020colbert,gao2021coil}. Following our initial motivation, we investigate the use of unsupervised dimensionality reduction for document retrieval scenarios where 
a supervised approach is not possible.
% , \eg when no such training data is available.   

\mypartight{Experimental protocol.}
We start from 768-dimensional features extracted from a model trained for Question Answering (QA), \ie ANCE~\citep{xiong2020approximate}. %\dl{Can we please define QA. "Question Answering"?}
We use Webis-Touché-2020~\citep{bondarenko2020overview, wachsmuth2017building} a conversational argument dataset composed of 380k documents to learn the dimensionality reduction function. 
% /=\onlyonarxiv{
\mypartight{Compared approaches.}
% }
We report results for three flavors of our approach. \textit{\method} uses a
linear %projector 
encoder while \textit{\flcne{1}} and \textit{\flcne{2}} use a factorized linear one with respectively one hidden layer and two hidden layers. We compare with PCA, which was the best performing competitor from Section~\ref{sec:results_image_retrieval}.
We also report retrieval results obtained with the 768-dimensional initial features.

\begin{figure}
\begin{center}
\resizebox{\linewidth}{!}{
    \begin{subfigure}{.47\linewidth}
        \centering
        \begin{tikzpicture}
\begin{axis}[%
  width=\linewidth,
  height=5cm,
  xlabel={$\dd$},
  xtick = {1,2,3,4,5,7.5},
  xticklabels = {8,16,32,64,128,768},
  ylabel={Recall@100},
  ylabel near ticks,
  xlabel near ticks,
  minor y tick num=3,
  legend pos=south east,
   legend style={font=\footnotesize},
  every tick label/.append style={font=\footnotesize}
  ]

\pgfplotstableread{
    d cnelinear cnelinear-std cnelinearone cnelinearone-std cnelineartwo cnelineartwo-std pca baseline
    1 73.31 1.17 76.89 0.61 76.84 1.37 50.14 nan
    2 86.67 0.72 88.98 0.46 88.68 0.89 73.83 nan
    3 92.50 0.19 93.97 0.39 94.01 0.19 87.84 nan
    4 94.45 0.28 95.68 0.18 95.68 0.19 93.67 nan
    5 94.56 0.27 96.26 0.24 96.22 0.15 95.66 nan
    7.5 nan nan nan nan nan nan nan 94.02
    }{\map}

   \addplot[cneplotlinear]      table[x=d,  y=cnelinear]   \map; \leg{\lcne}
    \addplot [name path=upper,draw=none, forget plot] table[x=d,y expr=\thisrow{cnelinear}+\thisrow{cnelinear-std}] {\map};
    \addplot [name path=lower,draw=none, forget plot] table[x=d,y expr=\thisrow{cnelinear}-\thisrow{cnelinear-std}] {\map};
    \addplot [fill=\cneplotlinearc!\stdgrad, forget plot] fill between[of=upper and lower];   
   
   \addplot[cneplotflinearone]      table[x=d,  y=cnelinearone]   \map; \leg{\flcne{1}}
    \addplot [name path=upper,draw=none, forget plot] table[x=d,y expr=\thisrow{cnelinearone}+\thisrow{cnelinearone-std}] {\map};
    \addplot [name path=lower,draw=none, forget plot] table[x=d,y expr=\thisrow{cnelinearone}-\thisrow{cnelinearone-std}] {\map};
    \addplot [fill=\cneplotflinearonec!\stdgrad, forget plot] fill between[of=upper and lower];

   \addplot[cneplotflineartwo]      table[x=d,  y=cnelineartwo]   \map; \leg{\flcne{2}}
    \addplot [name path=upper,draw=none, forget plot] table[x=d,y expr=\thisrow{cnelineartwo}+\thisrow{cnelineartwo-std}] {\map};
    \addplot [name path=lower,draw=none, forget plot] table[x=d,y expr=\thisrow{cnelineartwo}-\thisrow{cnelineartwo-std}] {\map};
    \addplot [fill=\cneplotflineartwoc!\stdgrad, forget plot] fill between[of=upper and lower];
   
   \addplot[pcaplot]      table[x=d,  y=pca]   \map; \leg{\pcaw}
   \addplot[baselineplot] table[x=d,  y=baseline]   \map; \leg{ANCE}

\end{axis}
\end{tikzpicture}
        % \caption{Recall@100}
        \label{fig:arguana_recall_nl}
    \end{subfigure}
    \begin{subfigure}{.47\linewidth}
        \centering
        \begin{tikzpicture}
\begin{axis}[%
  width=\linewidth,
  xlabel={$\dd$},
  height=5cm,
  xtick = {1,2,3,4,5,7.5},
  xticklabels = {8,16,32,64,128,768},
  ylabel={Recall@100},
  ylabel near ticks,
  xlabel near ticks,
    minor y tick num=3,
   legend style={font=\footnotesize},
  legend pos=south east,
  every tick label/.append style={font=\footnotesize}
  ]

\pgfplotstableread{
    d cnelinear cnelinear-std cnelinearone cnelinearone-std cnelineartwo cnelineartwo-std pca baseline
    1 76.84 1.37 76.40 1.13 73.36 1.38 50.14 nan
    2 88.68 0.89 88.85 0.55 88.83 0.41 73.83 nan
    3 94.01 0.19 93.95 0.31 93.27 0.30 87.84 nan
    4 95.68 0.19 95.63 0.11 95.35 0.19 93.67 nan
    5 96.22 0.15 96.24 0.18 95.86 0.14 95.66 nan
    7.5 nan nan nan nan nan nan nan 94.02
    }{\map}

   \addplot[cneplotflineartwo]      table[x=d,  y=cnelinear]   \map; \leg{\flcne{2,k=3}}
    \addplot [name path=upper,draw=none, forget plot] table[x=d,y expr=\thisrow{cnelinear}+\thisrow{cnelinear-std}] {\map};
    \addplot [name path=lower,draw=none, forget plot] table[x=d,y expr=\thisrow{cnelinear}-\thisrow{cnelinear-std}] {\map};
    \addplot [fill=\cneplotflineartwoc!\stdgrad, forget plot] fill between[of=upper and lower];   
   
   \addplot[cneplotflineartwokt]      table[x=d,  y=cnelinearone]   \map; \leg{\flcne{2,k=10}}
    \addplot [name path=upper,draw=none, forget plot] table[x=d,y expr=\thisrow{cnelinearone}+\thisrow{cnelinearone-std}] {\map};
    \addplot [name path=lower,draw=none, forget plot] table[x=d,y expr=\thisrow{cnelinearone}-\thisrow{cnelinearone-std}] {\map};
    \addplot [fill=\cneplotflineartwoktc!\stdgrad, forget plot] fill between[of=upper and lower];

   \addplot[cneplotflineartwokh]      table[x=d,  y=cnelineartwo]   \map; \leg{\flcne{2,k=100}}
    \addplot [name path=upper,draw=none, forget plot] table[x=d,y expr=\thisrow{cnelineartwo}+\thisrow{cnelineartwo-std}] {\map};
    \addplot [name path=lower,draw=none, forget plot] table[x=d,y expr=\thisrow{cnelineartwo}-\thisrow{cnelineartwo-std}] {\map};
    \addplot [fill=\cneplotflineartwokhc!\stdgrad, forget plot] fill between[of=upper and lower];
   
   \addplot[pcaplot]      table[x=d,  y=pca]   \map; \leg{\pcaw}
   \addplot[baselineplot] table[x=d,  y=baseline]   \map; \leg{ANCE}

\end{axis}
\end{tikzpicture}
        % \caption{Recall@100}
        \label{fig:arguana_recall_k}
    \end{subfigure}
}
% \vspace{-4pt}
\caption{\textbf{Argument retrieval results on ArguAna} for different values of output dimensions $\dd$. On the left we vary the amount of factorized layers, with fixed $k=3$, on the right we fix the amount of factorized layers to 2 and test $k=[3,10,100]$. Factorized linear is fixed to 512 hidden dimensions.}
\label{fig:arguana}
\end{center}
\end{figure}
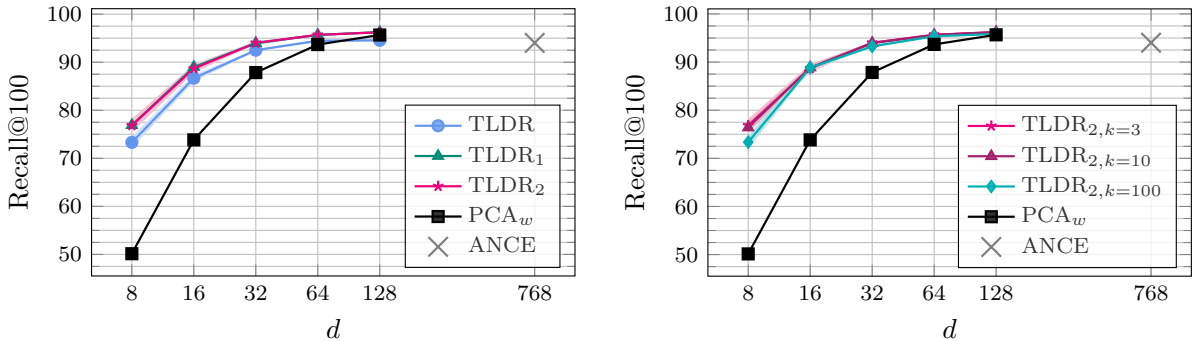

\mypartight{Results.} Figure \ref{fig:arguana} reports retrieval results on ArguAna, for different output dimensions $\dd$. We observe that the linear version of \method outperforms PCA for almost all values of $\dd$. The linear-factorized ones outperforms PCA in all scenarios. We see that the gain brought by \method over PCA increases as $\dd$ decreases. Note that we achieve results equivalent to the initial ANCE representation using only $4\%$ of the original dimensions; PCA, needs twice as many dimensions to achieve similar performance.

% -------------------------------------------------------------------------
% ArguAna results
% -------------------------------------------------------------------------
%------------------------------------------------------------------------------

% -------------------------------------------------------------------------
{\subsection{Analysis and Impact of hyper-parameters \label{sec:results_analysis}}}
% -------------------------------------------------------------------------

\mypartight{Impact of hyper-parameters.}
We observed surprising robustness for \tldr across multiple hyper-parameters. For brevity, we merely summarize the main findings here, while complete figures are presented in the appendix. 
When we vary the \textbf{architecture of the projector} $\pg$, an important module of \tldr, we see that having hidden layers generally helps. As also noted in~\citet{zbontar2021barlow}, having a high auxiliary dimension $\ddd$ for computing the loss is very important and highly impacts performance. When it comes to the \textbf{numbers of neighbors} $k$, we see that \tldr is surprisingly consistent across a wide range of $k$. We observe the same stability across several \textbf{batch sizes}. More details are presented in the appendix.
% and results varying all aforementioned parameters can be found in the Appendix Sections~\ref{sec:varyingprojectorandk},~\ref{sec:varyingbatchsize}.

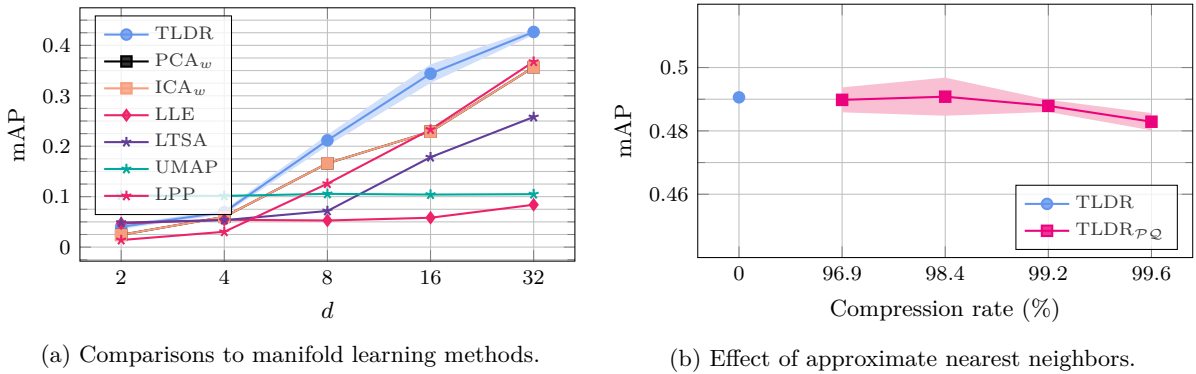
\begin{figure}
\begin{center}
\resizebox{\linewidth}{!}{
    \begin{subfigure}{.5\linewidth}
        \centering
        \begin{tikzpicture}
\begin{axis}[%
  width=\linewidth,
  xlabel={$\dd$},
  xtick = {1,2,3,4,5},
  xticklabels = {2,4,8,16,32},
  ylabel={mAP},
  legend pos=north west,
  ylabel near ticks, xlabel near ticks, legend style={font=\scriptsize},
  height=5cm,
  minor y tick num=3,
  label style={font=\small},
  title style={font=\small}
  ]

\pgfplotstableread{
d CNE_0-h4096 CNE_0-h4096-std PCA-whiten ICA-whiten UMAP LLE LTSA LPP
1	0.04032	0.00455821236890077	0.0242947	0.02429225	0.1022422	0.0469755	0.048558 0.01401
2	0.06936	0.00431662483892219	0.06000055	0.06000055	0.10152985	0.05418	0.0534965 0.03036
3	0.21170	0.0109935663003413	0.1660485	0.1660526	0.10557445	0.052789	0.071597 0.1257
4	0.34387	0.0184670097741892	0.22884035	0.228841	0.10412845	0.058211	0.178217 0.2331
5	0.42663	0.00596566844536302	0.356696	0.35670	0.10496875	0.08395	0.258008 0.3677
% 6	0.45228	0.00502573377727074	0.43195	0.43305	0.10496875	0.1072085	0.35336715 0.44645
% 7	0.48798	0.00520508405311576	0.44950	0.45470	0.10496875	0.167608	0.402092 0.47194
}{\map}
    
    \addplot[cneplotlinear]      table[x=d,  y=CNE_0-h4096]   \map; \leg{\lcne}
    \addplot [name path=upper,draw=none, forget plot] table[x=d,y expr=\thisrow{CNE_0-h4096}+\thisrow{CNE_0-h4096-std}] {\map};
    \addplot [name path=lower,draw=none, forget plot] table[x=d,y expr=\thisrow{CNE_0-h4096}-\thisrow{CNE_0-h4096-std}] {\map};
    \addplot [fill=\cneplotlinearc!\stdgrad, forget plot] fill between[of=upper and lower];

    \addplot[pcaplot]      table[x=d,  y=PCA-whiten]   \map; \leg{\pcaw}
    
    \addplot[drlimplot]      table[x=d,  y=ICA-whiten]   \map; \leg{\icaw}
    
    \addplot[cneplotgaussian]      table[x=d,  y=LLE]   \map; \leg{LLE}
    
    \addplot[mseplot]      table[x=d,  y=LTSA]   \map; \leg{LTSA}
    
    \addplot[contrastiveplot] table[x=d,  y=UMAP]   \map; \leg{UMAP}
    
    \addplot[lppplot] table[x=d,  y=LPP]   \map; \leg{LPP}

\end{axis}
\end{tikzpicture}
        \caption{Comparisons to manifold learning methods.}
        \label{fig:roxford5k_umap}
    \end{subfigure}%
    \begin{subfigure}{.5\linewidth}
        \centering
        \begin{tikzpicture}
\begin{axis}[%
  width=\linewidth,
  xlabel={Compression rate (\%)},
  xtick = {5,4,3,2,1},
  xticklabels = {99.6,99.2,98.4,96.9,0},
  ylabel={mAP},
  ytick = {0.46,0.48,0.50},
  legend pos=south east,
  ylabel near ticks, xlabel near ticks, legend style={font=\scriptsize},
  height=5cm,
  minor y tick num=1,
  label style={font=\small},
  title style={font=\small},
  ymax=0.52, ymin=0.44
  ]

\pgfplotstableread{
d CNE_0-h2048 CNE_0-h2048-std PQCNE_0-h2048 PQCNE_0-h2048-std
1 0.49060 0.0086457648591666 nan nan
2 nan nan 0.4898 0.0039406217783492
3 nan nan 0.4908 0.00601040764008565
4 nan nan 0.4879 0.0019173549488814
5 nan nan 0.4829 0.002725371534305
}{\map}
    
    \addplot[cneplotlinear]      table[x=d,  y=CNE_0-h2048]   \map; \leg{\lcne}
    \addplot [name path=upper,draw=none, forget plot] table[x=d,y expr=\thisrow{CNE_0-h2048}+\thisrow{CNE_0-h2048-std}] {\map};
    \addplot [name path=lower,draw=none, forget plot] table[x=d,y expr=\thisrow{CNE_0-h2048}-\thisrow{CNE_0-h2048-std}] {\map};
    \addplot [fill=\cneplotlinearc!\stdgrad, forget plot] fill between[of=upper and lower];

    \addplot[cneplotlinearpq]      table[x=d,  y=PQCNE_0-h2048]   \map; \leg{\ltldrpq}
    \addplot [name path=upper,draw=none, forget plot] table[x=d,y expr=\thisrow{PQCNE_0-h2048}+\thisrow{PQCNE_0-h2048-std}] {\map};
    \addplot [name path=lower,draw=none, forget plot] table[x=d,y expr=\thisrow{PQCNE_0-h2048}-\thisrow{PQCNE_0-h2048-std}] {\map};
    \addplot [fill=\cneplotlinearpqc!\stdgrad, forget plot] fill between[of=upper and lower];

\end{axis}
\end{tikzpicture}
        %\caption{ROxford5K-Hard test set.}
        \caption{Effect of approximate nearest neighbors.}
        \label{fig:roxford5k_avg_ann}
    \end{subfigure}
}
\caption{Left: Comparisons to manifold learning methods for small output dimensions $\dd \le 128$; we report Mean average precision (mAP) on \rox using \gemap representations as a function of the output dimensions $\dd$. 
Right: The effect of nearest neighbor approximation for $d=128$. We plot mAP as a function of the compression rate used during nearest neighbor computation. Note that the baseline (compression rate = 0) is using the 2048-dimensional (8192 bytes) \gemap  representations during nearest neighbor computation.}
\label{fig:roxford5k_small_d_and_ann}
\end{center}
\end{figure}

\mypartight{Comparisons to manifold learning methods on smaller output dimensions.}
In Figure~\ref{fig:roxford5k_umap} we present results for TLDR when the output dimensionality is $\dd \le 32$; in this regime, a few more manifold learning methods can be run, \eg ICA, Locally Linear Embedding (LLE)~\citep{roweis2000nonlinear}, Local Tangent Space Alignment (LTSA)~\citep{zhang2004principal}, Locality Preserving Projections (LPP)~\citep{he2003locality} and UMAP~\citep{mcinnes2018umap}. Unfortunately, even at smaller output dimensions we had to subsample the dataset to run some of the methods, due to their scalability issues. Specifically, we are forced to use only 5\% of the training set ($\sim$75K images) for learning LLE and LTSA, 10\% for LPP ($\sim$150K - this setting gives the best results) and 50\% ($\sim$750K images) for UMAP. We see that \tldr generally dominates for $d \ge 8$, and that linear methods tend to be the highest performing ones as the output dimensionality grows. 

\mypartight{How sensitive are \tldr-reduced vectors to subsequent quantization?}
In Figure~\ref{fig:pq_study} we verify that the gains from \tldr persist after compressing the dimensionality reduced vectors to only a few bytes, \eg via Product Quantization (PQ)~\citep{jegou2010product}. We conduct the study on \rox, starting from the \gemap \resnet representations. First of all, we see that it is highly desirable to first use dimensionality reduction and then PQ-compress the vectors; starting from the 2048-dimensional vectors results to much lower performance. Secondly, we see that \tldr once again highly outperforms PCA, with +8.16\% and +4.28\% in mAP for compression to 64 or 128 \textit{bits}, respectively.

\mypartight{How sensitive is \tldr to approximate nearest neighbors?}
To verify that our system is robust to an approximate computation of nearest neighbors, we test its performance using product quantization~\citep{jegou2010product,ge2013optimized} while varying the quantization budget (i.e. the amount of bytes used for each image during the nearest neighbor search). Compression is done using optimized product quantization (OPQ)~\citep{ge2013optimized} via the FAISS library~\citep{faiss} and results are reported in Figure~\ref{fig:roxford5k_avg_ann}. We present results on \rox, starting from the \gemap \resnet representations. We see that \tldr is quite robust to quantization during the nearest neighbor search and that even when the quantization is pretty strong 
(1/64 the default size or merely 16 bytes per vector) 
\tldr still retains its high performance.

% -------------------------------------------------------------------------
% More baselines
% -------------------------------------------------------------------------
% -------------------------------------------------------------------------
% \arxivorsub{
% \section{Discussion and related work \label{sec:discussion}}

\section{Related work}
\label{sec:discussion}

The basic idea behind \tldr is embarrassingly simple and links to a large number of related methods, from PCA to manifold learning and neighborhood embedding.
In this section we discuss a few such relations; more are discussed in Appendix~\ref{appendix:discussion}.
% due to lack of space.
% }
% {
% \section{Discussion} \label{sec:discussion}
% }

\mypartight{Linear dimensionality reduction.} 
We refer the reader to~\citet{cunningham2015linear} for an extensive review of linear dimensionality reduction. It is beyond the scope of this paper to exhaustively discuss many such related works, we will therefore focus on PCA~\citep{pca} which is the de facto standard linear dimensionality method, in particular for large-scale  retrieval. 

One can derive the learning objective of PCA~\citep{pca} by setting $\ef(x) = W^Tx$ and $\proj(x) = Wx$ in the model of Figure~\ref{fig:overview}, \ie use a linear encoder and projector with $W \in \mathrm{R}^{\di \times \dd}$, and optimize $W$ via minimizing the Frobernius norm of the matrix of reconstruction errors over the whole training set, subject to orthogonality constraints:
\begin{equation}
\label{eq:pca}
    W^* = \arg \min_W ||x - \proj(\ef(x))||_F, \quad  \mathrm{s.t.} \quad W^TW = I_\dd.
\end{equation}
This equation has a closed form solution that can be obtained via the eigendecomposition of the data covariance matrix and then keeping the largest $\dd$ eigenvectors\footnote{We refer the reader to Chapter 2 of the Deep Learning book~\citep{goodfellow2016deep} for the derivation of the PCA solution.}. 

Unlike PCA, \tldr does not constrain the projector to be a linear model, nor the loss to be a reconstruction loss. 
In fact, the redundancy reduction term in the Barlow Twins loss encourages the \textit{whitening} of the batch representations as a soft constraint~\citep{zbontar2021barlow}, in a way analogous to the orthogonality constraint of Eq.(\ref{eq:pca}).
We see from Figure~\ref{fig:roxford5k_paris6k_ablative} that part of the performance gains of \tldr over PCA is precisely due to this asymmetry in the architecture, \ie when the projector is an MLP with hidden layers. Looking at MSE results in Figures~\ref{fig:roxford5k_paris6k_avg}, \ie a version of \tldr with a reconstruction loss, we also see that the Barlow Twins loss and the flexibility of computing it in an arbitrarily high $\ddd$-dimensional space further contributes to this gain. One can therefore interpret \tldr \emph{as a more generic way of optimizing a linear encoder, \ie using an arbitrary decoder and approximating the constraint of Eq.(\ref{eq:pca}) in a soft way,} further incorporating a weak notion of whitening. 

Another way of learning a linear encoder is presented in~\citet{he2003locality} where Locality Preserving Projections are given by the optimal linear approximations to the eigenfunctions of the Laplace-Beltrami operator on the data manifold. We find the \tldr optimization to not only be more scalable, but also more versatile as it is stochastically and independently optimizing for different neighborhoods of the manifold.

% (see Figure~\ref{fig:roxford5k_umap}); we believe that  a linear encoder with
%  able to focus on more \textit{local} manifold structure, stochastic and flexible, leading to a better solution to approximating the manifold structure
% LPP lacks in scalability--we had to subsample to compare in Figure~\ref{fig:roxford5k_umap}; we also believe the randomness and stochasicity of learning a linear encoder with }
% Although resulting in a linear encoder, LPP's optimization requires computing a generalized eigenvector problem relying on computing laplacians on the data adjaceny graph, }

% \onlyonarxiv{
\mypartight{Manifold learning and neighborhood embedding methods.} 
Manifold learning methods define objective functions that try to preserve the local structure of the input manifold, usually expressed via a \knn graph. 
Non-linear unsupervised dimensionality-reduction methods usually require the \knn graph of the input data, while most further require eigenvalue decompositions~\citep{roweis2000nonlinear, donoho2003hessian, zhang2004principal} and shortest-path~\citep{tenenbaum2000global} or computation of the graph Laplacian~\citep{belkin2003laplacian}. Others involve more complex optimization~\citep{mcinnes2018umap, agrawal2021minimum}. 
Moreover, many manifold learning methods were created to solely operate on the data they were learned on. Although ``out-of-sample'' extensions for many of such methods have been proposed~\citep{bengio2004out}, methods like Spectral Embeddings, pyMDE~\citep{agrawal2021minimum} or the very popular \tsne~\citep{van2008visualizing} can only be used for the data they were trained on. Finally, UMAP~\citep{mcinnes2018umap} was recently proposed as not only a competitor of \tsne on 2-dimensional outputs, but as a general purpose dimension reduction technique. Yet, all our experiments with UMAP, even after exhaustive hyperparameter tuning, resulted in very low performance for  $\dd \ge 8$ for all the tasks we evaluated. % in the main paper and the Appendix. 

\mypartight{Nearest neighbors as ``supervision'' for contrastive learning.}
The seminal method DrLIM~\citep{hadsell2006dimensionality} uses a contrastive loss over neighborhood pairs for representation learning. Experimenting only on simple datasets like MNIST, it learns a CNN backbone and the dimensionality reduction function in a single stage, using a max-margin loss. \tldr resembles DrLIM~\citep{hadsell2006dimensionality} with respect to the encoder input and the way pairs are constructed; a crucial difference, however, is the loss function and the space in which it is computed: DrLIM uses a contrastive loss which is computed directly on the lower dimensional space. Despite our best effort to make this approach work as described, performance was very low without a projector. Using the contrastive loss from~\citet{hadsell2006dimensionality}, together with the projector we use for \method, we were able to get more meaningful results (reported as \contrproj in our experiments), although still underperforming the Barlow Twins loss. This difference may be due to two reasons: first, and as discussed above, Barlow Twins encourages the whitening of the representations which makes it more suitable for this task. Second, and like many other pair-wise losses, the contrastive loss further requires sampling hard/meaningful negatives~\citep{wu2017sampling,radenovic2018fine}.
Very recently, \citet{iscen2022learning} use features space neighbor consistency as a regularization term for learning under label noise by encouraging the prediction of each example to be similar to its nearest neighbours'.

\looseness=-1
\mypartight{Relation to Deep Canonical Correlation Analysis.} 
The MLP variant of \tldr is closely related to the Deep Canonical Correlation Analysis (DCCA) introduced by~\citet{andrew2013deep}. The Barlow Twins loss can be seen as a more modern, simpler way to optimize DCCA. First, the addition of the projector is shown to 
significantly and positively impact the results.
%empirically make a big difference, 
Second, the gradient of DCCA involves computing the gradient of the trace of the square-root of a matrix, a step which is not required for computing the Barlow Twins loss.
%while deriving the gradient of DCCA involves for example the gradient of the trace of the matrix square-root, something that is not needed for computing the Barlow Twins loss. 
Moreover, Deep CCA obtains optimal results with
%showed their best results  using 
full-batch optimization and L-BFGS, \ie a far less scalable setting. 

\mypartight{Temporal neighborhoods.} 
Component analysis methods that take
%Also related are component analysis methods for that take 
into account \dlt{the} temporal structure of the data are also related, something critical for time series.
Dynamical Components Analysis (DCA)~\citep{bai2020representation}, for example, uses \textit{temporal} neighborhoods, to discover a subspace with maximal predictive information, defined as the mutual information between the past and future. Similarly, Deep Autoencoding Predictive Components (DAPC)~\citep{clark2019unsupervised} builds on the same notion of predictive mutual information from temporal neighbors, but further adopts a masked reconstruction task to enforce the latent representations to be more informative.

\mypartight{Graph diffusion for harder pairs.} \citet{iscen2018mining} improve the method from~\citet{hadsell2006dimensionality} by mining harder positives and negative pairs for the contrastive loss via diffusion over the \knn graph. Similar to~\citet{hadsell2006dimensionality}, they learn (fine-tune) the whole network and not just the dimensionality-reduction layer. Although it would be interesting to incorporate such ideas in \tldr, we consider it complementary and beyond the scope of this paper. 
Methods for learning descriptor matching are also related; \eg \citep{simonyan2014learning} formulates dimensionality reduction as a convex optimisation problem. Although the redundancy reduction objective can be formulated in many ways, \eg via  stochastic proximal gradient methods like Regularised Dual Averaging in~\citet{simonyan2014learning}, we believe that the \emph{simplicity, immediacy and clarity} in which the Barlow Twins objective optimizes the output space is a strong advantage of \method.
% }

\section{Discussion}

% When operating on high-dimensional vector spaces, it is a standard assumption shared by all manifold learning methods and therefore also by \tldr that there exists a locally linear/euclidean manifold in the input space that we want to preserve.
% Augmentations like the ones used in visual representation learning are based on the fact that we know that the source signal is pixels and therefore \eg visual semantics should be preserved after shifting, spatial cropping, or color jittering. In our case, we start from vectors in high-dimensional spaces without further assumptions, i.e. we assume that the input space is robust and invariant to common perturbations like the ones mentioned above for images. 

% Paragraph discussing BYOL, SimSiam variants [FuBN, hxnp] and the possible use case of 
\mypartight{The Barlow Twins and related SSL losses.} 
We chose the Barlow Twins loss from~\citet{zbontar2021barlow} as we believe it fits dimensionality reduction exceptionally. The second term of the loss does not only provide Barlow Twins with an elegant way of avoiding collapse, something not trivial for other related losses like SimSiam~\citep{chen2020exploring,zhang2022does}, but decorrelating the output space is a property that highly suits the dimensionality reduction task. Moreover, from our experiments and ablations, we see that the Barlow Twins loss is highly robust to most hyper-parameters. 
We could however devise a variant of \tldr that uses any other ``symmetrical'' loss, \eg SimCLR~\citep{chen2020simple}, SimSiam~\citep{chen2020exploring} or the recently proposed VICReg~\citep{bardes2021vicreg} that builds on Barlow Twins. Yet, leveraging SSL losses that use self-distillation and the Exponential Moving Average trick such as BYOL~\citep{grill2020bootstrap} or MoCo~\citep{he2020momentum} with neighbor pairs might be far from trivial. 
% It is worth noting, however, that the first term of the Barlow Twins loss is very similar to the non-contrastive loss from BYOL.

% TLDR as a signal for distillation [hxnp]
\looseness=-1
\mypartight{\tldr beyond dimensionality reduction.} 
As we discuss in the previous paragraph, 
%we do not consider neighbors to be a replacement for image transformations for learning large visual representation models from scratch.
we do not consider \tldr to be a method for learning large visual representation models from scratch, where neighbors would simply replace image transformations.
We however believe that this signal can be useful as another way of performing model distillation, by \eg sampling neighbors from a larger model or a model trained on more data. It is also clear from the results of \tldr learning on top of the generic \imnet-trained DINO and the \gld dataset in Figure~\ref{fig:dino_results} (left), that our method can be used for unsupervised domain adaptation.
\tldr is also agnostic to the way neighbors are extracted, and other similarity measures could be used. The choice of this measure would then be a hyper-parameter, potentially taking prior information into account.
%and which distance to used for creating neighbor pairs would become a hyper-parameter, for instance to take into account prior information. 
\tldr could then learn an (output) embedding space for which the Euclidean distance is suitable, a highly desirable property for many tasks.
%an embedding (output) space for which the Euclidean distance is to be used, something really desirable because of its simplicity.

% Paragraph on computational complexity, clarifying “scalability” and training times [FuBN, j6W2, hxnp]
\mypartight{A scalable manifold learning method.} 
By ``scalability'' we refer to the ability of the proposed \tldr to be efficiently applied to arbitrary large datasets and output spaces. We believe this to be an important property, given that most manifold learning methods are unable to ``scale'' to large datasets or output dimensions.
% : many require constructing and navigating large graphs, computing eigendecompositions of large laplacians or other optimization processes that scale super-linearly with either the output dimension or dataset size. 
This is why, in Figure~\ref{fig:roxford5k_umap} for example, we had to subsample the training dataset %to make some of the competing methods 
so some of the competing methods could
fit in memory and/or converge within a couple of days (see also Appendix~\ref{sec:varyingtrainingsizetime} for training time comparisons). \tldr uses mini-batch Stochastic Gradient Descent and has linear time and memory complexity with respect to both the training database size and the output dimension.
In that regard, this is a highly scalable manifold learning method.
%and is a highly scalable manifold learning method in that regard. 

\mypartight{Computational cost and training time.}
In terms of computation, it is worth noting that 
training \tldr, \ie learning the dimensionality reduction encoder,
%learning a dimensionality reduction encoder for \tldr 
is generally orders of magnitude less expensive than than \eg learning visual representation. We always start from pre-extracted feature vectors and learn a small (linear or MLP) encoder and an MLP projector. There is no large backbone to learn and even datasets like \gld~\citep{weyand2020GLDv2} (composed of 1.5M vectors in 2048 dimensions for the \gemap model) can easily fit in memory. When it comes to training time, \tldr is noticeably slower than PCA. For example, for $d=128$ and for the \gld dataset, learning PCA takes approximately 18 minutes (multi-core), while learning a linear \tldr over 100 epochs takes approximately 63 minutes (on a single GPU). The latter includes 13 minutes for computing (exact) k-NNs for the dataset. 

Although slower at training time, linear \tldr highly outperforms PCA, while all other manifold learning methods that we compare to can neither reach the performance of PCA nor be applied to large training datasets for $d\ge 32$. 
More importantly, as we show in Appendix~\ref{sec:varyingtrainingsize},
the performance of PCA saturates early and, unlike \tldr, does not benefit from adding more data; Our study on a dataset of 1.5 million images suggests that the gap with PCA increases as the training set gets larger.
Finally, although the training time of \tldr is higher than PCA, both methods share the same \textit{linear encoding complexity} during testing, something highly important as this is a process repeated at every single retrieval request.

\mypartight{\tldr out of its comfort zone.}
Although \tldr can be seen as a way of generalizing recent self-supervised visual representation learning methods to cases where handcrafted transformations of the data are challenging or impossible to define, we want to emphasize that it \textit{is not suited} for self-supervised representation learning from pixels; augmentation invariance is a much more suited prior in that regard, while it is also practically impossible to define meaningful neighboring pairs from the input pixel space.
Additionally, although visualization is a common manifold learning application, \tldr is neither designed not recommended for 2D outputs; there are other methods like $t$-SNE, UMAP or MDE that specialize for such tasks (see also Appendix).
% ~\citet{van2008visualizing,mcinnes2018umap,agrawal2021minimum} that specialize for such tasks. 

\mypartight{What is \tldr suitable for?}
% \looseness=-1
\tldr excels for dimensionality reduction to mid-size outputs, \eg when $\dd$ is between 32 and 256 dimensions, a range to which the vast majority of manifold learning methods cannot scale. This is very useful in practice for retrieval. \tldr further provides a computationally efficient way of \textit{adapting} pre-trained representations, coming \eg from large pre-trained models, to a new domain or a new task, without the need for any labeled data from the downstream task, and without fine-tuning large encoders.
%without the need of labeled data for the downstream task nor the compute and know-how to fine-tune large encoders.} 
%% At the same time, \tldr enables the community to utilize a powerful learning framework initially tailored for visual representation learning~\citep{zbontar2021barlow} for dimensionality reduction and for a much wider set of tasks and domains like natural language. 
% \\
% \emph{\textbf{Easy-to-use code for TLRD will be made publicly available}.}

% \input{tex/4_discussion}
\section{Conclusions}

In this paper we introduce \method, a dimensionality-reduction method that combines neighborhood embedding learning with the simplicity and effectiveness of recent self-supervised learning losses. 

By simply replacing PCA with \tldr, one can significantly increase the 
performance of generic (DINO) or specialized (\gemap) models on retrieval tasks like \imnet or \rox, especially if one has access to \textit{unlabeled} data from the downstream retrieval domain. Those gains are achieved by learning a linear encoder on top of pre-extracted features and without the need to fine-tune a large backbone.
To summarize, \tldr offers a number of desirable properties:

%By simply replacing PCA with \tldr one can significantly increase the state-of-the-art landmark retrieval performance of \gemap~\citep{revaud2019aploss} and boost argument retrieval performance without additional computational cost.
% \method further offers a number of desirable properties:
\begin{inlinelist} 
    \item \emph{Scalability:} learned via stochastic gradient descent, \tldr can easily be parallelized across GPUs and machines, while for even the largest datasets, approximate nearest neighbor methods can be used to create input pairs in sub-linear complexity~\citep{babenko2014inverted,kalantidis2014locally}
    \item \emph{Simplicity:} The Barlow Twins~\citep{zbontar2021barlow} objective is robust and easy to optimize, and does not have trivial solutions 
    \item \emph{Out-of-sample generalization}
    \item \emph{Linear encoding complexity:} \tldr is highly effective  with a linear encoder, offering a direct replacement of PCA without extra encoding cost.
    % \item \emph{Robustness to hyper-parameters} like $k$, batch size and width.
\end{inlinelist}

\section*{Acknowledgements}
The authors want to sincerely thank Christopher Dance for his early comments that helped shape this work. We also want to thank our colleagues at NAVER LABS Europe\footnote{https://europe.naverlabs.com/} for providing feedback that made this work significantly better. Specifically we would like to thank Gabriela, J\'erome, Boris, Martin, St\'ephane, Bulent, Rafael, Gregory, Riccardo and Florent for reviewing our work and providing us with constructive comments. Last but not least, we want to thank Audi for hand-drawing our gorgeous teaser Figure. Finally, we want to thank all anonymous reviewers, whose comments also made this work better. This work was supported in part by
MIAI@Grenoble Alpes (ANR-19-P3IA-0003).

\bibliography{main}
\bibliographystyle{tmlr}
\newpage
\appendix
\renewcommand \thepart{}
\renewcommand \partname{}

\part{Appendix}
\parttoc
\renewcommand\thefigure{\Alph{figure}}    
\setcounter{figure}{0}    
\renewcommand\thetable{\Alph{table}}    
\setcounter{table}{0}    

\section{Appendix Summary}

In this appendix we present a number of additional details, results and Figures that we could not fit in the main text: 
\begin{itemize}
    % \onlyinsub{\item We present a table with a \textbf{summary of all compared methods} for the landmark retrieval experiments of the main paper (Table~\ref{tab:comparedmethods}).}
    \item We  report additional experiments for the \textbf{landmark retrieval} task in Appendix~\ref{app:landmark}. Specifically, we present results for the med/hard splits separately for \gemap (Appendix~\ref{sec:med_hard}), an experiment with oracle neighbors (Appendix~\ref{sec:oracle}), an experiment with features from a larger ResNet-101 backbone (Appendix~\ref{sec:resnetoneowone}). We also present some additional results with DINO representations (Appendix~\ref{sec:dinoresnetresults}).
    \item We present ablations when we \textbf{vary the projector architecture} and the \textbf{number of neighbors $k$} (Appendix~\ref{sec:varyingprojectorandk}), as well as when  \textbf{varying the training set size} (Appendix~\ref{sec:varyingtrainingsize}) and the \textbf{batch size} (Appendix~\ref{sec:varyingbatchsize}). We also compare the training times of \tldr and other manifold learning methods as we vary the size of the training set (Appendix~\ref{sec:varyingtrainingsizetime})
    \item We report additional experiments for the \textbf{document retrieval}  in Appendix~\ref{appendix:first_stage_retrieval}. Specifically, we extend our evaluation protocol and report result on a new task: \textbf{duplicate query retrieval}. We further investigate not only dimensionality reduction for the same task, but also the case of \textbf{dimensionality reduction transfer}.
    \item Although \tldr is not suited for such applications, as a proof of concept we present results on the FashionMNIST dataset in Appendix~\ref{sec:experiments_fashionmnist}, \ie when \textbf{learning from raw pixel data}. We also present some results when using \tldr for \textbf{visualization}, \ie when the output dimension is $\dd=2$ in Figure~\ref{fig:2dviz}.
    \item 
    % \arxivorsub{
    We extend Section~\ref{sec:discussion} with
    % }{We present}
    \textbf{further discussion} on related topics and more related works in Appendix~\ref{appendix:discussion}. We conclude with a brief discussion on limitations of \tldr in Appendix~\ref{sec:limitations}
    \item Finally, in Appendix~\ref{sec:pseudocode} we give pytorch-style pseudocode for initializing and training TLDR.
\end{itemize}
 
% \onlyinsub{

% \noindent\emph{\textbf{Easy-to-use code for \tldr is also part of the supplementary; usage instruction are included in the accompanying README.}}
% \section{Table of compared approaches}

% In Table~\ref{tab:comparedmethods} we present a summary of the approaches we compare to in the main text.
% \input{tables/table_baselines}
% }

\section{Additional experiments: Landmark image retrieval}
\label{app:landmark}

\subsection{Results on Medium and Hard protocols separately}
\label{sec:med_hard}

In Figure~\ref{fig:roxford5k_paris6k} we report the mAP metric for the Medium and Hard splits of the Revisited Oxford and Paris
datasets~\citep{radenovic2018revisiting} separately. 

\begin{figure}
\begin{center}
% -------------------------------------------------------------------------
% Oxford results
% -------------------------------------------------------------------------
\resizebox{\linewidth}{!}{
    \begin{subfigure}{.5\linewidth}
        \centering
        \begin{tikzpicture}
\begin{axis}[%
  width=\linewidth,
  xlabel={$\dd$},
  xtick = {1,2,3,4,5,6.2},
  xticklabels = {32,64,128,256,512,2048},
  ylabel={mAP},
  legend pos=south east,
  ylabel near ticks, xlabel near ticks, legend style={font=\scriptsize}, legend columns=2,
  height=5.5cm,
  minor y tick num=4,
  title={ROxford5K-Medium},
  label style={font=\small},
  title style={font=\small}
  ]

\pgfplotstableread{
d CNE_0-h2048 CNE_0-h2048-std CNE_1-h2048 CNE_1-h2048-std CNE_2-h2048 CNE_2-h2048-std CNE_1*-h2048 CNE_1*-h2048-std CNE_2*-h2048 CNE_2*-h2048-std Recon_0 Recon_0-std Synth-nn Synth-nn-std Contrastive Contrastive-std Contrastive-MLP Contrastive-MLP-std Contrastive+Proj Contrastive+Proj-std Supervised Supervised-std PCA-whiten baseline
1 0.56274 0.00541 0.56987 0.00684 0.57017 0.00625 0.58959 0.01027 0.52845 0.01613 0.52819 0.01336 0.50792 0.00211 0.38229 0.00063 0.35797 0.02813 0.52277 0.00598 0.54856 0.00471 0.49181 nan
2 0.59348 0.00394 0.59473 0.00378 0.59654 0.00479 0.62193 0.00766 0.58192 0.00959 0.56403 0.00583 0.56687 0.00088 0.38325 0.00151 0.33996 0.00292 0.56877 0.00238 0.58876 0.00617 0.55430 nan
3 0.61959 0.00552 0.62049 0.00121 0.62181 0.00292 0.62220 0.00777 0.60256 0.00641 0.57574 0.00351 0.59003 0.00153 0.38236 0.00282 0.34167 0.02649 0.60252 0.00239 0.62311 0.00340 0.58579 nan
4 0.64392 0.00434 0.64640 0.00522 0.63724 0.00245 0.61817 0.00798 0.59405 0.00754 0.58456 0.00205 0.60281 0.00009 0.38488 0.00183 0.34548 0.01590 0.61428 0.00129 0.64592 0.00135 0.61374 nan
5 0.65697 0.00240 0.65506 0.00419 0.64806 0.00365 0.61472 0.00556 0.60307 0.00392 0.58713 0.00136 0.60619 0.00063 0.38791 0.00046 0.33751 0.02313 0.62133 0.00044 0.65853 0.00077 0.63608 nan
6.2 nan nan nan nan nan nan nan nan nan nan nan nan nan nan nan nan nan nan nan nan nan nan nan 0.61855
}{\map}
    
    \addplot[cneplotlinear]      table[x=d,  y=CNE_0-h2048]   \map; \leg{\lcne}
    \addplot [name path=upper,draw=none, forget plot] table[x=d,y expr=\thisrow{CNE_0-h2048}+\thisrow{CNE_0-h2048-std}] {\map};
    \addplot [name path=lower,draw=none, forget plot] table[x=d,y expr=\thisrow{CNE_0-h2048}-\thisrow{CNE_0-h2048-std}] {\map};
    \addplot [fill=\cneplotlinearc!\stdgrad, forget plot] fill between[of=upper and lower];
    
    \addplot[cneplotflinearone]      table[x=d,  y=CNE_1-h2048]   \map; \leg{\flcne{1}}
    \addplot [name path=upper,draw=none, forget plot] table[x=d,y expr=\thisrow{CNE_1-h2048}+\thisrow{CNE_1-h2048-std}] {\map};
    \addplot [name path=lower,draw=none, forget plot] table[x=d,y expr=\thisrow{CNE_1-h2048}-\thisrow{CNE_1-h2048-std}] {\map};
    \addplot [fill=\cneplotflinearonec!\stdgrad, forget plot] fill between[of=upper and lower];

    \addplot[cneplotmlpone]      table[x=d,  y=CNE_1*-h2048]   \map; 
    \leg{\mlpcne{1}}
    \addplot [name path=upper,draw=none, forget plot] table[x=d,y expr=\thisrow{CNE_1*-h2048}+\thisrow{CNE_1*-h2048-std}] {\map};
    \addplot [name path=lower,draw=none, forget plot] table[x=d,y expr=\thisrow{CNE_1*-h2048}-\thisrow{CNE_1*-h2048-std}] {\map};
    \addplot [fill=\cneplotmlponec!\stdgrad, forget plot] fill between[of=upper and lower];

    \addplot[cneplotgaussian]      table[x=d,  y=Synth-nn]   \map; 
    \leg{\methodgauss}
    \addplot [name path=upper,draw=none, forget plot] table[x=d,y expr=\thisrow{Synth-nn}+\thisrow{Synth-nn-std}] {\map};
    \addplot [name path=lower,draw=none, forget plot] table[x=d,y expr=\thisrow{Synth-nn}-\thisrow{Synth-nn-std}] {\map};
    \addplot [fill=\cneplotgaussianc!\stdgrad, forget plot] fill between[of=upper and lower];
    
    \addplot[mseplot]      table[x=d,  y=Recon_0]   \map; 
    \leg{MSE}
    \addplot [name path=upper,draw=none, forget plot] table[x=d,y expr=\thisrow{Recon_0}+\thisrow{Recon_0-std}] {\map};
    \addplot [name path=lower,draw=none, forget plot] table[x=d,y expr=\thisrow{Recon_0}-\thisrow{Recon_0-std}] {\map};
    \addplot [fill=\mseplotc!\stdgrad, forget plot] fill between[of=upper and lower];

    \addplot[contrastiveplot]      table[x=d,  y=Contrastive+Proj]   \map; 
    \leg{\contrproj}
    \addplot [name path=upper,draw=none, forget plot] table[x=d,y expr=\thisrow{Contrastive+Proj}+\thisrow{Contrastive+Proj-std}] {\map};
    \addplot [name path=lower,draw=none, forget plot] table[x=d,y expr=\thisrow{Contrastive+Proj}-\thisrow{Contrastive+Proj-std}] {\map};
    \addplot [fill=\contrastiveplotc!\stdgrad, forget plot] fill between[of=upper and lower];

    \addplot[pcaplot]      table[x=d,  y=PCA-whiten]   \map; \leg{\pcaw}

    \addplot[baselineplot] table[x=d,  y=baseline]   \map; \leg{\gemap}
    
\end{axis}
\end{tikzpicture}
        \label{fig:roxford5k_medium}
    \end{subfigure}%
    \begin{subfigure}{.5\linewidth}
        \centering
        \begin{tikzpicture}
\begin{axis}[%
  width=\linewidth,
  xlabel={$\dd$},
  xtick = {1,2,3,4,5,6.2},
  xticklabels = {32,64,128,256,512,2048},
  ylabel={mAP},
  ymax=0.43,
  legend pos=south east,
  ylabel near ticks, xlabel near ticks, legend style={font=\scriptsize},
  legend columns=2,
  height=5.5cm,
  minor y tick num=4,
  title={ROxford5K-Hard},
  label style={font=\small},
  title style={font=\small}
  ]

\pgfplotstableread{
d CNE_0-h2048 CNE_0-h2048-std CNE_1-h2048 CNE_1-h2048-std CNE_2-h2048 CNE_2-h2048-std CNE_1*-h2048 CNE_1*-h2048-std CNE_2*-h2048 CNE_2*-h2048-std Recon_0 Recon_0-std Synth-nn Synth-nn-std Contrastive Contrastive-std Contrastive-MLP Contrastive-MLP-std Contrastive+Proj Contrastive+Proj-std Supervised Supervised-std PCA-whiten baseline
1 0.28533 0.00363 0.29343 0.00740 0.29129 0.00707 0.31559 0.01335 0.27183 0.01861 0.24376 0.01395 0.23251 0.00645 0.15187 0.00050 0.12609 0.03047 0.25146 0.00527 0.27449 0.00274 0.23427 nan
2 0.31552 0.01233 0.32173 0.00780 0.32212 0.00889 0.35906 0.01347 0.32963 0.01190 0.28309 0.00794 0.29695 0.00078 0.15150 0.00000 0.10668 0.01314 0.29170 0.00275 0.31810 0.00678 0.30960 nan
3 0.36160 0.01091 0.35912 0.00547 0.36379 0.00228 0.36027 0.01622 0.33827 0.01325 0.29220 0.00369 0.31996 0.00211 0.15095 0.00061 0.13933 0.01103 0.33118 0.00533 0.37405 0.00363 0.31321 nan
4 0.38932 0.00706 0.38502 0.01159 0.37964 0.00552 0.35573 0.01252 0.32485 0.01067 0.30108 0.00279 0.33247 0.00042 0.15194 0.00062 0.11776 0.01095 0.34705 0.00202 0.39264 0.00222 0.34909 nan
5 0.40347 0.00302 0.40210 0.00445 0.39352 0.00690 0.34948 0.00956 0.33525 0.00545 0.30423 0.00193 0.33677 0.00105 0.15191 0.00013 0.12963 0.01444 0.35583 0.00046 0.40530 0.00446 0.39243 nan
6.2 nan nan nan nan nan nan nan nan nan nan nan nan nan nan nan nan nan nan nan nan nan nan nan 0.34763
}{\map}
    
    \addplot[cneplotlinear]      table[x=d,  y=CNE_0-h2048]   \map; \leg{\lcne}
    \addplot [name path=upper,draw=none, forget plot] table[x=d,y expr=\thisrow{CNE_0-h2048}+\thisrow{CNE_0-h2048-std}] {\map};
    \addplot [name path=lower,draw=none, forget plot] table[x=d,y expr=\thisrow{CNE_0-h2048}-\thisrow{CNE_0-h2048-std}] {\map};
    \addplot [fill=\cneplotlinearc!\stdgrad, forget plot] fill between[of=upper and lower];
    
    \addplot[cneplotflinearone]      table[x=d,  y=CNE_1-h2048]   \map; \leg{\flcne{1}}
    \addplot [name path=upper,draw=none, forget plot] table[x=d,y expr=\thisrow{CNE_1-h2048}+\thisrow{CNE_1-h2048-std}] {\map};
    \addplot [name path=lower,draw=none, forget plot] table[x=d,y expr=\thisrow{CNE_1-h2048}-\thisrow{CNE_1-h2048-std}] {\map};
    \addplot [fill=\cneplotflinearonec!\stdgrad, forget plot] fill between[of=upper and lower];
    
    \addplot[cneplotmlpone]      table[x=d,  y=CNE_1*-h2048]   \map; 
    \leg{\mlpcne{1}}
    \addplot [name path=upper,draw=none, forget plot] table[x=d,y expr=\thisrow{CNE_1*-h2048}+\thisrow{CNE_1*-h2048-std}] {\map};
    \addplot [name path=lower,draw=none, forget plot] table[x=d,y expr=\thisrow{CNE_1*-h2048}-\thisrow{CNE_1*-h2048-std}] {\map};
    \addplot [fill=\cneplotmlponec!\stdgrad, forget plot] fill between[of=upper and lower];

    \addplot[cneplotgaussian]      table[x=d,  y=Synth-nn]   \map; 
    \leg{\methodgauss}
    \addplot [name path=upper,draw=none, forget plot] table[x=d,y expr=\thisrow{Synth-nn}+\thisrow{Synth-nn-std}] {\map};
    \addplot [name path=lower,draw=none, forget plot] table[x=d,y expr=\thisrow{Synth-nn}-\thisrow{Synth-nn-std}] {\map};
    \addplot [fill=\cneplotgaussianc!\stdgrad, forget plot] fill between[of=upper and lower];
        
    \addplot[mseplot]      table[x=d,  y=Recon_0]   \map; 
    \leg{MSE}
    \addplot [name path=upper,draw=none, forget plot] table[x=d,y expr=\thisrow{Recon_0}+\thisrow{Recon_0-std}] {\map};
    \addplot [name path=lower,draw=none, forget plot] table[x=d,y expr=\thisrow{Recon_0}-\thisrow{Recon_0-std}] {\map};
    \addplot [fill=\mseplotc!\stdgrad, forget plot] fill between[of=upper and lower];
    
    \addplot[contrastiveplot]      table[x=d,  y=Contrastive+Proj]   \map; 
    \leg{\contrproj}
    \addplot [name path=upper,draw=none, forget plot] table[x=d,y expr=\thisrow{Contrastive+Proj}+\thisrow{Contrastive+Proj-std}] {\map};
    \addplot [name path=lower,draw=none, forget plot] table[x=d,y expr=\thisrow{Contrastive+Proj}-\thisrow{Contrastive+Proj-std}] {\map};
    \addplot [fill=\contrastiveplotc!\stdgrad, forget plot] fill between[of=upper and lower];

    \addplot[pcaplot]      table[x=d,  y=PCA-whiten]   \map; \leg{\pcaw}

    \addplot[baselineplot] table[x=d,  y=baseline]   \map; \leg{\gemap}

\end{axis}
\end{tikzpicture}
        \label{fig:roxford5k_hard}
    \end{subfigure}
}
% -------------------------------------------------------------------------
% Paris results
% -------------------------------------------------------------------------
\resizebox{\linewidth}{!}{
    \begin{subfigure}{.5\linewidth}
        \centering
        \begin{tikzpicture}
\begin{axis}[%
  width=\linewidth,
  xlabel={$\dd$},
  xtick = {1,2,3,4,5,6.2},
  xticklabels = {32,64,128,256,512,2048},
  ylabel={mAP},
  legend pos=south east,
  ylabel near ticks, xlabel near ticks, legend style={font=\scriptsize}, legend columns=2,
  height=5.5cm,
  title={RParis6K-Medium},
    minor y tick num=4,
  label style={font=\small},
  title style={font=\small}
  ]

\pgfplotstableread{
d CNE_0-h2048 CNE_0-h2048-std CNE_1-h2048 CNE_1-h2048-std CNE_2-h2048 CNE_2-h2048-std CNE_1*-h2048 CNE_1*-h2048-std CNE_2*-h2048 CNE_2*-h2048-std Recon_0 Recon_0-std Synth-nn Synth-nn-std Contrastive Contrastive-std Contrastive-MLP Contrastive-MLP-std Contrastive+Proj Contrastive+Proj-std Supervised Supervised-std PCA-whiten baseline
1 0.71949 0.00425 0.71725 0.00785 0.71298 0.00762 0.71351 0.00331 0.70018 0.01051 0.67783 0.00917 0.65651 0.00309 0.57078 0.00034 0.56535 0.00522 0.66931 0.00465 0.71775 0.00221 0.62729 nan
2 0.74347 0.00307 0.74264 0.00122 0.74282 0.00265 0.75305 0.00560 0.72769 0.00702 0.72664 0.00450 0.72742 0.00047 0.57144 0.00046 0.56391 0.00049 0.71525 0.00032 0.73989 0.00224 0.71828 nan
3 0.77018 0.00123 0.76962 0.00111 0.76717 0.00208 0.76103 0.00414 0.74299 0.00769 0.74812 0.00272 0.74884 0.00033 0.56996 0.00190 0.56926 0.00474 0.74096 0.00120 0.76327 0.00291 0.74679 nan
4 0.77895 0.00279 0.78014 0.00169 0.77644 0.00246 0.76472 0.00383 0.75025 0.00375 0.74504 0.00158 0.75645 0.00026 0.57618 0.00261 0.55473 0.00879 0.75033 0.00139 0.77671 0.00135 0.75747 nan
5 0.78397 0.00148 0.78381 0.00177 0.78048 0.00129 0.76046 0.00412 0.75407 0.00589 0.74597 0.00079 0.75916 0.00053 0.57601 0.00050 0.56196 0.00213 0.75438 0.00115 0.77983 0.00405 0.76171 nan
6.2 nan nan nan nan nan nan nan nan nan nan nan nan nan nan nan nan nan nan nan nan nan nan nan 0.76532
}{\map}
    
    \addplot[cneplotlinear]      table[x=d,  y=CNE_0-h2048]   \map; \leg{\lcne}
    \addplot [name path=upper,draw=none, forget plot] table[x=d,y expr=\thisrow{CNE_0-h2048}+\thisrow{CNE_0-h2048-std}] {\map};
    \addplot [name path=lower,draw=none, forget plot] table[x=d,y expr=\thisrow{CNE_0-h2048}-\thisrow{CNE_0-h2048-std}] {\map};
    \addplot [fill=\cneplotlinearc!\stdgrad, forget plot] fill between[of=upper and lower];
    
    \addplot[cneplotflinearone]      table[x=d,  y=CNE_1-h2048]   \map; \leg{\flcne{1}}
    \addplot [name path=upper,draw=none, forget plot] table[x=d,y expr=\thisrow{CNE_1-h2048}+\thisrow{CNE_1-h2048-std}] {\map};
    \addplot [name path=lower,draw=none, forget plot] table[x=d,y expr=\thisrow{CNE_1-h2048}-\thisrow{CNE_1-h2048-std}] {\map};
    \addplot [fill=\cneplotflinearonec!\stdgrad, forget plot] fill between[of=upper and lower];

    \addplot[cneplotmlpone]      table[x=d,  y=CNE_1*-h2048]   \map; 
    \leg{\mlpcne{1}}
    \addplot [name path=upper,draw=none, forget plot] table[x=d,y expr=\thisrow{CNE_1*-h2048}+\thisrow{CNE_1*-h2048-std}] {\map};
    \addplot [name path=lower,draw=none, forget plot] table[x=d,y expr=\thisrow{CNE_1*-h2048}-\thisrow{CNE_1*-h2048-std}] {\map};
    \addplot [fill=\cneplotmlponec!\stdgrad, forget plot] fill between[of=upper and lower];
    
    % \addplot[cneplotflinearone]      table[x=d,  y=Contrastive]   \map; \leg{Contr}
    % \addplot [name path=upper,draw=none, forget plot] table[x=d,y expr=\thisrow{Contrastive}+\thisrow{Contrastive-std}] {\map};
    % \addplot [name path=lower,draw=none, forget plot] table[x=d,y expr=\thisrow{Contrastive}-\thisrow{Contrastive-std}] {\map};
    % \addplot [fill=\cneplotflinearonec!\stdgrad, forget plot] fill between[of=upper and lower];
    
    % \addplot[cneplotflineartwo]      table[x=d,  y=Contrastive-MLP]   \map; \leg{$\textrm{Contr}^\star$}
    % \addplot [name path=upper,draw=none, forget plot] table[x=d,y expr=\thisrow{Contrastive-MLP}+\thisrow{Contrastive-MLP-std}] {\map};
    % \addplot [name path=lower,draw=none, forget plot] table[x=d,y expr=\thisrow{Contrastive-MLP}-\thisrow{Contrastive-MLP-std}] {\map};
    % \addplot [fill=\cneplotflineartwoc!\stdgrad, forget plot] fill between[of=upper and lower];
    
    \addplot[cneplotgaussian]      table[x=d,  y=Synth-nn]   \map; 
    \leg{\methodgauss}
    \addplot [name path=upper,draw=none, forget plot] table[x=d,y expr=\thisrow{Synth-nn}+\thisrow{Synth-nn-std}] {\map};
    \addplot [name path=lower,draw=none, forget plot] table[x=d,y expr=\thisrow{Synth-nn}-\thisrow{Synth-nn-std}] {\map};
    \addplot [fill=\cneplotgaussianc!\stdgrad, forget plot] fill between[of=upper and lower];
    
    \addplot[mseplot]      table[x=d,  y=Recon_0]   \map; 
    \leg{MSE}
    \addplot [name path=upper,draw=none, forget plot] table[x=d,y expr=\thisrow{Recon_0}+\thisrow{Recon_0-std}] {\map};
    \addplot [name path=lower,draw=none, forget plot] table[x=d,y expr=\thisrow{Recon_0}-\thisrow{Recon_0-std}] {\map};
    \addplot [fill=\mseplotc!\stdgrad, forget plot] fill between[of=upper and lower];

    \addplot[contrastiveplot]      table[x=d,  y=Contrastive+Proj]   \map; 
    \leg{\contrproj}
    \addplot [name path=upper,draw=none, forget plot] table[x=d,y expr=\thisrow{Contrastive+Proj}+\thisrow{Contrastive+Proj-std}] {\map};
    \addplot [name path=lower,draw=none, forget plot] table[x=d,y expr=\thisrow{Contrastive+Proj}-\thisrow{Contrastive+Proj-std}] {\map};
    \addplot [fill=\contrastiveplotc!\stdgrad, forget plot] fill between[of=upper and lower];

    % \addplot[tldrplotsup]      table[x=d,  y=Supervised]   \map; 
    % \leg{Supervised}
    % \addplot [name path=upper,draw=none, forget plot] table[x=d,y expr=\thisrow{Supervised}+\thisrow{Supervised-std}] {\map};
    % \addplot [name path=lower,draw=none, forget plot] table[x=d,y expr=\thisrow{Supervised}-\thisrow{Supervised-std}] {\map};
    % \addplot [fill=\tldrplotsupc!\stdgrad, forget plot] fill between[of=upper and lower];

    % \addplot[drlimplot]      table[x=d,  y=Contrastive]   \map; \leg{DrLIM}
    % \addplot [name path=upper,draw=none, forget plot] table[x=d,y expr=\thisrow{Contrastive}+\thisrow{Contrastive-std}] {\map};
    % \addplot [name path=lower,draw=none, forget plot] table[x=d,y expr=\thisrow{Contrastive}-\thisrow{Contrastive-std}] {\map};
    % \addplot [fill=\drlimplotc!\stdgrad, forget plot] fill between[of=upper and lower];

    \addplot[pcaplot]      table[x=d,  y=PCA-whiten]   \map; \leg{\pcaw}

    \addplot[baselineplot] table[x=d,  y=baseline]   \map; \leg{\gemap}

    \end{axis}
\end{tikzpicture}
        \label{fig:rparis6k_medium}
    \end{subfigure}%
    \begin{subfigure}{.5\linewidth}
        \centering
        \begin{tikzpicture}
\begin{axis}[%
  width=\linewidth,
  xlabel={$\dd$},
  xtick = {1,2,3,4,5,6.2},
  xticklabels = {32,64,128,256,512,2048},
  ylabel={mAP},
  legend pos=south east,
  ylabel near ticks, xlabel near ticks, legend style={font=\scriptsize},
  legend columns=2,
  height=5.5cm, 
  minor y tick num=4,
  title={RParis6K-Hard},
  label style={font=\small},
  title style={font=\small}
  ]

\pgfplotstableread{
d CNE_0-h2048 CNE_0-h2048-std CNE_1-h2048 CNE_1-h2048-std CNE_2-h2048 CNE_2-h2048-std CNE_1*-h2048 CNE_1*-h2048-std CNE_2*-h2048 CNE_2*-h2048-std Recon_0 Recon_0-std Synth-nn Synth-nn-std Contrastive Contrastive-std Contrastive-MLP Contrastive-MLP-std Contrastive+Proj Contrastive+Proj-std Supervised Supervised-std PCA-whiten baseline
1 0.50429 0.00744 0.50450 0.01100 0.49736 0.00988 0.51615 0.00454 0.51279 0.01223 0.45357 0.00888 0.42449 0.00412 0.34741 0.00015 0.37342 0.00032 0.44199 0.00435 0.49793 0.00217 0.39592 nan
2 0.53145 0.00544 0.52841 0.00284 0.52635 0.00403 0.56025 0.00967 0.53528 0.00813 0.50917 0.00771 0.51333 0.00050 0.34779 0.00008 0.36179 0.00057 0.49199 0.00198 0.52476 0.00610 0.49353 nan
3 0.56486 0.00168 0.56054 0.00263 0.55787 0.00261 0.56151 0.00521 0.55048 0.01022 0.53946 0.00218 0.53658 0.00066 0.34776 0.00079 0.37201 0.00218 0.52545 0.00304 0.55337 0.00324 0.52378 nan
4 0.57777 0.00401 0.57738 0.00304 0.57340 0.00338 0.56398 0.00862 0.56176 0.00676 0.53437 0.00200 0.54605 0.00031 0.35181 0.00244 0.36305 0.01373 0.53513 0.00224 0.57509 0.00249 0.53775 nan
5 0.58548 0.00239 0.58266 0.00224 0.57958 0.00154 0.55688 0.00406 0.56023 0.01145 0.53639 0.00121 0.54972 0.00055 0.35325 0.00010 0.35877 0.00160 0.54296 0.00189 0.57793 0.00595 0.54850 nan
6.2 nan nan nan nan nan nan nan nan nan nan nan nan nan nan nan nan nan nan nan nan nan nan nan 0.55850
}{\map}
    
    \addplot[cneplotlinear]      table[x=d,  y=CNE_0-h2048]   \map; \leg{\lcne}
    \addplot [name path=upper,draw=none, forget plot] table[x=d,y expr=\thisrow{CNE_0-h2048}+\thisrow{CNE_0-h2048-std}] {\map};
    \addplot [name path=lower,draw=none, forget plot] table[x=d,y expr=\thisrow{CNE_0-h2048}-\thisrow{CNE_0-h2048-std}] {\map};
    \addplot [fill=\cneplotlinearc!\stdgrad, forget plot] fill between[of=upper and lower];
    
    \addplot[cneplotflinearone]      table[x=d,  y=CNE_1-h2048]   \map; \leg{\flcne{1}}
    \addplot [name path=upper,draw=none, forget plot] table[x=d,y expr=\thisrow{CNE_1-h2048}+\thisrow{CNE_1-h2048-std}] {\map};
    \addplot [name path=lower,draw=none, forget plot] table[x=d,y expr=\thisrow{CNE_1-h2048}-\thisrow{CNE_1-h2048-std}] {\map};
    \addplot [fill=\cneplotflinearonec!\stdgrad, forget plot] fill between[of=upper and lower];
    
    \addplot[cneplotmlpone]      table[x=d,  y=CNE_1*-h2048]   \map; 
    \leg{\mlpcne{1}}
    \addplot [name path=upper,draw=none, forget plot] table[x=d,y expr=\thisrow{CNE_1*-h2048}+\thisrow{CNE_1*-h2048-std}] {\map};
    \addplot [name path=lower,draw=none, forget plot] table[x=d,y expr=\thisrow{CNE_1*-h2048}-\thisrow{CNE_1*-h2048-std}] {\map};
    \addplot [fill=\cneplotmlponec!\stdgrad, forget plot] fill between[of=upper and lower];

    \addplot[cneplotgaussian]      table[x=d,  y=Synth-nn]   \map; 
    \leg{\methodgauss}
    \addplot [name path=upper,draw=none, forget plot] table[x=d,y expr=\thisrow{Synth-nn}+\thisrow{Synth-nn-std}] {\map};
    \addplot [name path=lower,draw=none, forget plot] table[x=d,y expr=\thisrow{Synth-nn}-\thisrow{Synth-nn-std}] {\map};
    \addplot [fill=\cneplotgaussianc!\stdgrad, forget plot] fill between[of=upper and lower];
        
    \addplot[mseplot]      table[x=d,  y=Recon_0]   \map; 
    \leg{MSE}
    \addplot [name path=upper,draw=none, forget plot] table[x=d,y expr=\thisrow{Recon_0}+\thisrow{Recon_0-std}] {\map};
    \addplot [name path=lower,draw=none, forget plot] table[x=d,y expr=\thisrow{Recon_0}-\thisrow{Recon_0-std}] {\map};
    \addplot [fill=\mseplotc!\stdgrad, forget plot] fill between[of=upper and lower];
    
    \addplot[contrastiveplot]      table[x=d,  y=Contrastive+Proj]   \map; 
    \leg{\contrproj}
    \addplot [name path=upper,draw=none, forget plot] table[x=d,y expr=\thisrow{Contrastive+Proj}+\thisrow{Contrastive+Proj-std}] {\map};
    \addplot [name path=lower,draw=none, forget plot] table[x=d,y expr=\thisrow{Contrastive+Proj}-\thisrow{Contrastive+Proj-std}] {\map};
    \addplot [fill=\contrastiveplotc!\stdgrad, forget plot] fill between[of=upper and lower];

    \addplot[pcaplot]      table[x=d,  y=PCA-whiten]   \map; \leg{\pcaw}

    \addplot[baselineplot] table[x=d,  y=baseline]   \map; \leg{\gemap}

\end{axis}
\end{tikzpicture}
        \label{fig:rparis6k_hard}
    \end{subfigure}
}
\caption{\textbf{Image retrieval experiments}. Mean average precision (mAP) on
  \rox (top) and \rpa (bottom), for the Medium (left) and Hard (right)
  test sets, as a function of the output dimensions $\dd$. We report \method
  with different encoders: \linear (\lcne), \flinear with 1 hidden layer
  (\flcne{1}), and a MLP with 1 hidden layer (\mlpcne{1}), the projector remains
  the same (MLP with 2 hidden layers). We compare with two baselines based on
  \lcne, but which respectively train with a reconstruction (MSE) and a
  contrastive (\contrproj) loss. Our main baselines are PCA with whitening, and the original
  2048-dimentional features (\gemap~\cite{revaud2019aploss}), \ie before projection.}

\label{fig:roxford5k_paris6k}
\end{center}
\end{figure}

\subsection{Results with ``Oracle'' nearest neighbors} 
\label{sec:oracle}
\begin{figure}
\begin{center}
\resizebox{\linewidth}{!}{
    \begin{subfigure}{.5\linewidth}
        \centering
        \begin{tikzpicture}
\begin{axis}[%
  width=\linewidth,
  xlabel={$\dd$},
  xtick = {1,2,3,4,5,6.2},
  xticklabels = {32,64,128,256,512,2048},
  ylabel={mAP},
  legend pos=south east,
  ylabel near ticks, xlabel near ticks, legend style={font=\scriptsize},
  height=5cm,
  title={ROxford5K-Medium},
  label style={font=\small},
  title style={font=\small}
  ]

\pgfplotstableread{
d CNE_0-h2048 CNE_0-h2048-std CNE_1-h2048 CNE_1-h2048-std CNE_2-h2048 CNE_2-h2048-std CNE_1*-h2048 CNE_1*-h2048-std CNE_2*-h2048 CNE_2*-h2048-std Recon_0 Recon_0-std Synth-nn Synth-nn-std Contrastive Contrastive-std Contrastive-MLP Contrastive-MLP-std Contrastive+Proj Contrastive+Proj-std Supervised Supervised-std CNE_0-h2048_whr CNE_0-h2048_whr-std PCA-whiten ICA-whiten ICA-whiten-std LLE LLE-std LTSA LTSA-std baseline
1 0.56274 0.00541 0.56987 0.00684 0.57017 0.00625 0.58959 0.01027 0.52845 0.01613 0.52819 0.01336 0.50792 0.00211 0.38229 0.00063 0.35797 0.02813 0.52277 0.00598 0.54856 0.00471 0.56274 0.00541 0.49181 0.48936 0.00000 0.07542 0.00000 0.09506 0.00000 nan
2 0.59348 0.00394 0.59473 0.00378 0.59654 0.00479 0.62193 0.00766 0.58192 0.00959 0.56403 0.00583 0.56687 0.00088 0.38325 0.00151 0.33996 0.00292 0.56877 0.00238 0.58876 0.00617 0.59348 0.00394 0.55430 0.55605 0.00000 0.08764 0.00000 0.13733 0.00000 nan
3 0.61959 0.00552 0.62049 0.00121 0.62181 0.00292 0.62220 0.00777 0.60256 0.00641 0.57574 0.00351 0.59003 0.00153 0.38236 0.00282 0.34167 0.02649 0.60252 0.00239 0.62311 0.00340 0.61959 0.00552 0.58579 0.58729 0.00000 0.10227 0.00000 0.24739 0.00000 nan
4 0.64392 0.00434 0.64640 0.00522 0.63724 0.00245 0.61817 0.00798 0.59405 0.00754 0.58456 0.00205 0.60281 0.00009 0.38488 0.00183 0.34548 0.01590 0.61428 0.00129 0.64592 0.00135 0.64392 0.00434 0.61374 0.61740 0.00000 0.11193 0.00000 0.40454 0.00000 nan
5 0.65697 0.00240 0.65506 0.00419 0.64806 0.00365 0.61472 0.00556 0.60307 0.00392 0.58713 0.00136 0.60619 0.00063 0.38791 0.00046 0.33751 0.02313 0.62133 0.00044 0.65853 0.00077 0.65697 0.00240 0.63608 0.64163 0.00000 0.10898 0.00000 0.49851 0.00000 nan
6.2 nan nan nan nan nan nan nan nan nan nan nan nan nan nan nan nan nan nan nan nan nan nan nan nan nan nan nan nan nan nan nan 0.61855
}{\map}
    
    \addplot[cneplotlinear]      table[x=d,  y=CNE_0-h2048]   \map; \leg{\lcne}
    \addplot [name path=upper,draw=none, forget plot] table[x=d,y expr=\thisrow{CNE_0-h2048}+\thisrow{CNE_0-h2048-std}] {\map};
    \addplot [name path=lower,draw=none, forget plot] table[x=d,y expr=\thisrow{CNE_0-h2048}-\thisrow{CNE_0-h2048-std}] {\map};
    \addplot [fill=\cneplotlinearc!\stdgrad, forget plot] fill between[of=upper and lower];
    
    \addplot[cneplotsup]      table[x=d,  y=Supervised]   \map; 
    \leg{\supervised}
    \addplot [name path=upper,draw=none, forget plot] table[x=d,y expr=\thisrow{Supervised}+\thisrow{Supervised-std}] {\map};
    \addplot [name path=lower,draw=none, forget plot] table[x=d,y expr=\thisrow{Supervised}-\thisrow{Supervised-std}] {\map};
    \addplot [fill=\cneplotsupc!\stdgrad, forget plot] fill between[of=upper and lower];

    \addplot[drlimplot]      table[x=d,  y=ICA-whiten]   \map; \leg{\icaw}
    
    \addplot[pcaplot]      table[x=d,  y=PCA-whiten]   \map; \leg{\pcaw}
 
    \addplot[baselineplot] table[x=d,  y=baseline]   \map; \leg{\gemap}
    
\end{axis}
\end{tikzpicture}
        \label{fig:roxford5k_medium_oracle}
    \end{subfigure}%
    \begin{subfigure}{.5\linewidth}
        \centering
        \begin{tikzpicture}
\begin{axis}[%
  width=\linewidth,
  xlabel={$\dd$},
  xtick = {1,2,3,4,5,6.2},
  xticklabels = {32,64,128,256,512,2048},
  ylabel={mAP},
  legend pos=south east,
  ylabel near ticks, xlabel near ticks, legend style={font=\scriptsize},
  height=5cm,
  title={ROxford5K-Hard},
  label style={font=\small},
  title style={font=\small}
  ]

\pgfplotstableread{
d CNE_0-h2048 CNE_0-h2048-std CNE_1-h2048 CNE_1-h2048-std CNE_2-h2048 CNE_2-h2048-std CNE_1*-h2048 CNE_1*-h2048-std CNE_2*-h2048 CNE_2*-h2048-std Recon_0 Recon_0-std Synth-nn Synth-nn-std Contrastive Contrastive-std Contrastive-MLP Contrastive-MLP-std Contrastive+Proj Contrastive+Proj-std Supervised Supervised-std CNE_0-h2048_whr CNE_0-h2048_whr-std PCA-whiten ICA-whiten ICA-whiten-std LLE LLE-std LTSA LTSA-std baseline
1 0.28533 0.00363 0.29343 0.00740 0.29129 0.00707 0.31559 0.01335 0.27183 0.01861 0.24376 0.01395 0.23251 0.00645 0.15187 0.00050 0.12609 0.03047 0.25146 0.00527 0.27449 0.00274 0.28533 0.00363 0.23427 0.22403 0.00000 0.01982 0.00000 0.02722 0.00000 nan
2 0.31552 0.01233 0.32173 0.00780 0.32212 0.00889 0.35906 0.01347 0.32963 0.01190 0.28309 0.00794 0.29695 0.00078 0.15150 0.00000 0.10668 0.01314 0.29170 0.00275 0.31810 0.00678 0.31552 0.01233 0.30960 0.31004 0.00000 0.02406 0.00000 0.03928 0.00000 nan
3 0.36160 0.01091 0.35912 0.00547 0.36379 0.00228 0.36027 0.01622 0.33827 0.01325 0.29220 0.00369 0.31996 0.00211 0.15095 0.00061 0.13933 0.01103 0.33118 0.00533 0.37405 0.00363 0.36160 0.01091 0.31321 0.32210 0.00000 0.02846 0.00000 0.06254 0.00000 nan
4 0.38932 0.00706 0.38502 0.01159 0.37964 0.00552 0.35573 0.01252 0.32485 0.01067 0.30108 0.00279 0.33247 0.00042 0.15194 0.00062 0.11776 0.01095 0.34705 0.00202 0.39264 0.00222 0.38932 0.00706 0.34909 0.35191 0.00000 0.03055 0.00000 0.19017 0.00000 nan
5 0.40347 0.00302 0.40210 0.00445 0.39352 0.00690 0.34948 0.00956 0.33525 0.00545 0.30423 0.00193 0.33677 0.00105 0.15191 0.00013 0.12963 0.01444 0.35583 0.00046 0.40530 0.00446 0.40347 0.00302 0.39243 0.40216 0.00000 0.02576 0.00000 0.23863 0.00000 nan
6.2 nan nan nan nan nan nan nan nan nan nan nan nan nan nan nan nan nan nan nan nan nan nan nan nan nan nan nan nan nan nan nan 0.34763
}{\map}
    
    \addplot[cneplotlinear]      table[x=d,  y=CNE_0-h2048]   \map; \leg{\lcne}
    \addplot [name path=upper,draw=none, forget plot] table[x=d,y expr=\thisrow{CNE_0-h2048}+\thisrow{CNE_0-h2048-std}] {\map};
    \addplot [name path=lower,draw=none, forget plot] table[x=d,y expr=\thisrow{CNE_0-h2048}-\thisrow{CNE_0-h2048-std}] {\map};
    \addplot [fill=\cneplotlinearc!\stdgrad, forget plot] fill between[of=upper and lower];
  
     \addplot[cneplotsup]      table[x=d,  y=Supervised]   \map; 
    \leg{\supervised}
    \addplot [name path=upper,draw=none, forget plot] table[x=d,y expr=\thisrow{Supervised}+\thisrow{Supervised-std}] {\map};
    \addplot [name path=lower,draw=none, forget plot] table[x=d,y expr=\thisrow{Supervised}-\thisrow{Supervised-std}] {\map};
    \addplot [fill=\cneplotsupc!\stdgrad, forget plot] fill between[of=upper and lower];

    \addplot[drlimplot]      table[x=d,  y=ICA-whiten]   \map; \leg{\icaw}
    
    \addplot[pcaplot]      table[x=d,  y=PCA-whiten]   \map; \leg{\pcaw}

    \addplot[baselineplot] table[x=d,  y=baseline]   \map; \leg{\gemap}

\end{axis}
\end{tikzpicture}
        \label{fig:roxford5k_hard_oracle}
    \end{subfigure}
}
\caption{\textbf{Neighbor-supervised with oracle}. Mean average precision (mAP) on ROxford5K~\citep{radenovic2018revisiting} for the Medium (left) and Hard (right) test sets, as a function of the output dimensions $\dd$. We compare TLDR with an oracle version that uses labels to select training pairs. We include as baselines both PCA and ICA with whitening, and the original 2048-dimensional features (GeM-AP [32]), i.e. before projection.}
\label{fig:roxford5k_oracle}
\end{center}
\end{figure}
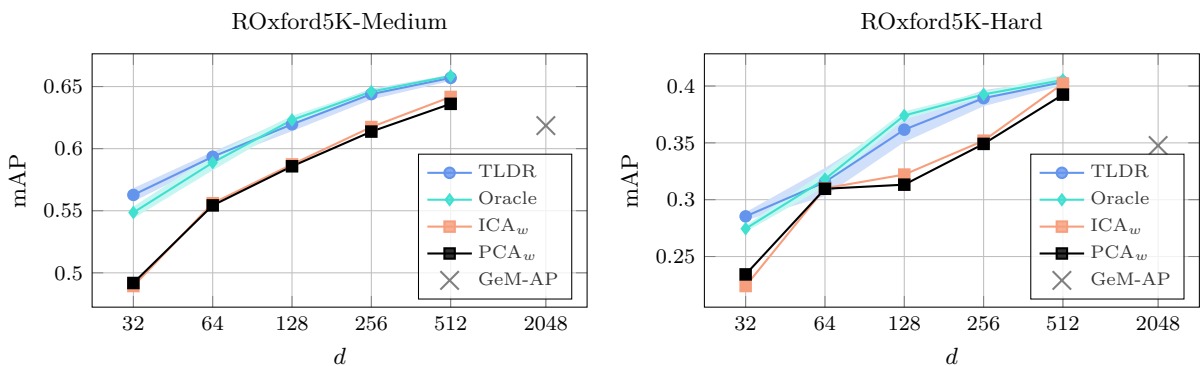

In Figure~\ref{fig:roxford5k_oracle} we present results using an oracle version of \tldr, \ie a version that uses labels to only keep as pairs neighbors that come from the same landmark in the training set. As we see, \tldr practically matches the oracle's performance.

% -------------------------------------------------------------------------
% Oxford results
% -------------------------------------------------------------------------
% -------------------------------------------------------------------------

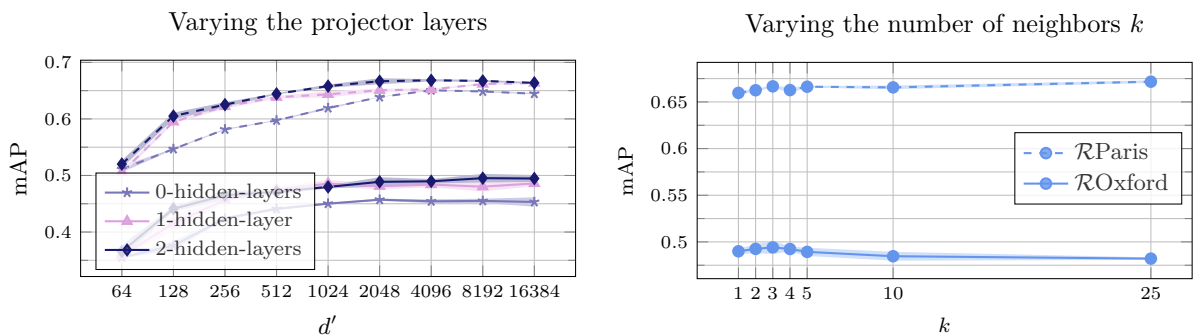
\begin{figure}
% \vspace{-10pt}
\begin{center}
\resizebox{\linewidth}{!}{
    \begin{subfigure}{.5\linewidth}
        \centering
        \begin{tikzpicture}
\begin{axis}[%
  width=\linewidth,
  xlabel={$\ddd$},
  xtick = {64,128,256,512,1024,2048,4096,8192,16384},
  xticklabels = {64,128,256,512,1024,2048,4096,8192,16384},
  ylabel={mAP},
  legend pos=south west,
      minor y tick num=1,
  xmode=log,
  ylabel near ticks, xlabel near ticks,
  height=4.5cm,
  title=Varying the projector layers,
 legend style={font=\footnotesize},
  tick label style={font=\scriptsize},
    label style={font=\small},
  ]

\pgfplotstableread{
d  TLDR-prl0 TLDR-prl0-std TLDR-prl1 TLDR-prl1-std TLDR-prl2 TLDR-prl2-std TLDR-prl0-Pa TLDR-prl0-std-Pa TLDR-prl1-Pa TLDR-prl1-std-Pa TLDR-prl2-Pa TLDR-prl2-std-Pa
64	0.35526	0.00204550727204769	0.358405	0.00201679200712419	0.365615	0.0110114326951582	0.511065	0.0054001157395004	0.50547	0.00471008492492439	0.520175	0.00675860932440987
128	0.375795	0.00786504291151676	0.414985	0.0136022222449128	0.441115	0.00477849348644528	0.546755	0.00107517440445725	0.59486	0.00254031494110474	0.60504	0.0059622017745125
256	0.42406	0.00229836898691224	0.456845	0.00496638701673561	0.46502	0.00740978407242748	0.58148	0.00104	0.62212	0.00401138380113397	0.62523	0.00435013218189976
512	0.441	0.00255033331154969	0.475575	0.00460895866763849	0.47091	0.00396823638408802	0.59731	0.000775016128864426	0.63831	0.00222073186134661	0.64429	0.00106660676915159
1024	0.45021	0.00160475854881661	0.48553	0.00610442462481109	0.479735	0.00265224433263604	0.619185	0.00181748452538117	0.643505	0.00551330209584057	0.657985	0.00283835339589699
2048	0.45715	0.00174884247432409	0.48096	0.00521080128195271	0.48878	0.00806953840067696	0.638605	0.000326879182573623	0.65071	0.00299282475263755	0.66663	0.00530143848403431
4096	0.45433	0.00347491726520215	0.48391	0.0049331784885609	0.489685	0.00401970770081607	0.65068	0.000733348484691964	0.65214	0.00132599019604219	0.668295	0.00202785601066742
8192	0.45505	0.00386791933731819	0.480165	0.00822601361535465	0.49525	0.00752736673744544	0.64848	0.00106491783720623	0.66169	0.00182002747232013	0.6672	0.00173834691589452
16384	0.453155	0.00822514437563256	0.485995	0.00503359215670082	0.494615	0.00608523623206199	0.644615	0.00109025226438655	0.66347	0.00158254541798964	0.66374	0.00122262013724623
}{\map}

    \addplot[cneplotlinearplzero]      table[x=d,  y=TLDR-prl0]   \map; \leg{0-hidden-layers}
    \addplot [name path=upper,draw=none, forget plot] table[x=d,y expr=\thisrow{TLDR-prl0}+\thisrow{TLDR-prl0-std}] {\map};
    \addplot [name path=lower,draw=none, forget plot] table[x=d,y expr=\thisrow{TLDR-prl0}-\thisrow{TLDR-prl0-std}] {\map};
    \addplot [fill=\cneplotlinearplzeroc!\stdgrad, forget plot] fill between[of=upper and lower];
    
    \addplot[cneplotlinearplone]      table[x=d,  y=TLDR-prl1]   \map; \leg{1-hidden-layer}
    \addplot [name path=upper,draw=none, forget plot] table[x=d,y expr=\thisrow{TLDR-prl1}+\thisrow{TLDR-prl1-std}] {\map};
    \addplot [name path=lower,draw=none, forget plot] table[x=d,y expr=\thisrow{TLDR-prl1}-\thisrow{TLDR-prl1-std}] {\map};
    \addplot [fill=\cneplotlinearplonec!\stdgrad, forget plot] fill between[of=upper and lower];

    \addplot[cneplotlinearpltwo]      table[x=d,  y=TLDR-prl2]   \map; \leg{2-hidden-layers}
    \addplot [name path=upper,draw=none, forget plot] table[x=d,y expr=\thisrow{TLDR-prl2}+\thisrow{TLDR-prl2-std}] {\map};
    \addplot [name path=lower,draw=none, forget plot] table[x=d,y expr=\thisrow{TLDR-prl2}-\thisrow{TLDR-prl2-std}] {\map};
    \addplot [fill=\cneplotlinearpltwoc!\stdgrad, forget plot] fill between[of=upper and lower];
    
    \addplot[cneplotlinearplzero, forget plot, dashed]      table[x=d,  y=TLDR-prl0-Pa]   \map; \leg{\lcne-0pl}
    \addplot [name path=upper,draw=none, forget plot] table[x=d,y expr=\thisrow{TLDR-prl0-Pa}+\thisrow{TLDR-prl0-std-Pa}] {\map};
    \addplot [name path=lower,draw=none, forget plot] table[x=d,y expr=\thisrow{TLDR-prl0-Pa}-\thisrow{TLDR-prl0-std-Pa}] {\map};
    \addplot [fill=\cneplotlinearplzeroc!\stdgrad, forget plot] fill between[of=upper and lower];
    
    \addplot[cneplotlinearplone, forget plot, dashed]      table[x=d,  y=TLDR-prl1-Pa]   \map; \leg{\lcne-1pl}
    \addplot [name path=upper,draw=none, forget plot] table[x=d,y expr=\thisrow{TLDR-prl1-Pa}+\thisrow{TLDR-prl1-std-Pa}] {\map};
    \addplot [name path=lower,draw=none, forget plot] table[x=d,y expr=\thisrow{TLDR-prl1-Pa}-\thisrow{TLDR-prl1-std-Pa}] {\map};
    \addplot [fill=\cneplotlinearplonec!\stdgrad, forget plot] fill between[of=upper and lower];

    \addplot[cneplotlinearpltwo, forget plot, dashed]      table[x=d,  y=TLDR-prl2-Pa]   \map; \leg{\lcne-2pl}
    \addplot [name path=upper,draw=none, forget plot] table[x=d,y expr=\thisrow{TLDR-prl2-Pa}+\thisrow{TLDR-prl2-std-Pa}] {\map};
    \addplot [name path=lower,draw=none, forget plot] table[x=d,y expr=\thisrow{TLDR-prl2-Pa}-\thisrow{TLDR-prl2-std-Pa}] {\map};
    \addplot [fill=\cneplotlinearpltwoc!\stdgrad, forget plot] fill between[of=upper and lower];
    
\end{axis}
\end{tikzpicture}
  
        \label{fig:rparis6k_med_pdims2}
    \end{subfigure}%
    \begin{subfigure}{.5\linewidth}
        \centering
        \begin{tikzpicture}
\begin{axis}[%
  width=\linewidth,
  xlabel={$k$},
  xtick = {1, 2, 3, 4, 5, 10, 25},
  xticklabels = {1, 2, 3, 4, 5, 10, 25},
  ylabel={mAP},
      minor y tick num=1,
  legend style={at={(0.98,0.51)},anchor=east, font=\small},
  ylabel near ticks, xlabel near ticks, height=4.5cm,
  label style={font=\footnotesize},
    tick label style={font=\scriptsize},
    title=Varying the number of neighbors $k$,
  ]

\pgfplotstableread{
d CNE CNE-std CNEpar CNEpar-std PCA
1	0.48995	0.00286007867024668	0.65969	0.00194092761328186	nan
2	0.492525	0.00471277519090398	0.66252	0.00163720798923045	nan
3	0.494145	0.00706150833745879	0.666715	0.000905013812049297	nan
4	0.492385	0.00267059918370391	0.66271	0.00278410667899059	nan
5	0.48932	0.00440323176769064	0.666305	0.000935334164884401	nan
10	0.48465	0.00473177556526089	0.665605	0.00193752419339734	nan
25	0.482085	0.00138311243216161	0.6717	0.00138003623140844	nan
}{\map}

\addplot[cneplotlinear, dashed]      table[x=d,  y=CNEpar]   \map; \leg{\rpa}
    \addplot [name path=upper,draw=none, forget plot] table[x=d,y expr=\thisrow{CNEpar}+\thisrow{CNEpar-std}] {\map};
    \addplot [name path=lower,draw=none, forget plot] table[x=d,y expr=\thisrow{CNEpar}-\thisrow{CNEpar-std}] {\map};
    \addplot [fill=\cneplotlinearc!\stdgrad, forget plot] fill between[of=upper and lower];

    \addplot[cneplotlinear]      table[x=d,  y=CNE]   \map; \leg{\rox}
    \addplot [name path=upper,draw=none, forget plot] table[x=d,y expr=\thisrow{CNE}+\thisrow{CNE-std}] {\map};
    \addplot [name path=lower,draw=none, forget plot] table[x=d,y expr=\thisrow{CNE}-\thisrow{CNE-std}] {\map};
    \addplot [fill=\cneplotlinearc!\stdgrad, forget plot] fill between[of=upper and lower];

    %\legend{};
\end{axis}
\end{tikzpicture}
  
        \label{fig:oxford_paris_hard_neighbors}
    \end{subfigure}
}
% \vspace{-4pt}
\caption{\textbf{Impact of TLDR hyper-parameters} with a linear encoder and $\dd=128$. Dashed (solid) lines are for RParis6K-Mean (ROxford5K-Mean).  (Left) Impact of the \emph{auxiliary} dimension $\ddd$ and the number of hidden layers in the projector. (Right) Impact of the number of neighbors $k$. We see how the algorithm is robust to the number of neighbors used. }
\label{fig:roxford5k_paris6k_ablative}
\end{center}
\end{figure}

\ykt{
% -------------------------------------------------------------------------
\subsection{Varying the projector architectures and the number of neighbors}
\label{sec:varyingprojectorandk}
Figure~\ref{fig:roxford5k_paris6k_ablative} studies the role of some of our
parameters. On the left size of Figure~\ref{fig:roxford5k_paris6k_ablative}, we
vary the architecture of the projector $\pg$, an important module of \tldr. We see that having hidden layers generally helps. As also noted in~\citet{zbontar2021barlow}, having a high auxiliary dimension $\ddd$ for computing the loss is very important and highly impacts performance. 
On the right side of Figure~\ref{fig:roxford5k_paris6k_ablative} we show the surprisingly consistent performance of \tldr across a wide range of numbers of neighbors $k$.
}

% -------------------------------------------------------------------------
% Projector dims
% -------------------------------------------------------------------------
% -------------------------------------------------------------------------

\subsection{Varying the size of the training set} 
\label{sec:varyingtrainingsize}

\begin{figure}
\begin{center}
% -------------------------------------------------------------------------
% Oxford results
% -------------------------------------------------------------------------
\resizebox{\linewidth}{!}{
    \begin{subfigure}{.5\linewidth}
        \centering
        \begin{tikzpicture}
\begin{axis}[%
  width=\linewidth,
  xlabel={Number of images (in thousands)},
  xtick = {10, 20, 40, 80, 160, 320, 640, 1280},
  xticklabels = {10, 20, 40, 80, 160, 320, 640, 1280},
  ylabel={mAP},
  legend pos=south east,
  ylabel near ticks, xlabel near ticks, height=4.5cm, xmode=log,
  title={ROxford5K-Medium},
  ]

\pgfplotstableread{
d CNE CNE-std PCA-whiten
15 0.54457 0.01304 0.57158
30 0.56852 0.00405 0.57998
45 0.57653 0.00773 0.57574
75 0.59421 0.00997 0.57788
150 0.60992 0.00492 0.58681
300 0.61031 0.00810 0.58898
600 0.61147 0.00247 0.58780
1500 0.62220 0.00581 0.58579
}{\map}
    
    \addplot[cneplotlinear]      table[x=d,  y=CNE]   \map; \leg{\lcne}
    \addplot [name path=upper,draw=none, forget plot] table[x=d,y expr=\thisrow{CNE}+\thisrow{CNE-std}] {\map};
    \addplot [name path=lower,draw=none, forget plot] table[x=d,y expr=\thisrow{CNE}-\thisrow{CNE-std}] {\map};
    \addplot [fill=\cneplotlinearc!\stdgrad, forget plot] fill between[of=upper and lower];
    
    \addplot[pcaplot]      table[x=d,  y=PCA-whiten]   \map; \leg{PCA}

\end{axis}
\end{tikzpicture}
  
        % \caption{ROxford5K-Medium test set.}
        \label{fig:roxford5k_medium_trperc}
    \end{subfigure}%
    \begin{subfigure}{.5\linewidth}
        \centering
        \begin{tikzpicture}
\begin{axis}[%
  width=\linewidth,
  xlabel={Number of images (in thousands)},
  xtick = {10, 20, 40, 80, 160, 320, 640, 1280},
  xticklabels = {10, 20, 40, 80, 160, 320, 640, 1280},
  ylabel={mAP},
  legend pos=south east,
  ylabel near ticks, xlabel near ticks, height=4.5cm, xmode=log,
  title={ROxford5K-Hard},
  ]

\pgfplotstableread{
d CNE CNE-std PCA-whiten
15 0.27048 0.01736 0.30602
30 0.29634 0.00205 0.30959
45 0.30005 0.01739 0.30825
75 0.33034 0.02021 0.31083
150 0.35314 0.01180 0.33084
300 0.34658 0.01331 0.33213
600 0.35331 0.00473 0.33200
1500 0.36830 0.00892 0.31321
}{\map}
    
    \addplot[cneplotlinear]      table[x=d,  y=CNE]   \map; \leg{\lcne}
    \addplot [name path=upper,draw=none, forget plot] table[x=d,y expr=\thisrow{CNE}+\thisrow{CNE-std}] {\map};
    \addplot [name path=lower,draw=none, forget plot] table[x=d,y expr=\thisrow{CNE}-\thisrow{CNE-std}] {\map};
    \addplot [fill=\cneplotlinearc!\stdgrad, forget plot] fill between[of=upper and lower];
    
    \addplot[pcaplot]      table[x=d,  y=PCA-whiten]   \map; \leg{PCA}

\end{axis}
\end{tikzpicture}
  
        % \caption{ROxford5K-Hard test set.}
        \label{fig:roxford5k_hard_trperc}
    \end{subfigure}
}
% -------------------------------------------------------------------------
% Paris results
% -------------------------------------------------------------------------
\resizebox{\linewidth}{!}{
    \begin{subfigure}{.5\linewidth}
        \centering
        \begin{tikzpicture}
\begin{axis}[%
  width=\linewidth,
  xlabel={Number of images (in thousands)},
  xtick = {10, 20, 40, 80, 160, 320, 640, 1280},
  xticklabels = {10, 20, 40, 80, 160, 320, 640, 1280},
  ylabel={mAP},
  legend pos=south east,
  ylabel near ticks, xlabel near ticks, height=4.5cm, xmode=log,
  title={RParis6K-Medium},
  ]

\pgfplotstableread{
d CNE CNE-std PCA-whiten
15 0.73287 0.00424 0.74136
30 0.73810 0.00254 0.74647
45 0.74627 0.00027 0.74455
75 0.75516 0.00345 0.74357
150 0.75733 0.00354 0.74416
300 0.76151 0.00404 0.74360
600 0.76549 0.00332 0.74167
1500 0.76974 0.00121 0.74679
}{\map}
    
    \addplot[cneplotlinear]      table[x=d,  y=CNE]   \map; \leg{\lcne}
    \addplot [name path=upper,draw=none, forget plot] table[x=d,y expr=\thisrow{CNE}+\thisrow{CNE-std}] {\map};
    \addplot [name path=lower,draw=none, forget plot] table[x=d,y expr=\thisrow{CNE}-\thisrow{CNE-std}] {\map};
    \addplot [fill=\cneplotlinearc!\stdgrad, forget plot] fill between[of=upper and lower];
    
    \addplot[pcaplot]      table[x=d,  y=PCA-whiten]   \map; \leg{PCA}

\end{axis}
\end{tikzpicture}
  
        % \caption{RParis6K-Medium test set.}
        \label{fig:rparis6k_medium_trperc}
    \end{subfigure}%
    \begin{subfigure}{.5\linewidth}
        \centering
        \begin{tikzpicture}
\begin{axis}[%
  width=\linewidth,
  xlabel={Number of images (in thousands)},
  xtick = {10, 20, 40, 80, 160, 320, 640, 1280},
  xticklabels = {10, 20, 40, 80, 160, 320, 640, 1280},
  ylabel={mAP},
  legend pos=south east,
  ylabel near ticks, xlabel near ticks, height=4.5cm, xmode=log,
  title={RParis6K-Hard},
  ]

\pgfplotstableread{
d CNE CNE-std PCA-whiten
15 0.51580 0.00304 0.51369
30 0.52086 0.00716 0.52010
45 0.53021 0.00237 0.51675
75 0.54048 0.00549 0.51678
150 0.54416 0.00697 0.51939
300 0.55271 0.00401 0.51925
600 0.56148 0.00328 0.51494
1500 0.56466 0.00214 0.52378
}{\map}
    
    \addplot[cneplotlinear]      table[x=d,  y=CNE]   \map; \leg{\lcne}
    \addplot [name path=upper,draw=none, forget plot] table[x=d,y expr=\thisrow{CNE}+\thisrow{CNE-std}] {\map};
    \addplot [name path=lower,draw=none, forget plot] table[x=d,y expr=\thisrow{CNE}-\thisrow{CNE-std}] {\map};
    \addplot [fill=\cneplotlinearc!\stdgrad, forget plot] fill between[of=upper and lower];
    
    \addplot[pcaplot]      table[x=d,  y=PCA-whiten]   \map; \leg{PCA}

\end{axis}
\end{tikzpicture}
  
        % \caption{RParis6K-Hard test set.}
        \label{fig:rparis6k_hard_trperc}
    \end{subfigure}
}
\caption{\textbf{\tldr benefits from larger training sets.}
Impact of the size of the training set on performance. \tldr uses a linear encoder and $d = 128$.}

\label{fig:roxford5k_rparis6k_trperc}
\end{center}
\end{figure}
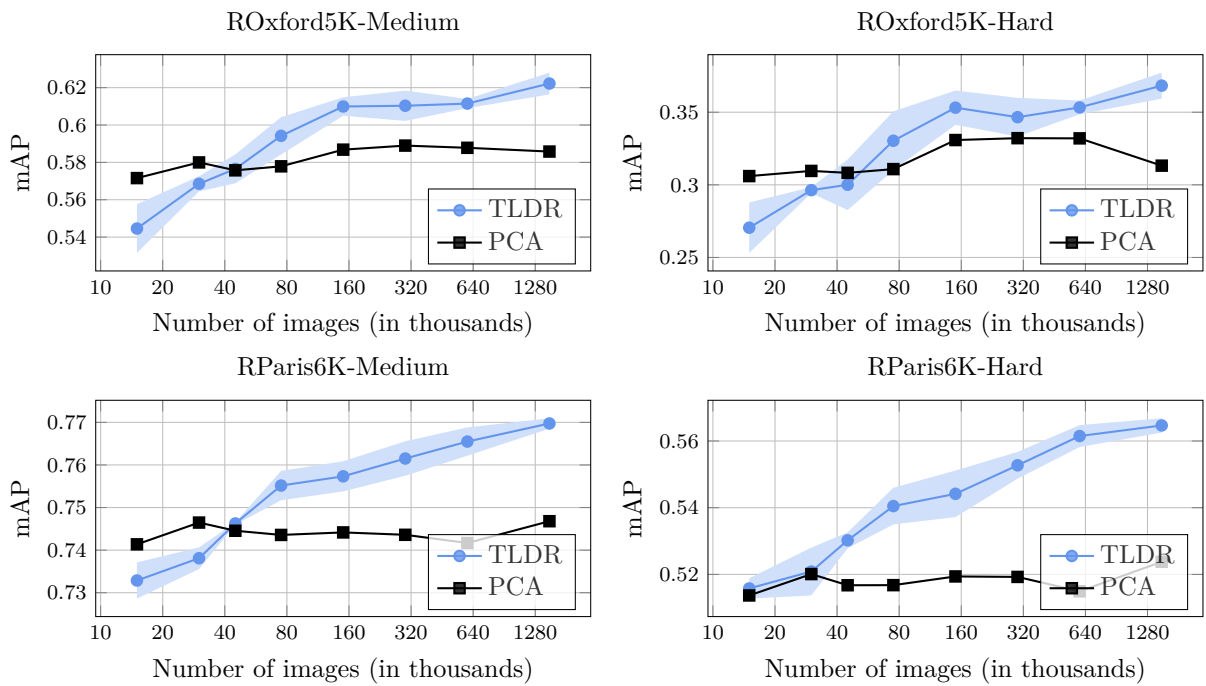

In Figure~\ref{fig:roxford5k_rparis6k_trperc} we show the impact of the size of the training set on \tldr's performance by randomly selecting subsets of images of increasing size from the Google Landmarks training set~\citep{weyand2020GLDv2}. As we see, PCA outperforms \tldr for a reduced number of images, however, it does not benefit from adding more data, keeping the same performance across all training set sizes. 
In contrast, \tldr does benefit from adding more data; all plots suggest that a larger training set could potentially boost the performance even further, increasing the gap with respect to PCA.

% -------------------------------------------------------------------------
% Training size
% -------------------------------------------------------------------------
% -------------------------------------------------------------------------

\subsection{Batch size ablation} 
\label{sec:varyingbatchsize}

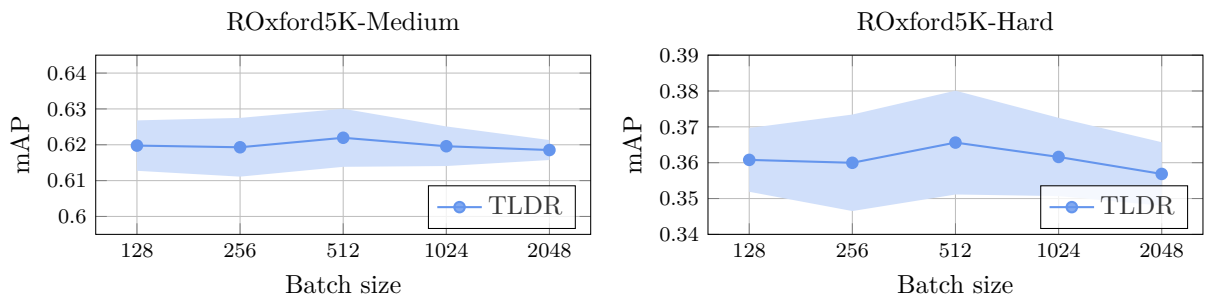
\begin{figure}
\begin{center}
\resizebox{\linewidth}{!}{
    \begin{subfigure}{.5\linewidth}
        \centering
        \begin{tikzpicture}
\begin{axis}[%
  width=\linewidth,
  xlabel={Batch size},
  xtick = {64, 128, 256, 512, 1024, 2048},
  xticklabels = {64, 128, 256, 512, 1024, 2048},
  ylabel={mAP}, ymin=0.595, ymax=0.645,
  legend pos=south east,
  ylabel near ticks, xlabel near ticks, height=4cm, xmode=log,
  title={ROxford5K-Medium},
  ]

\pgfplotstableread{
d CNE CNE-std PCA
128 0.61977 0.00703 nan
256 0.61930 0.00819 nan
512 0.62195 0.00810 nan
1024 0.61959 0.00552 nan
2048 0.61853 0.00278 nan
}{\map}

    \addplot[cneplotlinear]      table[x=d,  y=CNE]   \map; \leg{\lcne}
    \addplot [name path=upper,draw=none, forget plot] table[x=d,y expr=\thisrow{CNE}+\thisrow{CNE-std}] {\map};
    \addplot [name path=lower,draw=none, forget plot] table[x=d,y expr=\thisrow{CNE}-\thisrow{CNE-std}] {\map};
    \addplot [fill=\cneplotlinearc!\stdgrad, forget plot] fill between[of=upper and lower];

\end{axis}
\end{tikzpicture}
  
        %\caption{ROxford5K-Medium test set.}
        \label{fig:roxford5k_bsize_med}
    \end{subfigure}%
    \begin{subfigure}{.5\linewidth}
        \centering
        \begin{tikzpicture}
\begin{axis}[%
  width=\linewidth,
  xlabel={Batch size},
  xtick = {64, 128, 256, 512, 1024, 2048},
  xticklabels = {64, 128, 256, 512, 1024, 2048},
  ymin=0.34, ymax=0.39,
  ylabel={mAP},
  legend pos=south east,
  ylabel near ticks, xlabel near ticks, height=4cm, xmode=log,
  title={ROxford5K-Hard},
  ]

\pgfplotstableread{
d CNE CNE-std PCA
128 0.36078 0.00889 nan
256 0.35997 0.01346 nan
512 0.36561 0.01448 nan
1024 0.36160 0.01091 nan
2048 0.35686 0.00885 nan
}{\map}
    
    \addplot[cneplotlinear]      table[x=d,  y=CNE]   \map; \leg{\lcne}
    \addplot [name path=upper,draw=none, forget plot] table[x=d,y expr=\thisrow{CNE}+\thisrow{CNE-std}] {\map};
    \addplot [name path=lower,draw=none, forget plot] table[x=d,y expr=\thisrow{CNE}-\thisrow{CNE-std}] {\map};
    \addplot [fill=\cneplotlinearc!\stdgrad, forget plot] fill between[of=upper and lower];

\end{axis}
\end{tikzpicture}
  
        %\caption{ROxford5K-Hard test set.}
        \label{fig:roxfor5k_bsize_hard}
    \end{subfigure}
}
\caption{\textbf{The surprising stability of \tldr across batch sizes}.
Impact of the size of the training mini-batch on performance. \tldr uses a linear encoder and $d = 128$.}
\label{fig:roxford5k_bsize}
\end{center}
\end{figure}

Finally, in Figure~\ref{fig:roxford5k_bsize}, we show results of \tldr varying the size of the training mini-batch. Surprisingly, we observe it is stable across a wide range of values, allowing training \tldr under limited memory resources.

% -------------------------------------------------------------------------
% Batch size
% -------------------------------------------------------------------------
% -------------------------------------------------------------------------

% -------------------------------------------------------------------------
\subsection{Training time comparisons when varyin the size of the training set} 
\label{sec:varyingtrainingsizetime}

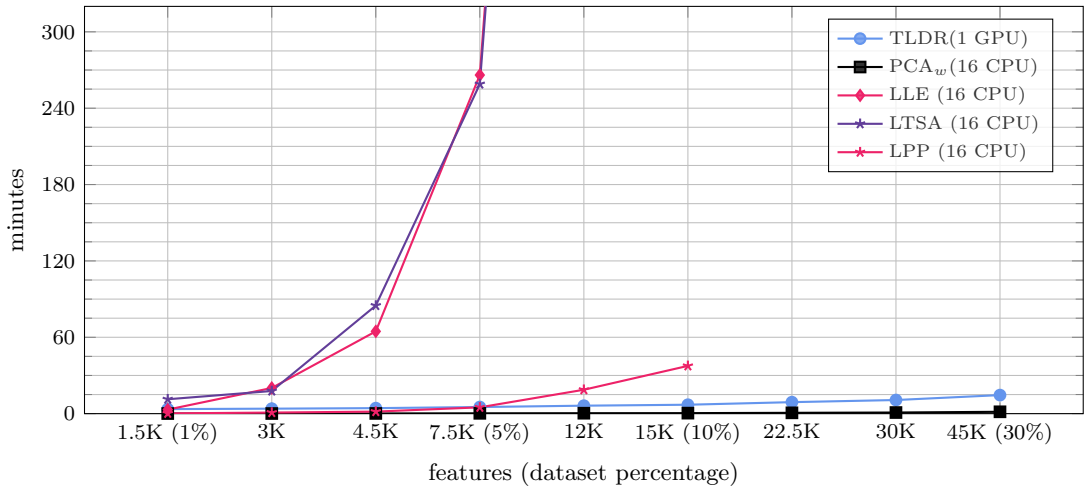
\begin{figure}
\begin{center}
% \resizebox{.5\linewidth}{!}{
        \begin{tikzpicture}
\begin{axis}[%
  width=.9\linewidth,
  xlabel={features (dataset percentage)},
  xtick = {1,2,3,4,5,6,7,8,9},
  xticklabels = {1.5K (1\%),3K , 4.5K ,7.5K (5\%),12K ,15K (10\%),22.5K, 30K, 45K (30\%)},
  ylabel={minutes},
  legend pos=north east,
  ylabel near ticks, xlabel near ticks, legend style={font=\scriptsize},
  height=7cm,
  minor y tick num=3,
  ytick = {0,60,120,180,240,300},
  yticklabels = {0,60,120,180,240,300},
  ymin=0, ymax=320,
  label style={font=\small},
  title style={font=\small}
  ]

\pgfplotstableread{
d  PCA-whiten  LLE LTSA LPP tldr
1   0.074 3.168 11.301 0.202 3.562
2	0.100 20.172 17.778 0.836 3.900
3	0.151 64.648 84.929 1.598 4.321
4	0.252 266.063 258.948 4.953 5.195
5	0.335 1360.462 1330.462 18.770 6.253
6	0.397 nan nan 37.514 7.002
7	0.636 nan nan nan 9.023
8	0.870 nan nan nan 10.697
9	1.448 nan nan nan 14.568
}{\map}

    \addplot[cneplotlinear, thick]      table[x=d,  y=tldr]   \map; \leg{\lcne (1 GPU)}
    
    \addplot[pcaplot, thick]      table[x=d,  y=PCA-whiten]   \map; \leg{\pcaw (16 CPU)}
    
    \addplot[cneplotgaussian, thick]      table[x=d,  y=LLE]   \map; \leg{LLE (16 CPU)}
    
        \addplot[mseplot, thick]      table[x=d,  y=LTSA]   \map; \leg{LTSA (16 CPU)}
    
    \addplot[lppplot, thick] table[x=d,  y=LPP]   \map; \leg{LPP (16 CPU)}

\end{axis}
\end{tikzpicture}
% }
\caption{\textbf{Training time as training dataset size increases} for $d = 32$. \tldr uses a linear encoder, a 2-hidden layer projector with $\ddd = 2048$ and is trained for 100 epochs.}
\label{fig:roxford5k_training_time}
\end{center}
\end{figure}

In Figure~\ref{fig:roxford5k_training_time} we report training time when learning a linear dimensionality reduction encoder while varying the size of the training set. We randomly select subsets of images of increasing size from the Google Landmarks training set~\citep{weyand2020GLDv2} (1.5M images) and measure the training time and \rox performance for \tldr and for a few manifold learning methods. All methods are run on the same servers. We used 16 CPUs and 300GB memory for all manifold learning methods, and one 32GB V100 GPU for \tldr. There is no publicly available easy-to-use GPU implementation for these methods, so we instead use the multi-core versions of scikit-learn.

The tables show that methods like LLE and LTSA scale exponentially with respect to both the time and memory needed. When sampling 5\% of the data, \ie 75k images, \tldr (and LPP) require 5 minutes to train while LLE and LTSA require over 4 hours. The LLE and LTSA runs took more than 3 (7) days when subsampling 8\% (10\%) of the data, while subsampling 15\% and above led to out-of-memory (OOM) related crashes despite 300GB of memory available. The LPP method is scaling better, but a) is 5x slower than TLDR when subsampling 10\% of the dataset and b) leads to OOM when subsampling 15\%. 
Obviously, reported times highly depend on their implementation. Here, we used the popular scikit-learn multi-core implementations for LLE and LTSA and the best implementation of LPP we could find\footnote{\url{https://github.com/jakevdp/lpproj}}.

% -------------------------------------------------------------------------
% Training Time
% -------------------------------------------------------------------------
% -------------------------------------------------------------------------

% -------------------------------------------------------------------------

% -------------------------------------------------------------------------
\subsection{ResNet-101 features} 
\label{sec:resnetoneowone}
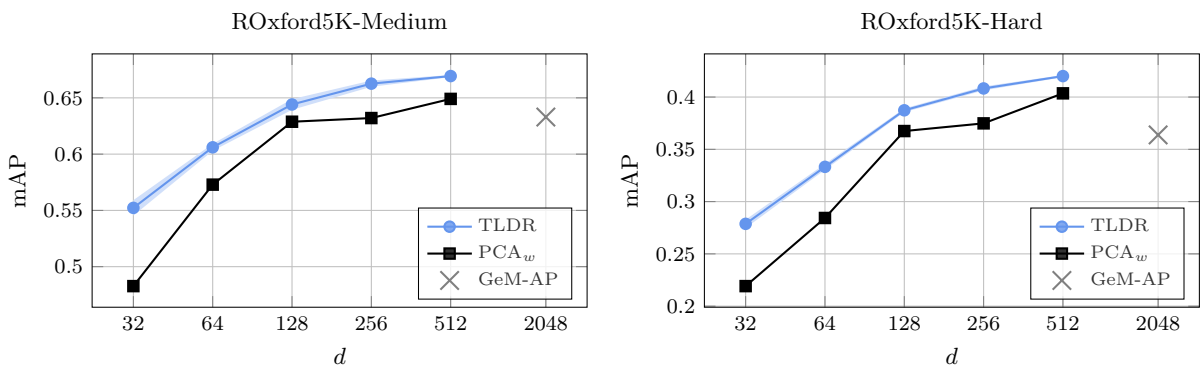
\begin{figure}
\begin{center}
\resizebox{\linewidth}{!}{
    \begin{subfigure}{.5\linewidth}
        \centering
        \begin{tikzpicture}
\begin{axis}[%
  width=\linewidth,
  xlabel={$\dd$},
  xtick = {1,2,3,4,5,6.2},
  xticklabels = {32,64,128,256,512,2048},
  ylabel={mAP},
  legend pos=south east,
  ylabel near ticks, xlabel near ticks, legend style={font=\scriptsize},
  height=5cm,
  title={ROxford5K-Medium},
  label style={font=\small},
  title style={font=\small}
  ]

\pgfplotstableread{
d CNE_0-h2048 CNE_0-h2048-std PCA-whiten baseline GeM-APw GeMw
1 0.55224 0.00686 0.48272 nan nan nan
2 0.60606 0.00310 0.57289 nan nan nan
3 0.64413 0.00474 0.62879 nan nan nan
4 0.66262 0.00277 0.63205 nan nan nan
5 0.66936 0.00095 0.64908 nan nan nan
6.2 nan nan nan 0.632956 0.675 0.678
}{\map}
    
    \addplot[cneplotlinear]      table[x=d,  y=CNE_0-h2048]   \map; \leg{\lcne}
    \addplot [name path=upper,draw=none, forget plot] table[x=d,y expr=\thisrow{CNE_0-h2048}+\thisrow{CNE_0-h2048-std}] {\map};
    \addplot [name path=lower,draw=none, forget plot] table[x=d,y expr=\thisrow{CNE_0-h2048}-\thisrow{CNE_0-h2048-std}] {\map};
    \addplot [fill=\cneplotlinearc!\stdgrad, forget plot] fill between[of=upper and lower];

    \addplot[pcaplot]      table[x=d,  y=PCA-whiten]   \map; \leg{\pcaw}

    \addplot[baselineplot] table[x=d,  y=baseline]   \map; \leg{\gemap}

\end{axis}
\end{tikzpicture}
        \label{fig:roxford5k_medium_r101}
    \end{subfigure}%
    \begin{subfigure}{.5\linewidth}
        \centering
        \begin{tikzpicture}
\begin{axis}[%
  width=\linewidth,
  xlabel={$\dd$},
  xtick = {1,2,3,4,5,6.2},
  xticklabels = {32,64,128,256,512,2048},
  ylabel={mAP},
  legend pos=south east,
  ylabel near ticks, xlabel near ticks, legend style={font=\scriptsize},
  height=5cm,
  title={ROxford5K-Hard},
  label style={font=\small},
  title style={font=\small}
  ]

\pgfplotstableread{
d CNE_0-h2048 CNE_0-h2048-std PCA-whiten baseline GeM-APw GeMw
1 0.27872 0.00400 0.21916 nan nan nan
2 0.33324 0.00282 0.28433 nan nan nan
3 0.38732 0.00178 0.36748 nan nan nan
4 0.40823 0.00188 0.37483 nan nan nan
5 0.42004 0.00105 0.40355 nan nan nan
6.2 nan nan nan 0.363779 0.428 0.417
}{\map}
    
    \addplot[cneplotlinear]      table[x=d,  y=CNE_0-h2048]   \map; \leg{\lcne}
    \addplot [name path=upper,draw=none, forget plot] table[x=d,y expr=\thisrow{CNE_0-h2048}+\thisrow{CNE_0-h2048-std}] {\map};
    \addplot [name path=lower,draw=none, forget plot] table[x=d,y expr=\thisrow{CNE_0-h2048}-\thisrow{CNE_0-h2048-std}] {\map};
    \addplot [fill=\cneplotlinearc!\stdgrad, forget plot] fill between[of=upper and lower];

    \addplot[pcaplot]      table[x=d,  y=PCA-whiten]   \map; \leg{\pcaw}

    \addplot[baselineplot] table[x=d,  y=baseline]   \map; \leg{\gemap}

\end{axis}
\end{tikzpicture}
        \label{fig:roxford5k_hard_r101}
    \end{subfigure}
}
\caption{\textbf{ResNet-101 features}. Mean average precision (mAP) on ROxford5K~\citep{radenovic2018revisiting} for different values of output dimensions $\dd$, using features obtained from the pre-trained ResNet-101 of~\citet{revaud2019aploss}.}

\label{fig:roxford5k_r101}
\end{center}
\end{figure}

We also experimented with features obtained from a larger pre-trained ResNet-101 model from~\citet{revaud2019aploss}\footnote{\url{https://github.com/naver/deep-image-retrieval}}. We see in Figure~\ref{fig:roxford5k_r101} that \tldr retains a significant gain over PCA and in fact surpasses the highest state-of-the-art numbers based on global features as reported in~\citet{tolias2020learning} for \rox.

% -------------------------------------------------------------------------
% R101 results
% -------------------------------------------------------------------------
% -------------------------------------------------------------------------

\ykt{
\subsection{Additional results with DINO features}
\label{sec:dinoresnetresults}

In Figure~\ref{fig:oxford_dino_r50_appendix} we present results on \rox using a \resnet DINO pretrained on \imnet; dimensionality reduction is learned on either \imnet or \gld. In Figure~\ref{fig:paris_dino_vit} we report performance on \rpa using a \vit DINO pretrained on \imnet.No labels are used at any stage. 
% Interestingly, we see that \rpa outperforms \tldr for 256 dimensions; this could possibly be due to  this to the overlap

\begin{figure}
\begin{center}
% -------------------------------------------------------------------------
% DINO  results
% -------------------------------------------------------------------------
\resizebox{\linewidth}{!}{
    \begin{subfigure}{.5\linewidth}
        \centering
        \begin{tikzpicture}
\begin{axis}[%
  width=\linewidth,
  xlabel={$\dd$},
  xtick = {2,3,4,5,6,7.5},
  xticklabels = {16,32,64,128,256,2048},
  ylabel={mAP},
  legend pos=south east,
  ylabel near ticks, xlabel near ticks, legend style={font=\scriptsize}, legend columns=2,
  height=5.5cm,
%   minor y tick num=4,
%   title={\rox},
  ytick = {0.1,0.15,0.2,0.25,0.29},
  yticklabels = {0.1,0.15,0.2,0.25,0.29},
  minor y tick num=1,
  label style={font=\small},
  title style={font=\small}
  ]

\pgfplotstableread{
d tldr-r-im-gld pca-r-im-gld dino-r-im tldr-r-im-im pca-r-im-im
% 1 0.12	0.08 nan 0.10	0.06
2 0.2025	0.1314 nan 0.1291	0.1014
3 0.2913	0.1928 nan 0.1582	0.1213
4 0.2904	0.2182 nan 0.1892	0.1550
5 0.2921	0.2546 nan 0.2146	0.1860
6 0.2879	0.2594 nan 0.2219	0.2114
7.5 nan nan 0.22 nan nan
}{\map}

    \addplot[cneplotlinear]     table[x=d,  y=tldr-r-im-gld]   \map; \leg{\tldr (\gld)}
    \addplot[pcaplot]           table[x=d,  y=pca-r-im-gld]   \map; \leg{\pcaw (\gld)}
    \addplot[cneplotlinear, dashed]     table[x=d,  y=tldr-r-im-im]   \map; \leg{\tldr (ImNet)}
    \addplot[pcaplot, dashed]           table[x=d,  y=pca-r-im-im]   \map; \leg{\pcaw (ImNet)}

    \addplot[baselineplot]      table[x=d,  y=dino-r-im]   \map; \label{dino_lbl} %\leg{DINO}
    
    % Second "Legend" node
\node [] at (rel axis cs: 0.913,0.68) {\scriptsize{{DINO}}};

\end{axis}
\end{tikzpicture}
        \caption{Results on \rox using DINO \resnet.}
        \label{fig:oxford_dino_r50_appendix}
    \end{subfigure}%
    \begin{subfigure}{.5\linewidth}
        \centering
        \begin{tikzpicture}
\begin{axis}[%
  width=\linewidth,
  xlabel={$\dd$},
  xtick = {2,3,4,5,6,6.5},
  xticklabels = {16,32,64,128,256,384},
  ylabel={mAP},
  legend pos=south east,
  ylabel near ticks, xlabel near ticks, legend style={font=\scriptsize}, legend columns=2,
  height=5.5cm,
%   minor y tick num=4,
%   title={DINO \vit (\imnet pretrained)},
%   ytick = {0.1,0.15,0.2,0.25,0.29},
%   yticklabels = {0.1,0.15,0.2,0.25,0.29},
  minor y tick num=1,
  label style={font=\small},
  title style={font=\small}
  ]

\pgfplotstableread{
d tldr-r-im-gld pca-r-im-gld dino-r-im tldr-r-im-im pca-r-im-im
2 0.4023	0.3373 nan 0.3345	0.2702
3 0.4723	0.3977 nan 0.4166	0.3270
4 0.5000	0.4505 nan 0.4539	0.3906
5 0.5278	0.5081 nan 0.4890	0.4739
6 0.5249	0.5497 nan 0.5062	0.4957
6.5 nan nan 0.4868 nan nan
}{\map}

    \addplot[cneplotlinear]     table[x=d,  y=tldr-r-im-gld]   \map; \leg{\tldr (\gld)}
    \addplot[pcaplot]           table[x=d,  y=pca-r-im-gld]   \map; \leg{\pcaw (\gld)}
    \addplot[cneplotlinear, dashed]     table[x=d,  y=tldr-r-im-im]   \map; \leg{\tldr (ImNet)}
    \addplot[pcaplot, dashed]           table[x=d,  y=pca-r-im-im]   \map; \leg{\pcaw (ImNet)}

    \addplot[baselineplot]      table[x=d,  y=dino-r-im]   \map; %\label{dino_lbl} %\leg{DINO}
    
    % Second "Legend" node
% \node [draw,fill=white] at (rel axis cs: 0.88,0.4) {\scriptsize{\ref{dino_lbl} DINO}};
\node [] at (rel axis cs: 0.92,0.6) {\scriptsize{{DINO}}};
    
\end{axis}
\end{tikzpicture}
        \caption{Results on \rpa using DINO \vit.}
        \label{fig:paris_dino_vit}
    \end{subfigure}
}
\caption{\textbf{Self-supervised landmark retrieval performance using a DINO backbones}. Both plots report mAP on \rox as a function of the output dimensions $\dd$. Left: Performance on \rox using a \resnet DINO pretrained on \imnet; dimensionality reduction is learned on either \imnet or \gld.
Right: Performance on \rpa using a \vit DINO pretrained on \imnet. No labels are used at any stage. 
  \label{fig:dino_extra_results}}
\end{center}
\end{figure}

}

\section{Additional experiments: Document retrieval}
\label{appendix:first_stage_retrieval}
In Section 3 of the main paper we studied first stage document retrieval under the task of argument retrieval, where both dimensionality reduction and evaluation are performed on datasets designed for the same task. In this section, we extend this evaluation protocol introducing a new task: duplicate query retrieval and now investigate not only dimensionality reduction for the same task, but also the case of \textit{dimensionality reduction transfer}. 

In the following paragraphs we first introduce the five datasets we use for first stage document retrieval, and then we discuss the additional experiments involving duplicate query datasets.

% -------------------------------------------------------------------------
% Summary of tasks and datasets for document retrieval
% -------------------------------------------------------------------------

\begin{table}
    \caption{Summary of \textbf{tasks and datasets} for the first stage document retrieval experiments presented in the Appendix.}
    \label{tab:task_summary_document_retrieval}
    \resizebox{\linewidth}{!}{
    \centering
    \begin{tabular}{l | c c | c | c c | c}
    \toprule
    Dataset & \# Documents & \# Queries & Avg positives per query & Avg query length & Avg document length & Retrieval type \\
    \midrule
    \multicolumn{7}{c}{Question answering (pretraining only)} \\
    \midrule
    MSMarco & 8.8M & 6980 & 1.1 & 6 & 56 & Asymmetric \\
    \midrule
    \multicolumn{7}{c}{Argument retrieval} \\
    \midrule
    ArguANA  & 8674 & 1406 & 1 & 193 & 167 & Asymmetric \\
    Webis-Touché 2020  & 380k & 49 & 49.2 & 7 & 292 & Asymmetric \\
    \midrule
    \multicolumn{7}{c}{Duplicate question retrieval} \\
    \midrule
    Quora  & 523k & 5000 & 1.6 & 10 & 11 & Symmetric \\
    CQADupStack & 457k & 13145 & 1.4 & 9 & 129 & Symmetric \\
    \bottomrule
    \end{tabular}
    }
    % \vspace{4pt}

\end{table}
% -------------------------------------------------------------------------

\begin{table}
    \caption{\textbf{Examples of queries and documents} from all the document retrieval datasets we use. Table extracted from~\cite{thakur2021beir}; Note the difference of length between query and document in some datasets.}
    \label{tab:examples}
    
    \small
    \resizebox{\textwidth}{!}{\begin{tabular}{l | l | l }
        \toprule
        \multicolumn{1}{l|}{\textbf{Dataset}}    &
        \multicolumn{1}{c}{\textbf{Query}}   &
        \multicolumn{1}{|c}{\textbf{Relevant-Document}} \\
        \midrule
   MSMARCO & \multicolumn{1}{p{8cm}|}{what fruit is native to australia} & \multicolumn{1}{p{12cm}}{\textit{<Paragraph>} Passiflora herbertiana. A rare passion fruit native to Australia. Fruits are green-skinned, white fleshed, with an unknown edible rating. Some sources list the fruit as edible, sweet and tasty, while others list the fruits as being bitter and inedible. assiflora herbertiana. A rare passion fruit native to Australia...} \\ \midrule
   ArguAna       & \multicolumn{1}{p{8cm}|}{Sexist advertising is subjective so would be too difficult to codify.  Effective advertising appeals to the social, cultural, and personal values of consumers. Through the connection of values to products, services and ideas, advertising is able to accomplish its goal of adoption... } & \multicolumn{1}{p{12cm}}{\textit{<Title>} media modern culture television gender house would ban sexist advertising \textit{<Paragraph>} Although there is a claim that sexist advertising is to difficult to codify, such codes have and are being developed to guide the advertising industry. These standards speak to advertising which demeans the status of women, objectifies them, and plays upon stereotypes about women which harm women and society in general. Earlier the Council of Europe was mentioned, Denmark, Norway and Australia as specific examples of codes or standards for evaluating sexist advertising which have been developed.} \\ \midrule
   Touche-2020 & \multicolumn{1}{p{8cm}|}{Should the government allow illegal immigrants to become citizens?} & \multicolumn{1}{p{12cm}}{\textit{<Title>} America should support blanket amnesty for illegal immigrants. \textit{<Paragraph>} Undocumented workers do not receive full Social Security benefits because they are not United States citizens " nor should they be until they seek citizenship legally. Illegal immigrants are legally obligated to pay taxes...} \\ \midrule 
   CQADupStack   & \multicolumn{1}{p{8cm}|}{Command to display first few and last few lines of a file} & \multicolumn{1}{p{12cm}}{\textit{<Title>} Combing head and tail in a single call via pipe \textit{<Paragraph>} On a regular basis, I am piping the output of some program to either `head` or `tail`. Now, suppose that I want to see the first AND last 10 lines of piped output, such that I could do something like ./lotsofoutput | headtail...} \\ \midrule
   Quora         & \multicolumn{1}{p{8cm}|}{How long does it take to methamphetamine out of your blood?} & \multicolumn{1}{p{12cm}}{\textit{<Paragraph>} How long does it take the body to get rid of methamphetamine?} \\ \midrule

    \end{tabular}}
    % \vspace{4pt}

\end{table}

\subsection{Tasks and dataset}

A summary of dataset statistics is available in Table~\ref{tab:task_summary_document_retrieval} and examples for each dataset are available in Table~\ref{tab:examples}.

\mypartight{MSMarco passages~\citep{nguyen2016msmarco}:} question and answer dataset based on Bing queries. Very sparse anotation with a high number of false negatives. Queries (Q) and Documents (D) are from different different domains due to size and content. Retrieval is asymmetric, because if you input a D as Q, the answer will not be D. Used only for pretraining as it has a set of training pairs for contrastive learning, while our aim is to perform self-supervision only. For this goal, we have chosen the other four datasets, that do not have a readily available set of training pairs (or triplets) for training, and thus self-supervision or unsupervised learning is required.

\mypartight{ArguAna~\citep{wachsmuth2018retrieval}:} Counter-argument retrieval dataset. Queries and documents belong to the same domain, with some queries being a part of the corpus, which makes it not suitable for training on this dataset. Queries and documents come from the same domain in both size and content, however associated query-document pairs have inverse context (Q defends a point, D is a rebuttal of Q), so input Q should not retrieve Q, if it is on the database. Retrieval is asymmetric as a query should not retrieve itself. 

\mypartight{Webis-Touché 2020~\citep{bondarenko2020overview, wachsmuth2017building}:} Argument retrieval dataset. Queries and documents are from different domains due to size and content, with queries being questions and documents being support arguments for the question. Retrieval is asymmetric as a query should not retrieve itself. 

\mypartight{CQADupStack~\citep{hoogeveen2016cqadupstack}:} Duplicate question retrieval from StackExchange subforums, composed of 12 different subforums. Corpuses are concatenated during training and mean result over all corpuses is used for testing (\ie every corpus has equal weight even if the number of queries is different). Queries are titles of recent submissions, while documents are concatenation of titles and descriptions of existing ones. Queries and documents are from different domains due to size and content, with the query domain being a part of the document one (Queries are contained in the documents). Retrieval is symmetric as a query should return itself.

\mypartight{Quora:} Duplicate question retrieval from the Quora platform. Queries are titles of recent submissions, while documents are titles of existing ones. Queries and documents are from the same domain concerning size and content. Retrieval is symmetric as a query should return itself.

\subsection{Experimental results}

We now test different combinations of the previously introduced datasets as dimensionality reduction and test datasets. The setup for all experiments is the same as the experiments in the main paper. In order to summarize our results (9 combinations of train/test datasets), we consider $\dd=64$ (second highest we investigate) as the comparison mark between PCA and \method. If \method outperforms PCA for all $\dd \leq 64$, we consider that it performed better than PCA, and otherwise we consider that PCA performed better than \method. Note that in all cases the lower the dimension the better \method performed against PCA. We also report which version of \method performed better, $L>0$ means that factorized linear is better than linear and $L=0$ the opposite. We present a summary of the experimental results in Table~\ref{tab:summary_results}, and provide depictions of some experiments in Figures~\ref{fig:arguana_ndcg} through~\ref{fig:webis_quoraa}. From the results presented on the table, we derive two conclusions about \method:

\begin{table}
    \caption{\textbf{Summary of the results on document retrieval.} (L=0) and (L>0) indicate which version of \method had better perfomance (linear and factorized linear respectively). Note that arguana is not suitable for training (not represented) and that we are not interested in using the same dataset for dimensionality reduction and test (thus the empty cells). }
    \label{tab:summary_results}
    
\resizebox{\textwidth}{!}{
\begin{tabular}{ccc|c|c|c|c|}
\cline{4-7}
                                                       &                                                          &             & \multicolumn{4}{c|}{Test dataset}                                               \\ \cline{4-7} 
                                                       &                                                          &             & \multicolumn{2}{c|}{Argument retrieval}  & \multicolumn{2}{c|}{Duplicate query} \\ \cline{4-7} 
                                                       &                                                          &             & ArguAna                      & Webis-Touché 2020    & Quora          & CQADupStack         \\ \hline
\multicolumn{1}{|l|}{\multirow{3}{*}{Dimensionality reduction}} & \multicolumn{1}{c|}{Argument Retrieval} & Webis-Touché 2020 & \method (L\textgreater{}0)                           &         & PCA (L\textgreater{}0)             &  PCA (L=0)                 \\  \cline{2-7} 
\multicolumn{1}{|l|}{}                                 &  \multicolumn{1}{c|}{\multirow{2}{*}{Duplicate Question}} & Quora       & \method (L\textgreater{}0) & \method (L=0) &              & PCA (L=0)                 \\ 
\multicolumn{1}{|l|}{}                                 & \multicolumn{1}{l|}{}                                    & CQADupStack & \method (L\textgreater{}0) & \method (L=0) & PCA (L\textgreater{}0)            &                   \\ \hline
\end{tabular}
}
% \vspace{4pt}

\end{table}

\begin{enumerate}
    \item \textbf{Differences in retrieval from pretraining to dimensionality reduction impacts results :} Looking into evaluation on argument retrieval, \method outperforms PCA. On the other hand, looking into evaluations on duplicate query retrieval, PCA is always able to outperform \method for $\dd \geq 64$. We infer that this must be derived from the difference in retrieval condition, as in all tests with asymmetric retrieval \method is able to outperform PCA. Note that symmetric retrieval and same domain for document and queries differs from the original pretraining task, and we posit that PCA is more robust to this type of change (which does not happen in our image retrieval experiments). Although we have this initial suspicion validated with 4 datasets a proper conclusion would need more dataset-pairs for experimentation, which we leave for future work.
    \item \textbf{Choosing linear or factorized linear depends on the statistics of the dataset:} Analyzing the results we are able to detect that the choice of which version of \method one should use depends on the length of queries and documents of the original dataset. If both lengths are equal, factorized linear is better (ArguANA and Quora), if not then linear is the better choice (Webis-Touché 2020 and CQADupStack). Even if by using ANCE representations we should not need to deal with these differences (we only tackle embeddings of fixed size), the statistics of the resulting embedding is different enough that it is detected by the batch normalization layer that is added for factorized linear.
\end{enumerate}

\begin{figure}
\begin{center}
\resizebox{\linewidth}{!}{
    \begin{subfigure}{.47\linewidth}
        \centering
        \begin{tikzpicture}
\begin{axis}[%
  width=\linewidth,
  height=5cm,
  xlabel={$\dd$},
  xtick = {1,2,3,4,5,7.5},
  xticklabels = {8,16,32,64,128,768},
  ylabel={Recall@100},
  ylabel near ticks,
  xlabel near ticks,
  minor y tick num=3,
  legend pos=south east,
   legend style={font=\footnotesize},
  every tick label/.append style={font=\footnotesize}
  ]

\pgfplotstableread{
    d cnelinear cnelinear-std cnelinearone cnelinearone-std cnelineartwo cnelineartwo-std pca baseline
    1 73.31 1.17 76.89 0.61 76.84 1.37 50.14 nan
    2 86.67 0.72 88.98 0.46 88.68 0.89 73.83 nan
    3 92.50 0.19 93.97 0.39 94.01 0.19 87.84 nan
    4 94.45 0.28 95.68 0.18 95.68 0.19 93.67 nan
    5 94.56 0.27 96.26 0.24 96.22 0.15 95.66 nan
    7.5 nan nan nan nan nan nan nan 94.02
    }{\map}

   \addplot[cneplotlinear]      table[x=d,  y=cnelinear]   \map; \leg{\lcne}
    \addplot [name path=upper,draw=none, forget plot] table[x=d,y expr=\thisrow{cnelinear}+\thisrow{cnelinear-std}] {\map};
    \addplot [name path=lower,draw=none, forget plot] table[x=d,y expr=\thisrow{cnelinear}-\thisrow{cnelinear-std}] {\map};
    \addplot [fill=\cneplotlinearc!\stdgrad, forget plot] fill between[of=upper and lower];   
   
   \addplot[cneplotflinearone]      table[x=d,  y=cnelinearone]   \map; \leg{\flcne{1}}
    \addplot [name path=upper,draw=none, forget plot] table[x=d,y expr=\thisrow{cnelinearone}+\thisrow{cnelinearone-std}] {\map};
    \addplot [name path=lower,draw=none, forget plot] table[x=d,y expr=\thisrow{cnelinearone}-\thisrow{cnelinearone-std}] {\map};
    \addplot [fill=\cneplotflinearonec!\stdgrad, forget plot] fill between[of=upper and lower];

   \addplot[cneplotflineartwo]      table[x=d,  y=cnelineartwo]   \map; \leg{\flcne{2}}
    \addplot [name path=upper,draw=none, forget plot] table[x=d,y expr=\thisrow{cnelineartwo}+\thisrow{cnelineartwo-std}] {\map};
    \addplot [name path=lower,draw=none, forget plot] table[x=d,y expr=\thisrow{cnelineartwo}-\thisrow{cnelineartwo-std}] {\map};
    \addplot [fill=\cneplotflineartwoc!\stdgrad, forget plot] fill between[of=upper and lower];
   
   \addplot[pcaplot]      table[x=d,  y=pca]   \map; \leg{\pcaw}
   \addplot[baselineplot] table[x=d,  y=baseline]   \map; \leg{ANCE}

\end{axis}
\end{tikzpicture}
        \label{fig:arguana_recall_nl_appendix}
    \end{subfigure}
    \begin{subfigure}{.47\linewidth}
        \centering
        \begin{tikzpicture}
\begin{axis}[%
  width=\linewidth,
  height=5cm,
  xlabel={$\dd$},
  xtick = {1,2,3,4,5,7.5},
  xticklabels = {8,16,32,64,128,768},
  ylabel={NDCG@10},
  ylabel near ticks,
  xlabel near ticks,
  legend pos=south east,
   legend style={font=\footnotesize},
  every tick label/.append style={font=\footnotesize}
  ]

\pgfplotstableread{
    d cnelinear cnelinear-std cnelinearone cnelinearone-std cnelineartwo cnelineartwo-std pca baseline
    1 19.38 0.60 20.91 1.00 20.88 0.31 09.61 nan
    2 31.22 0.66 32.59 0.64 32.66 0.58 22.97 nan
    3 38.73 0.29 39.67 0.41 39.63 0.51 36.15 nan
    4 41.88 0.27 43.17 0.32 42.99 0.28 42.47 nan
    5 43.11 0.21 45.02 0.30 45.09 0.38 45.77 nan
    7.5 nan nan nan nan nan nan nan 43.34
    }{\map}

   \addplot[cneplotlinear]      table[x=d,  y=cnelinear]   \map; \leg{\lcne}
    \addplot [name path=upper,draw=none, forget plot] table[x=d,y expr=\thisrow{cnelinear}+\thisrow{cnelinear-std}] {\map};
    \addplot [name path=lower,draw=none, forget plot] table[x=d,y expr=\thisrow{cnelinear}-\thisrow{cnelinear-std}] {\map};
    \addplot [fill=\cneplotlinearc!\stdgrad, forget plot] fill between[of=upper and lower];   
   
   \addplot[cneplotflinearone]      table[x=d,  y=cnelinearone]   \map; \leg{\flcne{1}}
    \addplot [name path=upper,draw=none, forget plot] table[x=d,y expr=\thisrow{cnelinearone}+\thisrow{cnelinearone-std}] {\map};
    \addplot [name path=lower,draw=none, forget plot] table[x=d,y expr=\thisrow{cnelinearone}-\thisrow{cnelinearone-std}] {\map};
    \addplot [fill=\cneplotflinearonec!\stdgrad, forget plot] fill between[of=upper and lower];

   \addplot[cneplotflineartwo]      table[x=d,  y=cnelineartwo]   \map; \leg{\flcne{2}}
    \addplot [name path=upper,draw=none, forget plot] table[x=d,y expr=\thisrow{cnelineartwo}+\thisrow{cnelineartwo-std}] {\map};
    \addplot [name path=lower,draw=none, forget plot] table[x=d,y expr=\thisrow{cnelineartwo}-\thisrow{cnelineartwo-std}] {\map};
    \addplot [fill=\cneplotflineartwoc!\stdgrad, forget plot] fill between[of=upper and lower];
   
   \addplot[pcaplot]      table[x=d,  y=pca]   \map; \leg{PCA}
   \addplot[baselineplot] table[x=d,  y=baseline]   \map; \leg{ANCE}

\end{axis}
\end{tikzpicture}
        \label{fig:arguana_ndcg_nl}
    \end{subfigure}
}
\caption{\textbf{Argument retrieval results on the ArguAna dataset using Webis-Touché 2020 for dimensionality reduction} for different values of output dimensions $\dd$. On the left we present Recall@100 and on the right we present NDCG@10.}
\label{fig:arguana_ndcg}
\end{center}
\end{figure}
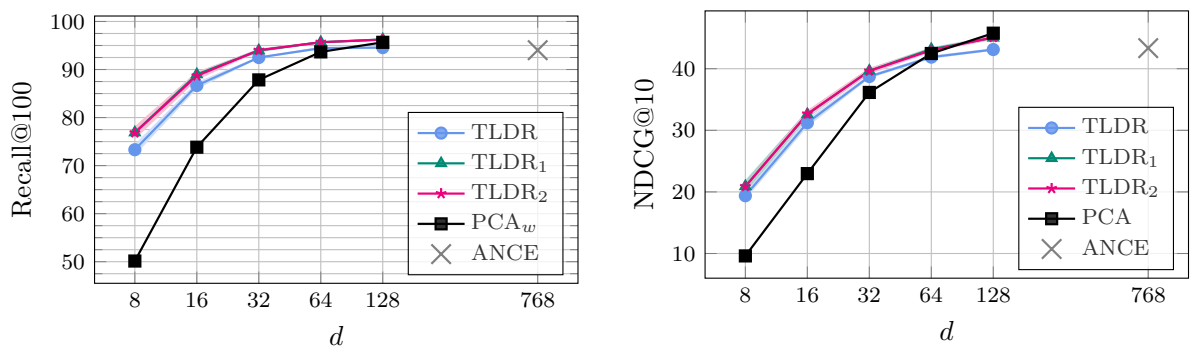  

\begin{figure}
\begin{center}
\resizebox{\linewidth}{!}{
    \begin{subfigure}{.47\linewidth}
        \centering
        \begin{tikzpicture}
\begin{axis}[%
  width=\linewidth,
  height=5cm,
  xlabel={$\dd$},
  xtick = {1,2,3,4,5,7.5},
  xticklabels = {8,16,32,64,128,768},
  ylabel={Recall@100},
  ylabel near ticks,
  xlabel near ticks,
  legend pos=south east,
   legend style={font=\footnotesize},
  every tick label/.append style={font=\footnotesize}
  ]

\pgfplotstableread{
    d cnelinear cnelinear-std cnelinearone cnelinearone-std cnelineartwo cnelineartwo-std pca baseline
    1 03.25 0 nan 0 04.11 0 01.06 nan
    2 14.30 0 nan 0 12.28 0 07.37 nan
    3 20.96 0 nan 0 18.59 0 15.56 nan
    4 25.90 0 nan 0 22.65 0 23.33 nan
    5 28.76 0 nan 0 23.96 0 27.10 nan
    7.5 nan nan nan nan nan nan nan 27.48
    }{\map}

   \addplot[cneplotlinear]      table[x=d,  y=cnelinear]   \map; \leg{\lcne}
    \addplot [name path=upper,draw=none, forget plot] table[x=d,y expr=\thisrow{cnelinear}+\thisrow{cnelinear-std}] {\map};
    \addplot [name path=lower,draw=none, forget plot] table[x=d,y expr=\thisrow{cnelinear}-\thisrow{cnelinear-std}] {\map};
    \addplot [fill=\cneplotlinearc!\stdgrad, forget plot] fill between[of=upper and lower];

   \addplot[cneplotflineartwo]      table[x=d,  y=cnelineartwo]   \map; \leg{\flcne{2}}
    \addplot [name path=upper,draw=none, forget plot] table[x=d,y expr=\thisrow{cnelineartwo}+\thisrow{cnelineartwo-std}] {\map};
    \addplot [name path=lower,draw=none, forget plot] table[x=d,y expr=\thisrow{cnelineartwo}-\thisrow{cnelineartwo-std}] {\map};
    \addplot [fill=\cneplotflineartwoc!\stdgrad, forget plot] fill between[of=upper and lower];
   
   \addplot[pcaplot]      table[x=d,  y=pca]   \map; \leg{PCA}
   \addplot[baselineplot] table[x=d,  y=baseline]   \map; \leg{ANCE}

\end{axis}
\end{tikzpicture}
        % \caption{Recall@100}
        \label{fig:quora_webis_recall}
    \end{subfigure}
    \begin{subfigure}{.47\linewidth}
        \centering
        \begin{tikzpicture}
\begin{axis}[%
  width=\linewidth,
  height=5cm,
  xlabel={$\dd$},
  xtick = {1,2,3,4,5,7.5},
  xticklabels = {8,16,32,64,128,768},
  ylabel={NDCG@10},
  ylabel near ticks,
  xlabel near ticks,
  legend pos=south east,
   legend style={font=\footnotesize},
  every tick label/.append style={font=\footnotesize}
  ]

\pgfplotstableread{
    d cnelinear cnelinear-std cnelinearone cnelinearone-std cnelineartwo cnelineartwo-std pca baseline
    1 01.75 0 nan 0 03.24 0 00.92 nan
    2 09.45 0 nan 0 08.75 0 05.98 nan
    3 17.72 0 nan 0 13.79 0 17.19 nan
    4 23.07 0 nan 0 15.66 0 21.46 nan
    5 24.85 0 nan 0 18.83 0 24.70 nan
    7.5 nan nan nan nan nan nan nan 25.42
    }{\map}

   \addplot[cneplotlinear]      table[x=d,  y=cnelinear]   \map; \leg{\lcne}
    \addplot [name path=upper,draw=none, forget plot] table[x=d,y expr=\thisrow{cnelinear}+\thisrow{cnelinear-std}] {\map};
    \addplot [name path=lower,draw=none, forget plot] table[x=d,y expr=\thisrow{cnelinear}-\thisrow{cnelinear-std}] {\map};
    \addplot [fill=\cneplotlinearc!\stdgrad, forget plot] fill between[of=upper and lower];

   \addplot[cneplotflineartwo]      table[x=d,  y=cnelineartwo]   \map; \leg{\flcne{2}}
    \addplot [name path=upper,draw=none, forget plot] table[x=d,y expr=\thisrow{cnelineartwo}+\thisrow{cnelineartwo-std}] {\map};
    \addplot [name path=lower,draw=none, forget plot] table[x=d,y expr=\thisrow{cnelineartwo}-\thisrow{cnelineartwo-std}] {\map};
    \addplot [fill=\cneplotflineartwoc!\stdgrad, forget plot] fill between[of=upper and lower];
   
   \addplot[pcaplot]      table[x=d,  y=pca]   \map; \leg{PCA}
   \addplot[baselineplot] table[x=d,  y=baseline]   \map; \leg{ANCE}

\end{axis}
\end{tikzpicture}
        % \caption{NDCG@10}
        \label{fig:quora_webis_ndcg}
    \end{subfigure}
}
\caption{\textbf{Argument retrieval results on the Webis-Touché 2020 dataset using Quora for dimensionality reduction} for different values of output dimensions $\dd$. On the left we present Recall@100 and on the right we present NDCG@10.}
\label{fig:quora_webis}
\end{center}
\end{figure}
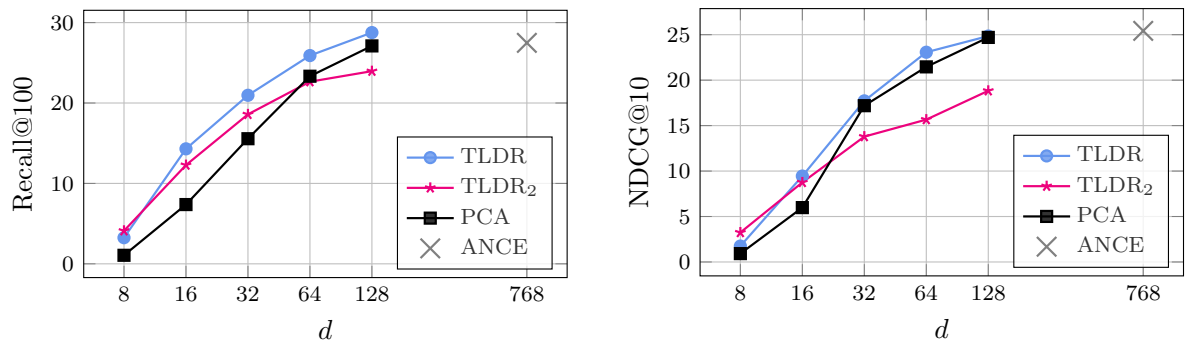  

\begin{figure}
\begin{center}
\resizebox{\linewidth}{!}{
    \begin{subfigure}{.47\linewidth}
        \centering
        \begin{tikzpicture}
\begin{axis}[%
  width=\linewidth,
  height=5cm,
  xlabel={$\dd$},
  xtick = {1,2,3,4,5,7.5},
  xticklabels = {8,16,32,64,128,768},
  ylabel={Recall@100},
  ylabel near ticks,
  xlabel near ticks,
  legend pos=south east,
   legend style={font=\footnotesize},
  every tick label/.append style={font=\footnotesize}
  ]

\pgfplotstableread{
    d cnelinear cnelinearone cnelineartwo pca baseline
    1 11.15 0.0 12.25 08.98 nan
    2 26.32 0.0 27.24 22.40 nan
    3 42.90 0.0 42.45 39.65 nan
    4 53.59 nan 53.39 53.99 nan
    5 59.62 nan 59.92 61.36 nan
    7.5 nan nan nan nan 60.84
    }{\map}

   \addplot[cneplotlinear]      table[x=d,  y=cnelinear]   \map; \leg{\flcne{0}}
   \addplot[cneplotflineartwo]      table[x=d,  y=cnelineartwo]   \map; \leg{\flcne{2}}
   \addplot[pcaplot]      table[x=d,  y=pca]   \map; \leg{PCA}
   \addplot[baselineplot] table[x=d,  y=baseline]   \map; \leg{baseline}

\end{axis}
\end{tikzpicture}
        % \caption{Recall@100}
        \label{fig:quora_cqadupstack_recall}
    \end{subfigure}
    \begin{subfigure}{.47\linewidth}
        \centering
        \begin{tikzpicture}
\begin{axis}[%
  width=\linewidth,
  height=5cm,
  xlabel={$\dd$},
  xtick = {1,2,3,4,5,7.5},
  xticklabels = {8,16,32,64,128,768},
  ylabel={NDCG@10},
  ylabel near ticks,
  xlabel near ticks,
  legend pos=south east,
   legend style={font=\footnotesize},
  every tick label/.append style={font=\footnotesize}
  ]

\pgfplotstableread{
    d cnelinear cnelinearone cnelineartwo pca baseline
    1 01.59 0.0 01.90 01.13 nan
    2 07.83 0.0 07.78 05.99 nan
    3 17.78 0.0 17.43 16.39 nan
    4 26.17 nan 25.03 26.46 nan
    5 30.42 nan 29.51 31.17 nan
    7.5 nan nan nan nan 31.62
    }{\map}

   \addplot[cneplotlinear]      table[x=d,  y=cnelinear]   \map; \leg{\flcne{0}}
   \addplot[cneplotflineartwo]      table[x=d,  y=cnelineartwo]   \map; \leg{\flcne{2}}
   \addplot[pcaplot]      table[x=d,  y=pca]   \map; \leg{PCA}
   \addplot[baselineplot] table[x=d,  y=baseline]   \map; \leg{baseline}

\end{axis}
\end{tikzpicture}
        % \caption{NDCG@10}
        \label{fig:quora_cqadupstack_ndcg}
    \end{subfigure}
}
\caption{\textbf{Duplicate question retrieval results on the CQADupstack dataset using Quora for dimensionality reduction} for different values of output dimensions $\dd$. On the left we present Recall@100 and on the right we present NDCG@10.}
\label{fig:quora_cqadupstack}
\end{center}
\end{figure}  

% -------------------------------------------------------------------------
% CQAdupstack results
% -------------------------------------------------------------------------
\begin{figure}
\begin{center}
\resizebox{\linewidth}{!}{
    \begin{subfigure}{.47\linewidth}
        \centering
        \begin{tikzpicture}
\begin{axis}[%
  width=\linewidth,
  height=5cm,
  xlabel={$\dd$},
  xtick = {1,2,3,4,5,7.5},
  xticklabels = {8,16,32,64,128,768},
  ylabel={Recall@100},
  ylabel near ticks,
  xlabel near ticks,
  legend pos=south east,
   legend style={font=\footnotesize},
  every tick label/.append style={font=\footnotesize}
  ]

\pgfplotstableread{
    d cnelinear cnelinear-std cnelinearone cnelinearone-std cnelineartwo cnelineartwo-std pca baseline
    1 62.54 0.61 64.45 0.21 64.64 0.48 53.31 nan
    2 88.20 0.21 89.40 0.33 89.37 0.28 84.11 nan
    3 95.61 0.14 95.81 0.09 95.81 0.10 95.43 nan
    4 97.53 0.13 97.98 0.09 97.94 0.06 98.33 nan
    5 98.43 0.07 98.62 0.03 98.59 0.10 98.78 nan
    7.5 nan nan nan nan nan nan nan 98.68
    }{\map}

   \addplot[cneplotlinear]      table[x=d,  y=cnelinear]   \map; \leg{\lcne}
    \addplot [name path=upper,draw=none, forget plot] table[x=d,y expr=\thisrow{cnelinear}+\thisrow{cnelinear-std}] {\map};
    \addplot [name path=lower,draw=none, forget plot] table[x=d,y expr=\thisrow{cnelinear}-\thisrow{cnelinear-std}] {\map};
    \addplot [fill=\cneplotlinearc!\stdgrad, forget plot] fill between[of=upper and lower];   
   
   \addplot[cneplotflinearone]      table[x=d,  y=cnelinearone]   \map; \leg{\flcne{1}}
    \addplot [name path=upper,draw=none, forget plot] table[x=d,y expr=\thisrow{cnelinearone}+\thisrow{cnelinearone-std}] {\map};
    \addplot [name path=lower,draw=none, forget plot] table[x=d,y expr=\thisrow{cnelinearone}-\thisrow{cnelinearone-std}] {\map};
    \addplot [fill=\cneplotflinearonec!\stdgrad, forget plot] fill between[of=upper and lower];

   \addplot[cneplotflineartwo]      table[x=d,  y=cnelineartwo]   \map; \leg{\flcne{2}}
    \addplot [name path=upper,draw=none, forget plot] table[x=d,y expr=\thisrow{cnelineartwo}+\thisrow{cnelineartwo-std}] {\map};
    \addplot [name path=lower,draw=none, forget plot] table[x=d,y expr=\thisrow{cnelineartwo}-\thisrow{cnelineartwo-std}] {\map};
    \addplot [fill=\cneplotflineartwoc!\stdgrad, forget plot] fill between[of=upper and lower];
   
   \addplot[pcaplot]      table[x=d,  y=pca]   \map; \leg{PCA}
   \addplot[baselineplot] table[x=d,  y=baseline]   \map; \leg{ANCE}

\end{axis}
\end{tikzpicture}
        % \caption{Recall@100}
        \label{fig:webis_quora_recall}
    \end{subfigure}
    \begin{subfigure}{.47\linewidth}
        \centering
        \begin{tikzpicture}
\begin{axis}[%
  width=\linewidth,
  height=5cm,
  xlabel={$\dd$},
  xtick = {1,2,3,4,5,7.5},
  xticklabels = {8,16,32,64,128,768},
  ylabel={NDCG@10},
  ylabel near ticks,
  xlabel near ticks,
  legend pos=south east,
   legend style={font=\footnotesize},
  every tick label/.append style={font=\footnotesize}
  ]

\pgfplotstableread{
    d cnelinear cnelinear-std cnelinearone cnelinearone-std cnelineartwo cnelineartwo-std pca baseline
    1 32.16 0.32 34.16 0.23 34.39 0.47 25.23 nan
    2 66.93 0.28 67.61 0.19 67.61 0.33 61.70 nan
    3 78.94 0.17 79.13 0.12 79.26 0.08 78.99 nan
    4 82.75 0.12 83.03 0.13 82.89 0.04 84.14 nan
    5 84.91 0.05 85.04 0.16 85.02 0.22 85.75 nan
%    6 85.50 0.10 85.69 0.17 85.40 0.10 85.68 nan
    7.5 nan nan nan nan nan nan nan 85.67
    }{\map}

   \addplot[cneplotlinear]      table[x=d,  y=cnelinear]   \map; \leg{\lcne}
    \addplot [name path=upper,draw=none, forget plot] table[x=d,y expr=\thisrow{cnelinear}+\thisrow{cnelinear-std}] {\map};
    \addplot [name path=lower,draw=none, forget plot] table[x=d,y expr=\thisrow{cnelinear}-\thisrow{cnelinear-std}] {\map};
    \addplot [fill=\cneplotlinearc!\stdgrad, forget plot] fill between[of=upper and lower];   
   
   \addplot[cneplotflinearone]      table[x=d,  y=cnelinearone]   \map; \leg{\flcne{1}}
    \addplot [name path=upper,draw=none, forget plot] table[x=d,y expr=\thisrow{cnelinearone}+\thisrow{cnelinearone-std}] {\map};
    \addplot [name path=lower,draw=none, forget plot] table[x=d,y expr=\thisrow{cnelinearone}-\thisrow{cnelinearone-std}] {\map};
    \addplot [fill=\cneplotflinearonec!\stdgrad, forget plot] fill between[of=upper and lower];

   \addplot[cneplotflineartwo]      table[x=d,  y=cnelineartwo]   \map; \leg{\flcne{2}}
    \addplot [name path=upper,draw=none, forget plot] table[x=d,y expr=\thisrow{cnelineartwo}+\thisrow{cnelineartwo-std}] {\map};
    \addplot [name path=lower,draw=none, forget plot] table[x=d,y expr=\thisrow{cnelineartwo}-\thisrow{cnelineartwo-std}] {\map};
    \addplot [fill=\cneplotflineartwoc!\stdgrad, forget plot] fill between[of=upper and lower];
   
   \addplot[pcaplot]      table[x=d,  y=pca]   \map; \leg{PCA}
   \addplot[baselineplot] table[x=d,  y=baseline]   \map; \leg{ANCE}

\end{axis}
\end{tikzpicture}
        % \caption{NDCG@10}
        \label{fig:webis_quora_ndcg}
    \end{subfigure}
}
\caption{\textbf{Duplicate question retrieval results on the Quora dataset using Webis-Touché 2020 for dimensionality reduction} for different values of output dimensions $\dd$. On the left we present Recall@100 and on the right we present NDCG@10.}\label{fig:webis_quoraa}
\end{center}
\end{figure}
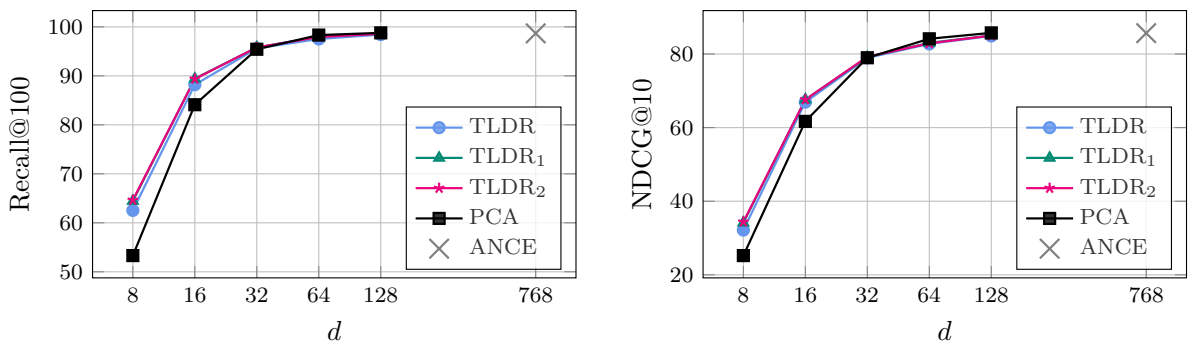  

% -------------------------------------------------------------------------
% FashionMNIST and 2D
\section{FashionMNIST: Learning from raw pixel data and visualization}
\label{sec:experiments_fashionmnist}

Although beyond the scope of what \method is designed for 
% \arxivorsub{(
(see also discussion at the end of Section~\ref{sec:discussion}),
% }{(see also discussion at the end of Section 5 in the main paper),}
in this section we present some basic results when using it for learning from raw pixel data and for visualizations, \ie when reducing the output dimension to only $\ddd = 2$.

\begin{figure*}[t]
\centering
    \begin{tikzpicture}
\begin{axis}[%
  width=.6\linewidth,
  xlabel={$\dd$},
  xtick = {1,2,3,4,5,6,7, 8},
  xticklabels = {2,8,32,64,128,256,512,784},
  ylabel={Accuracy},
  legend pos=outer north east,
    minor y tick num=3,
    ymin=0.5, ymax=0.88,
  ylabel near ticks, xlabel near ticks, 
  height=7cm,
  ]

\pgfplotstableread{
d TLDRmlp TLDR PCA UMAP PCAw ICA Isomap  
1 0.7592 0.6183 0.5597 0.7236 0.5602 0.5602 0.618
2 0.8205 0.8144 0.7768 0.8081 0.7823 0.7823 0.7851
3 0.8312 0.838 0.8274 0.3531 0.8354 0.8361 0.8095
4 0.835 0.84 0.8309 0.1931 0.8278 0.8265 0.8078
5 0.8348 0.8395 0.8303 0.1 0.7943 0.7955 0.8083
6 0.8361 0.8397 0.825 0.1 0.6624 0.6626 0.8085
7 0.8375 0.8391 0.8183 0.1 0.4548 0.4545 0.8037
8 0.8366 0.8382 0.8164 0.1 0.3735 0.3735 0.7979
}{\map}
    
    \addplot[cneplotlinear]      table[x=d,  y=TLDR]   \map; \leg{\lcne}
    
    \addplot[mseplot]      table[x=d,  y=PCA]   \map; \leg{PCA}
    
    \addplot[pcaplot]      table[x=d,  y=PCAw]   \map; \leg{\pcaw}
    
    \addplot[drlimplot]      table[x=d,  y=ICA]   \map; \leg{\icaw}

    \addplot[contrastiveplot] table[x=d,  y=UMAP]   \map; \leg{UMAP}
    
    \addplot[cneplotgaussian]      table[x=d,  y=Isomap]   \map; \leg{Isomap}

\end{axis}
\end{tikzpicture}
    \caption{\textbf{Results on the FashionMNIST dataset} as a function of the output dimensions $\dd$. We compare \method with PCA, PCA with whitening, UMAP and Isomap and report accuracy after $k^\prime$-NN classifiers (with $k^\prime=100$) following~\citep{mcinnes2018umap}. For \method and UMAP we set the number of neighbors $k = 100$. The performance of UMAP was very low for $\ddd > 32$. \label{fig:fashionmnist_knn}}
\end{figure*}
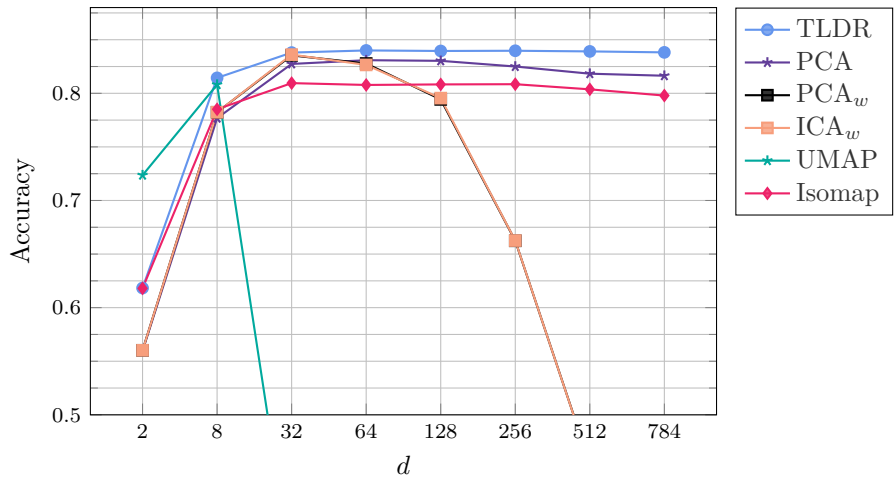

\mypartight{Learning from raw pixel data.} In Figure~\ref{fig:fashionmnist_knn} we present results when learning directly from raw pixel data. We use the predefined splits and, following related work~\citep{mcinnes2018umap}, we measure and report accuracy after $k$-NN classifiers. We see that \method retains its gains over any other manifold learning method we tested. We have to note however that these results have to be taken with a pinch of salt, as a) the input pixel space is relatively simple compared to higher resolution natural images and b) to achieve such results we use the prior knowledge that we only have 10 classes and set high values for hyper-parameter $k$, \ie $k=100$ for all methods compared.

% -------------------------------------------------------------------------
% FashionMNIST results
% -------------------------------------------------------------------------
% -------------------------------------------------------------------------

\begin{figure*}[t]
\centering
\begin{tabular}{c c}
     \includegraphics[width=0.45\linewidth]{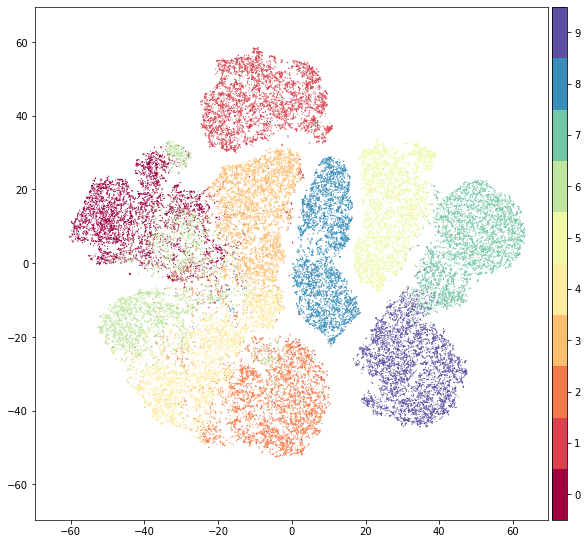} &  \includegraphics[width=0.45\linewidth]{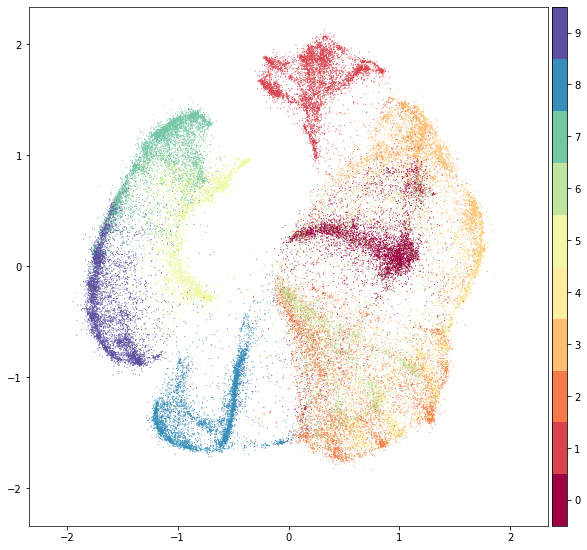} \\\vspace{20pt}
     \tsne~\citep{van2008visualizing} & MDE~\citep{agrawal2021minimum} \\ 
     
    \includegraphics[width=0.45\linewidth]{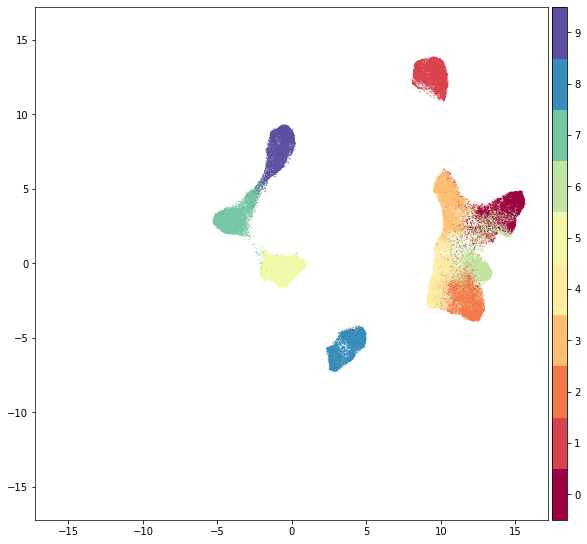} &
    \includegraphics[width=0.45\linewidth]{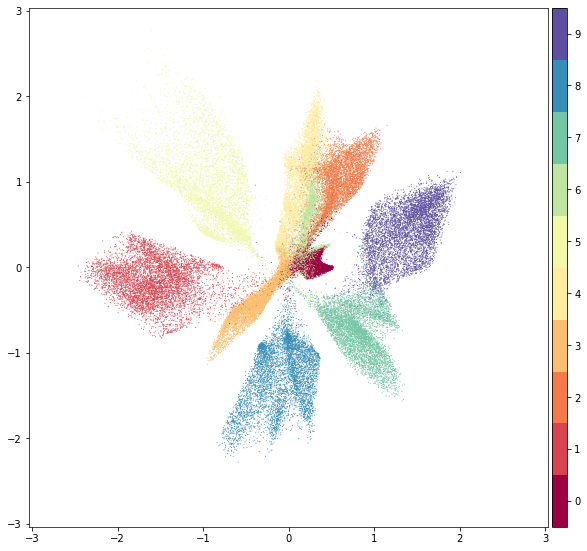} \\
     UMAP~\citep{mcinnes2018umap} ($k=100$)& \method (ours) ($k=100$)
\end{tabular}

\caption{\textbf{2D visualizations of the training set of FashionMNIST}. From top to bottom and left to right: \tsne, MDE, UMAP and \method.}
\label{fig:2dviz}
\end{figure*}

\mypartight{2D visualizations.}
Let us first clarify that \tldr was not created with 2D outputs in mind; in fact, there are other excellent choices for visualization like \tsne, UMAP~\citep{mcinnes2018umap}, TriMAP~\citep{amid2019trimap} or the recent Minimum-Distortion Embedding (MDE)~\citep{agrawal2021minimum} that we would use instead. 
In Figure~\ref{fig:2dviz} we show 2d visualizations when reducing the 60k training set of FashionMNIST to $\dd=2$ dimensions. We present results for \method, \tsne~\citep{van2008visualizing}, UMAP~\citep{mcinnes2018umap} and PyMDE~\citep{agrawal2021minimum}. It is interesting how \method seems to be optimized for linear separability even for 2-dimensional outputs. For visualizations, we used the pyMDE library\footnote{\href{https://pymde.org/}{https://pymde.org/}} provided by the authors of~\citep{agrawal2021minimum}.

% -------------------------------------------------------------------------

\section{Further discussions and related works}
\label{appendix:discussion}

% \onlyinsub{
% \mypartight{Graph diffusion for harder \knn pairs.} \citet{iscen2018mining} improve the method presented in~\citet{hadsell2006dimensionality} by mining harder positives and negative pairs for the contrastive loss via diffusion over the \knn graph. Similar to~\citet{hadsell2006dimensionality}, they are interested in learning (fine-tuning) the whole network and not just the dimensionality-reduction layer. Although it would be interesting to incorporate such ideas in \tldr, we consider it complementary and beyond the scope of this paper. Methods used for learning descriptor matching are also related; \eg \citep{simonyan2014learning} formulates dimensionality reduction as a convex optimisation problem. Although the redundancy reduction objective can be formulated in many ways, \eg via  stochastic proximal gradient methods like Regularised Dual Averaging in~\citep{simonyan2014learning}, we believe that the \emph{simplicity, immediacy and clarity} in which the Barlow Twins objective optimizes the output space is a strong advantage of \method.
% }

\mypartight{Graph diffusion for query expansion.} 
For the task of retrieval, assuming access to the search (test) database, methods like~\citep{iscen2017efficient,iscen2018fast,liu2019guided} utilize manifold learning on the the \knn graph of the database to facilitate \textit{query expansion}. We note that while these methods have shown great empirical performance on the same image retrieval datasets as we experiment on, we do not directly compare to them as their methodology and goals greatly differs from ours. We aim at being \textit{invariant to the target dataset} (thus not performing learning on them), differently from the aforementioned methods %that use the target dataset for learning and 
they need access to the target dataset for learning, and to its \knn graph during testing. \method is complementary to such  graph diffusion techniques for query expansion.

\mypartight{Relation to knowledge distillation.} 
Knowledge distillation (KD)~\citep{hinton2015distilling} aims at transferring knowledge from a pre-trained teacher network to a student one, often %used in the case of neural network compression. 
for neural network compression. 
One way to perform KD is relational KD (RKD)~\citep{park2019relational,tian2019contrastive,lin2020distilling}, which transfers knowledge using relations between samples such as distance and angles. \method can be seen as a method for RKD. It enforces the student network (encoder) to reproduce a relational property (neighborhood) found on the teacher (the input space). However, there are some main differences to traditional distillation methods: \begin{inlinelist} \item the application: self-supervised retrieval instead of supervised classification~\citep{hinton2015distilling}, contrastive~\citep{tian2019contrastive,lin2020distilling} or self-supervision for classification~\citep{fang2021seed}\item the definition of the relations: abstract (neighbors), instead of measurable ones (distance, angle), which avoids normalization problems due to the dimensionality difference between teacher and student \item the link between teacher and student: in our case, the student becomes a part of the teacher network at the end, instead of being a separate network.\end{inlinelist}

\mypartight{Relation to node embedding.} 
Node embedding methods aim at generating representations to graph nodes that are representative of the sample and its relations on the graph. In that sense, \method could be seen as learning embeddings for nodes on a graph. Compared to the traditional methods in this space, such as LINE~\citep{tang2015line}, Node2Vec~\citep{grover2016node2vec}, DeepWalk~\citep{perozzi2014deepwalk},  \method has three clear differences: i) does not rely on the edge strength; ii) regularization of the space based on the decorrelation of dimensions instead of L2-norm or orthogonality; and iii) only the 1-hop neighborhood information is used. More recent node embedding solutions are based on deep learning architectures that incorporate diffusion properties in the architecture like GCNs~\citep{kipf2016semi,you2020does}, while \method achieves a similar effect via the Barlow Twins loss. 

\subsection{Limitations of our work}
\label{sec:limitations}

In the context of document retrieval we also tested \method on another task: duplicate question retrieval. In duplicate question retrieval \method was only able to outperform PCA for the lower dimension values (d=8,16,32). We posit that \method does not achieve significant gains for the rest of the dimensions because duplicate task differs too much  from the original pretraining task (QA on MSMarco dataset) in that the duplicate retrieval is symmetric (the documents retrieved by a query should also appear when we use the document as query), while pretraining and argument retrieval is assymetric. In order to verify this, we performed ablations with different pairs of (dimensionality reduction,target dataset) and confirm that if the target dataset is a duplicate retrieval task \method is not able to outperform the compared method, but if we use duplicate retrieval only for dimensionality reduction and test on argument retrieval \method is able to outperform the compared methods. For full discussion and results \cf Section~\ref{appendix:first_stage_retrieval}.

\begin{algorithm2e}[tb]
   \caption{PyTorch-style pseudocode of \method.}
   \label{alg:pseudocode}
   
    \definecolor{codeblue}{rgb}{0,0.5,0.5}
    \lstset{
      basicstyle=\fontsize{7.2pt}{7.2pt}\ttfamily\bfseries,
      commentstyle=\fontsize{7.2pt}{7.2pt}\color{codeblue},
      keywordstyle=\fontsize{7.2pt}{7.2pt},
    }
\begin{lstlisting}[language=python]
# X: training set of M D-dimensional vectors
# D: input dimension
# d': projector's output dimension
# d: encoder's output dimension

# B: batch size
# k: number of neighbours
# compute_knn_graph: calculates the k nearest neighbours of each vector
# RandomBatchSampler: randomly samples batch_size indices

# Initialization
model = initialize_model()  # initialize both encoder f and projector g
N = compute_knn_graph(X, k)  # returns matrix of size Mxk

# Training
for e in num_epochs:
  for indices in RandomBatchSampler(len(X), batch_size=B):
    x = X[indices]  # BxD
    y = X[random_sample(N[indices], 1)]  # BxD
    x = model(x)  # Bxd'
    y = model(y)  # Bxd'
    loss = BarlowTwinsLoss(x, y)
    loss.backward()
    model.update()
    
model.discard_projector()  # discard projector g
Z = model(X)  # Mxd
   
\end{lstlisting}
\end{algorithm2e}

\section{Pseudocode of \method}
\label{sec:pseudocode}
In Algorithm~\ref{alg:pseudocode} we show the pseudocode of \method, which includes initialization, training, and projection.

\end{document}